\documentclass[msom]{informs3}

\OneAndAHalfSpacedXI

\usepackage{natbib}
 \bibpunct[, ]{(}{)}{,}{a}{}{,}%
 \def\bibfont{\small}%
\TheoremsNumberedThrough     
\ECRepeatTheorems

\EquationsNumberedThrough    

\MANUSCRIPTNO{}

\usepackage[titletoc]{appendix}
\usepackage[final]{microtype}
\usepackage{subcaption}
\usepackage{varwidth}
\usepackage{capt-of}
\usepackage{boldline}
\usepackage{epsfig}
\usepackage{latexsym}
\usepackage{url}
\usepackage{float}
\usepackage[hypertexnames=false,hyperfootnotes=false]{hyperref}
\usepackage{texnansi}
\usepackage{color}
\usepackage{tikz}
\usepackage{afterpage}
\usepackage{enumerate}
\usepackage[normalem]{ulem}
\usepackage{lmodern}
\usepackage{rotating}
\usepackage{graphicx}
\usepackage{caption}
\usepackage{multicol}
\usepackage{booktabs}
\usepackage{ifthen}
\usepackage{enumitem}
\setlist[enumerate]{topsep=0pt,itemsep=-1ex,partopsep=1ex,parsep=1ex}

\usepackage[compact]{titlesec}

\makeatletter
\g@addto@macro{\UrlBreaks}{\UrlOrds}
\makeatother

\newcommand{\R}{\ensuremath{\field{R}}} 

\newtheorem{informal-theorem}{Theorem}
\newtheorem{informal-proposition}{Proposition}

\usetikzlibrary{arrows,patterns,plotmarks}
\tikzstyle{every picture} += [>=stealth]

\makeatletter
\def\@seccntformat#1{\csname the#1\endcsname.\quad}
\makeatother

\floatstyle{ruled}
\provideboolean{fastcompile}
\newcommand{\hidefastcompile}[1]{\ifthenelse{\boolean{fastcompile}}{}{#1}}

\usepackage[top=1in, bottom=1in, left=1in, right=1in]{geometry}

\def\draft{1}  

\usepackage{dsfont}
\usepackage{tcolorbox}

\usepackage{mathtools}

\usepackage{algorithm}
\usepackage[noend]{algpseudocode}
\usepackage[normalem]{ulem}

\ifnum\draft=1

\else

\fi

\def\<{\langle}
\def\>{\rangle}

\def\R{\mathbb{R}}

\newcommand{\norm}[1]{\left\lVert\mspace{1mu} #1 \mspace{1mu}\right\rVert}

\newcommand{\tr}{{\rm tr}}
\newcommand{\E}[1] {{\mathbb{E}}\left(#1\right)}
\newcommand{\Ex}[2] {{\mathbb{E}}_{#1}\left(#2\right)}
\newcommand {\Prob}[1] {{\mathbb{P}}\left(#1\right)}
\newcommand {\Probx}[2] {{\mathbb{P}}_{#1}\left(#2\right)}

\newcommand {\sgn} {\mathrm{sgn}}

\newcommand {\F } {{\mathrm{F}}}

\newcommand {\Ber} {{\rm Ber}}
\newcommand {\1}[1] {{\mathds{1}}\left\{#1\right\}}

\newcommand{\TPR}{\mathrm{TPR}}
\newcommand{\FPR}{\mathrm{FPR}}
\newcommand{\cost}{\mathrm{cost}}
\newcommand{\Ps}{\mathcal{P^{*}}}
\newcommand{\Ano}{\mathrm{Anom}}
\newcommand{\Pos}{\mathrm{Poisson}}
\newcommand{\SVD}{\mathrm{SVD}}
\newcommand{\AD}{\mathrm{EW}}
\newcommand{\EW}{\mathrm{EW}}
\newcommand{\PAD}{\mathcal{P}^{\mathrm{EW}}}
\newcommand{\rank}{\mathrm{rank}}
\newcommand{\LL}{\mathrm{L}}
\newcommand{\RR}{\mathrm{R}}
\renewcommand{\O}{\mathrm{O}}
\newcommand{\A}{\mathrm{A}}

\usepackage[capitalise]{cleveref}

\begin{document}




\TITLE{Fixing Inventory Inaccuracies At Scale}

\ARTICLEAUTHORS{%
\AUTHOR{Vivek F. Farias}
\AFF{Operations Research Center, Massachusetts Institute of Technology, \EMAIL{vivekf@mit.edu}} 
\AUTHOR{Andrew A. Li}
\AFF{Tepper School of Business, Carnegie Mellon University, 
\EMAIL{aali1@cmu.edu}}

\AUTHOR{Tianyi Peng}
\AFF{Department of Aeronautics and Astronautics, Massachusetts Institute of Technology, 
\EMAIL{tianyi@mit.edu}}
} 

\ABSTRACT{%
\noindent Inaccurate records of inventory occur frequently, and by some measures cost retailers approximately 4\% in annual sales. Detecting inventory inaccuracies manually is cost-prohibitive, and existing algorithmic solutions rely almost exclusively on learning from longitudinal data, which is insufficient in the dynamic environment induced by modern retail operations. Instead, we propose a solution based on cross-sectional data over stores and SKUs, observing that detecting inventory inaccuracies can be viewed as a problem of identifying anomalies in a (low-rank) Poisson matrix. State-of-the-art approaches to anomaly detection in low-rank matrices apparently fall short. Specifically, from a theoretical perspective, recovery guarantees for these approaches require that non-anomalous entries be observed with vanishingly small noise (which is not the case in our problem, and indeed in many applications). 

So motivated, we propose a conceptually simple entry-wise approach to anomaly detection in low-rank Poisson matrices. Our approach accommodates a general class of probabilistic anomaly models. We show that the cost incurred by our algorithm approaches that of an optimal algorithm at a min-max optimal rate.  Using synthetic data and real data from a consumer goods retailer, we show that our approach provides up to a 10$\times$ cost reduction over incumbent approaches to anomaly detection. 
Along the way, we build on recent work that seeks entry-wise error guarantees for matrix completion, establishing such guarantees for sub-exponential matrices, a result of independent interest.  }%


\KEYWORDS{inventory record inaccuracies, phantom inventory, anomaly detection, matrix completion, high-dimensional inference} 

\maketitle

%


\section{Introduction} \label{sec:intro}

Consider the problem of tracking the inventory level of a single product (SKU) at a brick and mortar shop. Among the slew of complex operational processes that a retailer performs, this one would appear to be relatively simple, so it is perhaps surprising that in fact inaccurate records of inventory occur frequently: past audits of major retail chains have found errors in 51--65\% of inventory records across all stores and SKUs \citep{raman2001execution,kang2005information,dehoratius2008inventory,rekik2019modeling}. These inventory inaccuracies are costly. For example, one set of consequences of these inaccuracies are events referred to as {\em phantom inventory}, wherein inventory is recorded to be present {\em on-shelf} while in reality it is missing, perhaps due to shrinkage/theft \citep{de2008break,fan2014benefits}, misplacement by employees or shoppers \citep{ton2010effect,wang2016impact}, point-of-sale errors \citep{nachtmann2010impact}, among other reasons. Detecting inventory inaccuracies is the motivation for this work.

To appreciate the difficulty of maintaining accurate inventory records, it is worth briefly reviewing a few `obvious' potential solutions, and why they have failed so far:
\begin{enumerate}
\item Manual Auditing: Having employees perform manual store counts would certainly be accurate, but again these inaccuracies occur frequently, to the point that the amount of labor needed to correct and counteract them is simply too costly \citep{chuang2015inventory}.
\item Technology: The most promising technological solution to date is based on RFID tracking, and indeed there is empirical evidence that this can reduce inventory inaccuracies \citep{lee2007unlocking,hardgrave2013rfid,goyal2016effectiveness}. However, after at least two decades of development, the cost of integrating RFID inventory systems (including RFID tags, scanners, networking, building infrastructure, etc.) has not yet fallen to the point that there is widespread adoption among retailers.
\item Ignore the Problem: While the previous approaches ultimately fail due to cost, the cost (e.g. in lost sales) of {\em ignoring} inventory inaccuracies is equally high -- phantom inventory events alone cost the retail industry up to 4\% in annual revenue \citep{fleisch2005inventory}.
\end{enumerate}
These problems are only exacerbated in the modern retail environment, which adds the complexity of omnichannel components such as ship-to-store \citep{akturk2021exploring} fulfillment, ship-from-store \citep{li2020reinvent} fulfillment, and third-party shopping and delivery.

\vspace{1em}
\noindent
\textsf{\textbf{Algorithmic Solutions:}}
At first glance, the problem of detecting inventory inaccuracies would appear to lend itself to fairly straightforward algorithmic solutions. For example, in the case of phantom inventory: observe the sales transactions of the SKU over time, and loosely speaking, detect whether the rate of transactions has slowed or stopped. This solution works in principle if we have an accurate model of (a) the transactions as a stochastic process (a Poisson process with known rate, say), and (b) the effect of the phantom inventory event we wish to detect (stopping all transactions, say). Given these two ingredients, one can formalize the problem of detecting whether the event in question has occurred -- we will describe one such formalization later on -- and build `optimal' algorithms for detecting these events or mitigating their effects \citep{kok2007inspection,dehoratius2008retail}. Such {\em longitudinal} approaches, which rely only on the observation of single-SKU data over time, are cheap to implement, and largely make up the current state of the art. They are effectively the gold standard, again assuming two ingredients are known: a stochastic process for transactions, and the effect of the event to be detected.


The primary challenge when applying this type of procedure in practice is that {\em neither} of the two requisite ingredients are necessarily known in advance, due to rapid changes in customers' demand and the assortments offered by retailers. For example, it may be reasonable to model the transactions of a particular SKU over time as a Poisson process, but that process' rate may change as quickly as the amount of time it would take to detect that the process has stopped (i.e. a few inter-arrival times). It is effectively impossible then to detect whether a sequence of sales (particularly a lack of sales) is anomalous.

However, there is still hope for large retailers offering {\em many} SKUs across {\em multiple} locations. In particular, a retailer might reasonably assume that the demand rate for a particular SKU at a particular store is related to the demand for both (a) the same SKU at other stores and (b) other SKUs at the same store. This model, which is in the same vein as the famous `Netflix Prize' model \citep{bennett2007netflix}, suggests that while the inventory inaccuracy problem cannot be solved for a single SKU-store pair using its single stream of longitudinal data, it may be possible to detect inaccuracies simultaneously across all SKU-store pairs using {\em cross-sectional} data. {\em This is exactly what we seek to accomplish in this paper.}\footnote{There are many scenarios in which longitudinal approaches suffice, such as for high-velocity SKUs, or environments that are homogeneous over time. We view our work on leveraging cross-sectional data as {\em orthogonal} to those approaches, filling in the gap created by the scenarios in which longitudinal approaches fail.}

\vspace{1em}
\noindent
\textsf{\textbf{Detecting Anomalies in Matrices:}}
The problem of detecting inventory accuracies, or {\em anomalies}, across multiple SKUs and stores from cross-sectional sales data can be formulated as one of {\em anomaly detection in a low-rank matrix}. Specifically, let $M^*$ be the matrix of average demands, whose rows correspond to individual stores, whose columns correspond to individual SKUs, and whose entries are the {\em expected} demands during some, typically short, period -- say, a week. $M^*$ is unknown to us, and is assumed to be {\em low-rank}, which is a typical mathematical formalization \citep{grover1987simultaneous,ansari2000internet,ansari2003customization,farias2019learning} for relating a single SKU across multiple stores, and multiple SKUs at a single store. Similarly, let $Y$ be a random matrix of the same dimensions as $M^*$, whose entries are independent and have expected values matching the corresponding entries of $M^*$. We can think of $Y$ as the matrix of hypothetical sales, assuming no anomalous events (such as phantom inventory) have occurred.
Finally, let $X = Y + A$, where $A$ is an unknown, sparse (in the sense that many of its entries are zero) matrix of anomalies, capturing the effect of anomalous events on sales. We observe only $X$, and only on some subset of matrix entries $\Omega$. Matrix anomaly detection problems  concern identifying the support of $A$ simply from these observations.

The previous allusion to the Netflix Prize problem might suggest that the problem here has already been solved, in the same way that the Netflix Prize problem has been `solved' (e.g. via matrix completion algorithms). This is {\em not} the case, and it is worth understanding why. To begin, state-of-the-art approaches to solving this problem do indeed stem from algorithms for matrix completion; for instance, consider solving the following convex optimization problem (referred to as `stable PCP' by \cite{zhou2010stable}), where $\lambda_1$ and $\lambda_2$ are regularization parameters:
\begin{equation}\label{eq:stablePCP}
\min_{\hat Y, \hat A}
\|\hat{Y} \|_{*} + \lambda_2 \| \hat A \|_1 + \lambda_1 \| P_{\Omega}(X - \hat Y - \hat A )\|_{\F}^2.
\end{equation}
The three matrix norms in the objective are, from left to right, the nuclear norm to promote low-rankedness in $\hat{Y}$, the 1-norm to promote sparsity in $\hat{A}$, and the Frobenius norm to promote fit to $X$ on the observed entries $\Omega$.
Upon solving problem \eqref{eq:stablePCP}, ideally $\hat{Y} \approx Y$ and $\hat{A} \approx A$, and so we may use $\hat A$ to estimate the support of $A$.

Now in the absence of anomalies, this optimization problem (after removing the $\hat A$ terms) is in essence optimal under a variety on assumptions on the distributions of $Y$ and $\Omega$. In contrast, the available results for anomaly detection are weaker. In particular, approaches based on solving \cref{eq:stablePCP} or similar formulations, along with their corresponding theoretical results, are insufficient for at least the following reasons:

\begin{enumerate}
\item Do not allow for sufficient noise: Perhaps most limiting, without additional assumptions on the anomaly model, results that guarantee the recovery of $A$ require that the total observation `noise' $\| Y - M^* \|_{\F}$ be bounded by a {\em constant} independent of the size of the matrix. 
In contrast,
if we want to model $Y$ as, e.g., a matrix of Poisson entries with mean $M^*$, then clearly $\mathbb{E}  \| Y - M^* \|_{\F}$ will scale with the size of the matrix, so theoretical guarantees for extant matrix anomaly detection approaches do not apply.
\item Do not incorporate realistic cost structures: Existing results guaranteeing recovery of $A$ are measured in simple metrics such as the number of entries for which $\hat{A}$ and $A$ match in terms of being zero or non-zero.
Thus, they neglect to incorporate the potentially imbalanced costs of missing an anomaly (the cost being in lost sales) vs. falsely identifying an anomaly (the cost being sending an employee to fix a non-existent inventory problem).
\item Do not allow for useful distributions: For technical reasons, results for standard matrix completion techniques by and large make a statistical assumption (sub-Gaussianity) which precludes certain distributions that are standard for modeling sales, such as the Poisson distribution \citep{conrad1976sales, shi2014production}.
\item Perform poorly in practice: Even ignoring these theoretical limitations, we will see using real sales data that 
the optimization approach above can perform quite poorly.
\end{enumerate}

\vspace{1em}
\noindent
\textsf{\textbf{This Paper:}} Against the above backdrop, we make the following contributions to the problem of detecting inventory inaccuracies, and more generally anomaly detection in matrices:
\begin{enumerate}
\item {\bf A near-optimal algorithm:} We develop a new anomaly detection algorithm for low-rank Poisson\footnote{Sub-exponential, more generally.} matrices, and prove that our approach is min-max optimal (up to logarithmic terms) under a broad class of probabilistic anomaly models. These results demonstrate that, as alluded to previously, our algorithm is able to accurately detect inventory inaccuracies across SKUs and stores using a single snapshot of cross-sectional data, even when the underlying demand model and anomaly model are unknown.

We frame the anomaly detection problem itself as one of minimizing an extremely flexible cost function that is additive, but not necessarily identical, across entries of the matrix, and can penalize {\em false positives} (i.e. falsely identifying an anomaly) and {\em false negatives} (i.e. missing an anomaly) differently. Our main results are stated vis-$\grave{\mathrm{a}}$-vis this cost:
\begin{informal-theorem}(Informal)
Let all matrices be of size $n \times n$. Under mild assumptions, our algorithm achieves, with high probability, a cost satisfying
\begin{align*}
\mathrm{cost} \leq \mathrm{cost}^* + O\left(\frac{\log^{1.5} n}{\sqrt{n}}\right).
\end{align*}
\end{informal-theorem}
The term $\mathrm{cost}^*$ here represents the lowest achievable cost among all policies which {\em know} the average demands $M^*$ and underlying anomaly model (in contrast, our algorithm knows neither). A useful interpretation of $\mathrm{cost}^*$ is that it corresponds to the lowest cost given sufficient longitudinal data, for every store-SKU pair, in an idealized time-homogeneous environment. Thus, this result implies that our algorithm achieves, within an additive factor, that same idealized cost with just a single cross-sectional snapshot of data. Moreover, this additive factor shrinks as matrix size (meaning the number of SKUs and stores) increases, at a rate which is in fact optimal up to logarithmic terms:

\begin{informal-proposition}(Informal)
For any algorithm, there exists an instance such that its cost satisfies
\[ \mathrm{cost}  \geq \mathrm{cost}^* + \Omega\left(\frac{1}{\sqrt{n}}\right).  \]
\end{informal-proposition}

Our results are powered by two ingredients. The first is a new result for Poisson matrix completion described in the third contribution below. The second is we show that combined with a moment matching approach to learning the anomaly model, we can jointly learn the anomaly model {\em along with} the true underlying rate matrix. This in turn suffices to build an algorithm that we show is near optimal in the sense that it achieves the cost that converges to the optimal cost at a min-max optimal rate. The min-max optimality is established through a hypothesis testing argument.

\item {\bf Experimental validation:} Testing our algorithm on both synthetic data and real data from a national retailer, we find that our approach significantly outperforms the existing optimization approach to detecting anomalies. Indeed, our algorithm achieves a lower cost than the incumbent `Stable PCP' (and other existing benchmarks) by factors of 3 to 10 on synthetic and real data. 

\item {\bf Entry-wise guarantee for sub-exponential matrices:} As part of our approach, we prove a new result of independent interest for matrix completion with sub-exponential noise that, for the first time, bounds the {\em entry-wise} error under {\em sub-exponential} noise:
\begin{informal-theorem}(Informal)
Let $M^*$ be of size $n \times n$ and rank $r$, and assume that the entries of $M^*$ are observed independently with probability $p$, along with additive sub-exponential noise. Under mild assumptions, there exists an estimator $\hat{M}$ such that, with high probability,
\[ \|\hat{M} - M^*\|_\mathrm{max} \le O\left(\frac{r \log n}{p \sqrt{n}} \right).  \]
\end{informal-theorem}
This result substantially improves upon previous results for sub-exponential matrices, all of which bound an aggregate error measure \citep{lafond2015low,sambasivan2018minimax,cao2015poisson,mcrae2019low}. 

\end{enumerate}

The remainder of this paper is organized as follows: we conclude this section by reviewing the related literature and quickly establishing a base set of notation. \cref{sec:model} formally introduces our model for data and anomalies, along with the anomaly detection problem we seek to solve, and our main results. We describe our algorithm in \cref{sec:algorithm}, and our experimental results in \cref{sec:experiments}.
Proof sketches of the main results are given in \cref{sec:proof-sketches}. 
Finally, conclusions are drawn in \cref{sec:conclusion}.

\vspace{1em}
\subsection{Related Literature}
There are two ongoing streams of work to which the present paper contributes:

\vspace{1em}
\noindent
\textsf{\textbf{Inventory Inaccuracies:}} The first, naturally, is in inventory record inaccuracies, which are well-studied in Operations Management, e.g.~see the survey by \cite{mou2018retail}. The phenomenon itself has been observed for some time \citep{raman2001execution,dehoratius2008inventory,kang2005information,rekik2019modeling}, and  inventory inaccuracies remain a primary challenge for retailers \citep{chen2015analytics,fleisch2005inventory}.
Observed causes range from shrinkage/theft \citep{fan2014benefits,de2008break}, to misplacement \citep{wang2016impact,ton2010effect}, to point-of-sale errors \citep{nachtmann2010impact}.

The success and costs of non-algorithmic solutions like manual auditing \citep{chuang2015inventory}, technological solutions using \citep{lee2007unlocking,hardgrave2013rfid,goyal2016effectiveness}, and simply ignoring the problem \citep{fleisch2005inventory} have been studied. Existing algorithmic solutions \citep{kok2007inspection,dehoratius2008retail} have focused on  adapting inventory management policies to uncertain inventory levels. Algorithmic detection, particularly in a form that leverages cross-sectional data, i.e. observations across products and stores, is the motivation for this work.

\noindent
\textsf{\textbf{Matrix Anomaly Detection and Statistical Inference:}} The second body of work concerns anomaly detection for matrices. The majority of existing work has focused on a formulation called {\em robust principal component analysis (robust PCA)} \citep{candes2011robust,chandrasekaran2011rank}. Most relevant to our problem (which allows for noise) are approaches for \textit{noisy} robust PCA \citep{zhou2010stable,agarwal2012noisy,wong2017matrix,zhang2018robust,chen2020bridging}. Despite a sequence of breakthroughs and improvements in algorithms for optimizing objectives in noisy robust PCA \citep{lin2009fast,lin2010augmented,yuan2009sparse,aybat2016algorithms,ma2018efficient,netrapalli2014non,yi2016fast,zhang2018robust}, progress in statistical guarantees for these formulations has been relatively slower \citep{zhou2010stable,wong2017matrix,klopp2017robust}. See \cref{table:comparison} for a summary of existing statistical guarantees. Note that any hope of identifying the anomalies $A$ would require, at the very least, that $\|\hat{M} - M^*\|_{\F} = o(n)$. Thus, with respect to the noisy problem we are studying, in which $\|E\| = \Omega(n)$, existing results are insufficient. In contrast, our algorithm not only  improves upon the recovery of $M^*$ to sufficiently allow for recovery of $A$, it also provides an additional guarantee on entrywise recovery: $\|\hat{M}-M^{*}\|_{\max}$. All our guarantees are min-max optimal, and beyond the recovery of $M^{*}$, to the best of our knowledge, we are also the first paper to analyze the matrix anomaly detection as a formal cost minimization problem.

\begin{table}
\centering
\begin{tabular}{@{}ccc@{}}
\toprule
&$\norm{\hat{M}-M^{*}}_{\F}$ & $\norm{\hat{M}-M^{*}}_{\max}$ \\
\midrule
\cite{zhou2010stable} & $n\norm{E}_{\F}$ & -- \\

\cite{wong2017matrix} & $\sqrt{n}\norm{E}_{\F}$ & -- \\
\cite{klopp2017robust} & $\sqrt{\log n}\norm{E}_{\F} $ & -- \\
This paper & $ \frac{\sqrt{\log n}}{\sqrt{n}} \norm{E}_{\F}$ & $\frac{\sqrt{\log n}}{n\sqrt{n}} \norm{E}_{\F}$ \\
\bottomrule
\end{tabular}

%

\caption{Comparison of our results with existing work under proper hyper-parameters. The reported quantities are the scalings of upper bounds on the error of $\|\hat{M} - M^*\|$, for two matrix norms, with respect to matrix size $n$ and noise $E := Y - M^*$.} 
\label{table:comparison}
\end{table}

Finally, our work contributes to the area of statistical inference in matrix completion. This stream \citep{abbe2017entrywise,CFMY:19,ma2019implicit} has recently produced tight statistical characterizations of various algorithms for random matrices. Our own algorithm necessitates proving a similar result, namely, the first entry-wise guarantee for sub-exponential (rather than sub-Gaussian) noise. Our proof of this result builds on techniques from \cite{abbe2017entrywise}, and also draws on a recent result from Poisson matrix completion \citep{mcrae2019low}.


\vspace{1em}
\noindent
\textsf{\textbf{Notation:}}
For matrix $A \in \R^{n\times m}$, we abbreviate $\sum_{(i,j)\in [n]\times [m]} A_{ij}$ as $\sum_{ij} A_{ij}$ when no ambiguity exists. We will require a few matrix norms: $\norm{A}_{2,\infty} := \max_{i}\sqrt{\sum_{j} A_{ij}^2}$, $\norm{A}_{\max} = \max_{ij} |A_{ij}|, \norm{A}_{\F} = \sqrt{\sum_{ij} A_{ij}^2}$, and the spectral norm of $A$ is denoted $\norm{A}_2.$ The letter $C$ (and $c$) represents a sufficiently large (and small) universal (i.e. not dependent on problem parameters) constant that may change between equations.

\vspace{1em}

\section{Model and Main Results} \label{sec:model}
The following is the core problem which we will study. There exists an {\em expected demand} matrix $M^* \in \mathbb{R}_{+}^{n \times m}$, whose $n$ rows correspond to individual stores, whose $m$ columns correspond to individual SKUs, and whose entries are non-negative. Without loss of generality, we will assume that $n \geq m$.\footnote{This assumption is without loss of generality because we can flip the store and SKU axes.} Let $r$ denote the matrix {\em rank} of $M^*$, which loosely speaking, controls the extent to which the expected demands are related across stores and SKUs. 
For example, the most restrictive case ($r=1$) is equivalent to the following structural equation: \[ M^*_{ij} = u_i v_j, \]
where $u_i$ and $v_j$ can be interpreted as fixed effects, respectively, for store $i$ and SKU $j$.
At the other extreme, the least restrictive case ($r = n$) corresponds to placing no restriction at all on $M^*$. All other cases interpolate between these two (precisely, $r$ corresponds to the dimensionality of the fixed effects).
A second {\em anomaly} matrix $B \in \{0,1\}^{n \times m}$, of the same dimensions as $M^*$, contains binary elements which serve to indicate the store-SKU pairs for which an anomalous event (such as phantom inventory) has occurred.

Given $M^*$ and $B$, a random {\em sales} matrix $X$ is generated with independent entries distributed according to\footnote{We focus here on a model with non-negative, integer-valued $X$ that fits inventory applications. Our results can easily be extended to more general sub-exponential noise, as we discuss in \cref{sec:AlgorithmDetails}.}
\begin{align*} X_{ij} \sim
 \begin{cases}  \mathrm{Poisson}(M_{ij}^{*}) & \text{if } B_{ij} = 0 \\
 \mathrm{Anom}(\alpha^{*}, M_{ij}^{*}) & \text{if } B_{ij} = 1.
\end{cases}
\end{align*}
In words, if an anomaly has {\em not} occurred for a particular store-SKU pair ($B_{ij} = 0$), then the demand is drawn according to the `usual' Poisson distribution with mean $M_{ij}^*$. If an anomaly {\em has} occurred, the demand is instead drawn according to an alternate distribution $\mathrm{Anom}(\cdot,\cdot)$, which is some non-negative, integer-valued random variable parameterized by $M^*_{ij}$ and an unknown parameter vector $\alpha^{*} \in \mathbb{R}^d$. This model for the effect of an anomaly is flexible enough to model many different types of real-world anomalous events -- one simple example is phantom inventory events, for which it is natural to use $\mathrm{Anom}(\alpha, M_{ij}^{*}) = \mathrm{Poisson}(\alpha M_{ij}^{*})$, meaning the original Poisson sales process terminates some fraction $\alpha \in (0,1)$ of the way through the measured time horizon \citep{dehoratius2008retail}.

None of $M^*$, $B$, or $\alpha^*$ are known in advance. Instead, we observe only $X$, and only on a subset of store-SKU pairs $\Omega \subset [n]\times[m]$. We denote this observation by $X_\Omega$.
The subset of entries $\Omega$ is observed independently with probability $p_{\O}$. In addition, we assume that the entries of $B$ are independent ${\rm Bernoulli}(p_{\A}^{*})$ variables, where $p_{\A}^{*}$ is bounded away from one by a constant (so that at least a constant fraction of store-SKU pairs are not anomalous).\footnote{For readability, we have taken care to ensure that unknown quantities are denoted with asterisks (*). The quantity $p_{\O}$ is an exception -- it is unknown, but it is also so trivially estimated (from $\Omega$) that we omit the asterisk.} The fact that the positions of the observed entries ($\Omega$) and the anomalies ($B$) are uniformly distributed across the matrix is a seemingly-restrictive assumption that is worth addressing immediately:
\begin{itemize}
	 \item This assumption, often referred to as the `random uniform model', is \textit{canonical} in the matrix completion and matrix anomaly detection literature, e.g.~since the seminal work of \cite{candes2011robust}.
	 \item While the assumption itself enables a cleaner theoretical analysis of our algorithm, the algorithm can still be used when the assumption does not hold. In fact, we observe that our algorithm is robust to non-uniform observations and anomalies. One example of such is the real-data experiment in \cref{sec:real-data}, where both $\Omega$ and $B$ are highly correlated.\footnote{The empirical success of matrix-completion-type methods for non-uniform patterns can be traced back to the Netflix prize \citep{bennett2007netflix}.}
	 \item Recent progress for deterministic matrix regression problems \citep{chatterjee2020deterministic,farias2021learning} suggests that this robustness to non-uniform patterns is theoretically justified.
\end{itemize}

Given the above model, our goal is to minimize a certain cost function, to be described shortly, which will rely on inferring $B$ from $X_\Omega$.
Before proceeding, we will state and discuss the assumptions we place on both $M^*$ and the anomaly distribution. First, on $M^*$, we make the following assumptions, which are by this point standard in the matrix completion literature (e.g.~\cite{abbe2017entrywise,ma2019implicit}):


\vspace{1em}
\noindent
\textsf{\textbf{Assumptions on $M^*$:}} Let $M^{*} = U \Sigma V^{T}$ be its singular value decomposition (SVD), where $\Sigma \in \R^{r\times r}$ is a diagonal matrix with singular values $\sigma_1^{*} \geq \sigma_2^{*} \geq \dotsc \geq \sigma_{r}^{*}$ (let $\kappa := \sigma_1^{*}/\sigma_{r}^{*}$), and $U \in \R^{n \times r}, V \in \R^{m \times r}$ are matrices whose columns are the left and right-singular vectors of $M^*$.
\begin{itemize}
    \item(Boundedness): $\norm{M^{*}}_{\max}$ is bounded away from 0 by a constant, and \[\norm{M^{*}}_{\max} \leq L.\]
    \item (Incoherence):  \[\norm{U}_{2,\infty} \leq \sqrt{\frac{\mu r}{n}}, \norm{V}_{2,\infty} \leq \sqrt{\frac{\mu r}{m}}\]
    \item (Sparsity): \[\sqrt{p_\O}  \geq C\frac{\log^{1.5}(m) \mu r \kappa^2}{\sqrt{m}}\] for some constant $C$.   
\end{itemize}
These assumptions include parameters ($\mu,L,\kappa$), all of which typically scale as $O(1)$ with respect to $n$, though we do not assume this scaling explicitly (instead, our theoretical guarantees depend on them). The first two assumptions together are meant to preclude the possibility that a single (or just a few) store-SKU pair's demand makes up an overwhelming proportion of that store's, or that SKU's, total demand. The third assumption ensures we observe sufficiently many entries of the sales matrix $X$.

Second, recall that we assume a probabilistic anomaly model ${\rm Anom}(\cdot, \cdot)$ parameterized by a finite number of unknown parameters $\alpha^* \in \mathbb{R}^d$, or more generally $\alpha^* \in \Gamma$ for some known $\Gamma \subset \mathbb{R}^d$.
We will make the following assumptions on ${\rm Anom}(\cdot, \cdot)$:
\vspace{1em}

\noindent
\textsf{\textbf{Assumptions on ${\rm Anom}(\cdot, \cdot)$:}} \begin{itemize}
    \item(Sub-exponential): $\mathrm{Anom}(\alpha^{*},M_{ij}^{*})$ is sub-exponential:\footnote{The {\em sub-exponential norm} of a random variable $X$ is defined as $ \norm{X}_{\psi_1} := \inf\{t > 0: \E{\exp(|X|/t)}\leq 2\}.$ For $X$ itself to be {\em sub-exponential} is equivalent to having finite sub-exponential norm. The set of sub-exponential distributions includes all sub-Gaussian distributions, along with others such as the Poisson. We also use the same $L$ as in the assumption for $\|M^{*}\|_{\max}$ for convenience. One can simply take $L := \max\{\|{\mathrm{Anom}(\alpha^{*},M_{ij}^{*})}\|_{\psi_1}, \|M^{*}\|_{\max}\}.$} \[\|{\mathrm{Anom}(\alpha^{*},M_{ij}^{*})}\|_{\psi_1} \leq L.\]
    \item(Smoothly-parameterized): For each $k \in \mathbb{N}$, the quantity $\Prob{ \mathrm{Anom}(\alpha, M ) = k }$, viewed as a real-valued function on $(\alpha,M)$, is $K$-Lipschitz.
    \item(Mean Decomposition): For any $M \in \R_{+}$ and $\alpha \in \Gamma$, we have that \begin{equation}\E{\mathrm{Anom}(\alpha,M)} = g(\alpha)M \label{eq:def-g}\end{equation} for some $g: \R^d \rightarrow [0,1]$, where $g(\alpha)$ is $K$-Lipschitz in $\alpha.$
  \end{itemize}

It is worth pausing to discuss these assumptions on ${\rm Anom}(\cdot, \cdot)$ (incidentally, our example anomaly model for phantom inventory, $\mathrm{Anom}(\alpha, M_{ij}^{*}) = \mathrm{Poisson}(\alpha M_{ij}^{*})$, does satisfy all three assumptions, and is a useful example to hold in mind while parsing through them).
First, sub-exponential distributions are actually {\em more} general than the usually assumed sub-Gaussian distributions. This is an integral component of this work, as existing results do not apply to certain useful distributions that are sub-exponential, but not sub-Gaussian, such as the Poisson.

The second and third assumptions together enable identification of $\alpha^*$. The second assumption is, loosely speaking, necessary (there must be {\em some} requirement that the parameterization of ${\rm Anom}(\cdot, \cdot)$ be meaningful).
The third assumption is new and is another critical component of this work. We will discuss this more in the following section, but for now, consider alternative approaches:
\begin{itemize}
\item The matrix anomaly detection literature by and large makes a {\em zero-mean} assumption, i.e.~that the net additive perturbation of all anomalies is zero $\E{\mathrm{Anom}(\alpha, M)} = 0$. This is {\em more} restrictive than our assumption, and would not apply to our inventory applications.
\item In contrast to this probabilistic model, one could consider an {\em adversarial} anomaly model. 
However, the adversarial model that allows for arbitrary anomalies will essentially either require \textit{noiseless} observations or incur \textit{non-identification} problems \citep{candes2011robust,zhang2018robust,chen2020bridging}. Hence, certain probabilistic assumptions for the anomalies are required to make the problem meaningful.
\end{itemize}

\vspace{1em}
\subsection{The Cost Function}
We will view any anomaly detection {\em algorithm} $\pi$ as a mapping from the observed sales $X_\Omega$ to a binary matrix $A^\pi \in \{0,1\}^{n \times m}$, which encodes the store-SKU pairs that algorithm $\pi$ identifies to be anomalous. In particular, $A^{\pi}_{ij} = 1$ indicates that the algorithm predicts that $B_{ij} = 1$, i.e.~an anomaly has occurred at entry $(i, j).$

The final component in fully stating our problem is to define the performance metric with which we will evaluate any algorithm. Now existing results are, for the most part, stated in terms of the number of errors, that is entries for which $A_{ij}^\pi \ne B_{ij}$. However, as discussed in the previous section, there are in reality two different types of errors, with potentially different costs. Thus, the metric we will use (and seek to minimize) is total {\em cost}, where the cost incurred at each entry $(i,j) \in \Omega$ depends on the type of error:
\begin{itemize}
	\item $c_{ij}^{(0)}$ is incurred if $A_{ij}^\pi = 1$ and $B_{ij} = 0$. This {\em false positive} cost might represent the labor-time wasted by sending an employee to verify and correct a (non-existent) inventory anomaly.
	\item $c_{ij}^{(1)}$ is incurred if $A_{ij}^\pi = 0$ and $B_{ij} = 1$. This {\em false negative} cost might represent lost sales from an inventory anomaly, minus the labor-time that would have been spent on sending an employee to verify and correct the inventory anomaly.
\end{itemize}
Both costs are assumed to be non-negative. 
Furthermore, the costs can be {\em heterogeneous} across stores and SKUs, e.g. because some SKUs are more sensitive to inconsistent inventory records than others, or different stores have different labor costs.

Conditioned on the observation $X_{\Omega}$,\footnote{When `conditioning' on $X_{\Omega}$, we are referring to the probability distribution over $X_{\Omega}$ and $B$ that is fully specified by $(M^*,\alpha^*,p_A^*)$ and $p_O$.} our performance metric is the expected average cost for an algorithm $\pi$, as given by
\begin{align*}
\mathrm{cost}^{\pi}(X_{\Omega}) := \frac{1}{|\Omega|}\E{\sum_{(i,j) \in \Omega} c_{ij} \Big| X_{\Omega}},
\end{align*}
where $c_{ij}$ is the cost incurred at entry $(i,j)$\footnote{Our model can also allow more general costs (or rewards) incurred  when $A_{ij}^{\pi}=1,B_{ij}=1$ or $A_{ij}^{\pi}=0, B_{ij}=0$. See Appendix \ref{sec:general-cost} for more discussion.}:
\begin{align*}
c_{ij} &:=   c_{ij}^{(1)} \1{A_{ij}^{\pi}=1, B_{ij} = 0}
 + c_{ij}^{(0)} \1{A_{ij}^{\pi}=0, B_{ij} = 1}.
\end{align*}

\vspace{1em}

\subsection{Our Main Results}
In the next section, we will outline our algorithm $\pi^{\EW}$, which we refer to as the {\em entrywise (EW)} algorithm. Before doing so, we will state our main result, which is an upper bound on the cost of $\pi^{\EW}$. We will specifically measure cost relative to $\pi^*$, which denotes the {\em optimal} algorithm\footnote{The fact that there even exists a well-defined `optimal' algorithm may not be obvious a priori. We will define $\pi^*$ explicitly in the next section.} if the expected demand $M^*$, the anomaly model parameter $\alpha^*$, and the anomaly likelihood $p_A^*$ are known exactly (in contrast, our algorithm $\pi^{\EW}$ knows none of these in advance). Note that having $(M^*,\alpha^*,p_A^*)$ is {\em not} sufficient to back out $B$ exactly, so $\pi^*$ incurs a non-zero cost.

Our main result is the following, which guarantees that the cost of the EW algorithm is within an additive factor of the cost of $\pi^{*}$, which vanishes with increasing $m$ (recall that $m$ is the smaller of the two matrix dimensions):

\begin{theorem}\label{thm:main-theorem}
The expected cost of the entrywise algorithm $\pi^{\mathrm{EW}}$ satisfies
\begin{align*}
\Ex{X_{\Omega}}{\mathrm{cost}^{\pi^{\mathrm{EW}}}(X_\Omega) - \mathrm{cost}^{\pi^{*}}(X_\Omega)} &= O\left(\frac{\log^{1.5}(m)}{\sqrt{m}}\right).
\end{align*}
\end{theorem}
Here, $O(\cdot)$ hides polynomial dependence on $K, L, \kappa, \mu, p_{\O}^{-1}$ and $r$, which typically scale as constant with respect to $m$ in applications. As a technical aside, \cref{thm:main-theorem} is stated in terms of the expected cost (with the expectation taken over the different random realizations of observed data $X_\Omega$), but we in fact prove a stronger guarantee that the same bound holds with probability $1-O(1/nm)$. The proof of \cref{thm:main-theorem} is sketched in \cref{sec:proof-sketches}.

A few comments on the consequences of \cref{thm:main-theorem}:
\begin{enumerate}
\item \cref{thm:main-theorem} makes explicit the value of cross-sectional data: as the number of stores and SKUs grow, the cost of our algorithm approaches the lowest cost we could incur had we known the expected demand and underlying anomaly model. Recall that the longitudinal approach relies on knowing (by learning over time) exactly these pieces of information, and so $\pi^*$ can be viewed as applying the longitudinal approach to each store-SKU pair separately, with data collected over a long period of time, in an idealized time-homogeneous environment. In short, \cref{thm:main-theorem} guarantees that our algorithm achieves nearly that same idealized cost, in a potentially time-heterogeneous environment, using a single cross-sectional `snapshot' of data.
\item Because the guarantee in \cref{thm:main-theorem} is additive, it would be meaningless if hypothetically the optimal cost, $\mathrm{cost}^{\pi^{*}}(X_\Omega)$, were itself vanishing quickly with $m$. This is {\em not} the case. Again, even knowing $(M^*,\alpha^*,p_A^*)$, there is a constant probability of misidentifying anomalies (we will make this explicit in the next section), and so in fact $\mathrm{cost}^{\pi^{*}}(X_\Omega) = \Omega(1)$.
\item The rate in \cref{thm:main-theorem} is in fact minimax optimal, up to a logarithmic factor. This is captured by our second main result, the following proposition:
\end{enumerate}
\begin{proposition}\label{prop:lower-bound}
For any $m \in \mathbb{N}$ and algorithm $\pi$, there exists $M^* \in \mathbb{R}^{m \times m}$ and an anomaly model, with $K,L,\kappa,\mu,r=O(1)$ and $p_{\O},p_{\A}^*=\Omega(1)$, such that the following holds:
\begin{align*}
    \Ex{X_{\Omega}}{\mathrm{cost}^{\pi}(X_{\Omega}) - \mathrm{cost}^{\pi^*}(X_{\Omega})} = \Omega\left(\frac{1}{\sqrt{m}}\right).
\end{align*}
\end{proposition}
The proof of \cref{prop:lower-bound-rate}, which can be found in \cref{sec:lower-bound-appendix}, is by an explicit construction of a randomized family of instances.

\vspace{1em}

\section{Our Entrywise (EW) Algorithm}\label{sec:algorithm}
We are now prepared to state our algorithm for the anomaly detection problem formulated in the previous section. To understand the approach, it is worth first working out the `optimal' algorithm $\pi^*$ alluded to above, i.e.~the optimal approach assuming that $(M^*,\alpha^*,p_A^*)$ were known. 

\vspace{1em}
\subsection{Preliminaries: Characterizing the Optimal Algorithm}

The key observation (which we show in the following paragraph) is that for any entry $(i,j)\in\Omega$, the optimal decision of whether or not to identify the entry as anomalous is completely characterized by the quantity
\begin{equation} 
f_{ij}^{*} := \Prob{B_{ij} = 0~|~X_{\Omega}}, \label{eq:fij}
\end{equation}
i.e.~the likelihood that entry $(i,j)$ is not anomalous given observation $X_{\Omega}$. 
This quantity can be calculated explicitly with $(M_{ij}^*,\alpha^*,p_A^*)$. We will show this calculation soon, but it is simply a statement of Bayes' Theorem. The precise optimal decision, given $f_{ij}^*$, is then a threshold rule:
\begin{align} A_{ij}^{\pi^*} =  \1{ \frac{c_{ij}^{(1)}}{c_{ij}^{(0)} + c_{ij}^{(1)}}  \geq f_{ij}^*}.   	\label{eq:decision-rule}
\end{align}
For intuition on \cref{eq:decision-rule}, consider an extreme case: if the cost of a false negative (meaning $B_{ij} = 0$, but $A_{ij}=1$) is relatively high (meaning $c_{ij}^{(1)} \gg c_{ij}^{(0)}$), then $A_{ij}^{\pi^*} = 1$, intuitively to avoid incurring this high false negative cost. The reverse holds at the other extreme: a relatively high false positive cost implies that $A_{ij}^{\pi^*} = 0$. In between those extremes, the optimal decision is dictated by the ratio of the two costs, along with $f_{ij}^*$: if the likelihood of an anomaly is sufficiently high (meaning $f_{ij}^*$ small), then $A_{ij}^{\pi^*} = 1$, as we would expect.

Finally, to show that \cref{eq:decision-rule} is indeed optimal, we can re-write the cost function as follows: 
\begin{align}
\mathrm{cost}^{\pi}(X_{\Omega}) &= \frac{1}{|\Omega|}\sum_{(i,j) \in \Omega} \Big\{   c_{ij}^{(0)} \Prob{A_{ij}^{\pi}=1, B_{ij} = 0 \Big| X_{\Omega}}  + c_{ij}^{(1)} \Prob{A_{ij}^{\pi}=0, B_{ij} = 1 \Big| X_{\Omega}}   \Big\} \label{eq:cost-pi-pre} \\
&= \frac{1}{|\Omega|}\sum_{(i,j) \in \Omega} \Big\{   c_{ij}^{(0)} \Prob{A_{ij}^{\pi}=1 \Big| X_{\Omega}} f_{ij}^*  + c_{ij}^{(1)} \Prob{A_{ij}^{\pi}=0 \Big| X_{\Omega}}(1-f_{ij}^*)   \Big\} \nonumber \\
&= \frac{1}{|\Omega|}\sum_{(i,j) \in \Omega} \Big\{   c_{ij}^{(0)}  f_{ij}^*  + \left(c_{ij}^{(1)} - (c_{ij}^{(0)}+c_{ij}^{(1)})f_{ij}^* \right) \Prob{A_{ij}^{\pi}=0 \Big| X_{\Omega}}   \Big\}. \nonumber
\end{align} 
The first line above is by linearity of expectations. The second line is by conditional independence between $A_{ij}^{\pi}$ and $B_{ij}$ with $X_{\Omega}$ given (since $\pi$ solely depends on $X_{\Omega}$). The third line follows from a consolidation of terms, and reveals the threshold rule in \cref{eq:decision-rule}.

\vspace{1em}
\subsection{Algorithm Overview}
The previous discussion makes clear that given $(M^*,\alpha^*,p_A^*)$, the optimal algorithm is a threshold rule, as defined in \cref{eq:decision-rule}, based on the quantity $f_{ij}^{*},$ as defined in \cref{eq:fij}.
Naturally, our algorithm, which we refer to as the {\em entrywise (EW)} algorithm, approximates $\pi^{*}$ by estimating $f_{ij}^{*}$. It is so-named because it leverages an entrywise matrix completion guarantee for sub-exponential noise that we will describe shortly. The crux of our algorithm is stated in \cref{alg:EW}, with certain details left to be defined in the next subsection.
\begin{algorithm}[h!]
\caption{Entrywise (EW) Algorithm $\pi^{\mathrm{EW}}$ } \label{alg:EW}
{\bf Input:} $X_\Omega$, $\gamma \in (0,1]$
\begin{algorithmic}[1]
\State{Compute \[\hat{M} = \frac{nm}{|\Omega|} \SVD(X_{\Omega})_{r},\] where $ \SVD(X_{\Omega})_{r} := \arg\min_{\rank(M)\leq r} \norm{M-X'}_{\F}$, and $X'$ is obtained from $X_{\Omega}$ by setting entries outside of $\Omega$ to 0.}    
\State{Estimate $(\hat{\alpha},\hat{p}_{\A})$, e.g.~using the moment matching estimator in \cref{eq:optimize-F}. } 
\State{Estimate $\hat{f}_{ij}$ for $f_{ij}^{*}$, for $(i, j) \in \Omega$ using the plug-in estimator in \cref{eq:plug-in-f}. }
\State{For every $(i, j) \in \Omega$, set $A_{ij} = 1$ if 
$$
\frac{c_{ij}^{(1)}}{c_{ij}^{(0)} + c_{ij}^{(1)}}  \geq \hat{f}_{ij}.
$$
Otherwise, set $A_{ij}=0.$
}
\end{algorithmic}
{\bf Output:} $A_\Omega$
\end{algorithm}

On first read, it is perhaps easiest to parse through \cref{alg:EW} in {\em backwards} order. The last step, Step 4, mimics the optimal decision rule in \cref{eq:decision-rule}, but uses estimates $\hat{f}_{ij}$ of each $f_{ij}^*$. These estimates are computed in Step 3 by taking the expression for $f_{ij}^*$ (see \cref{eq:fij-star} in the next subsection), which depends on $(M_{ij}^*,\alpha^*,p_A^*)$, and `plugging-in' estimates $(\hat{M}_{ij},\hat{\alpha},\hat{p}_A)$ for these quantities. One critical point to note here is that since Steps 3 and 4 are performed separately at each entry, each estimate $\hat{M}_{ij}$ of $M^*_{ij}$ must be sufficiently accurate: hence the need for an {\em entrywise} guarantee.

Steps 1 and 2 together produce these estimates $(\hat{M},\hat{\alpha},\hat{p}_A)$. Step 1 constructs a `de-noised' estimate $\hat{M}$ of $M^*$ via a simple SVD-based matrix completion algorithm. The fact that this relatively simple algorithm works here  is a substantial advantage in terms of scalability, and is not obvious in the context of the matrix completion results preceding our work. In particular, previous guarantees either (a) bound global (but not necessarily entrywise) errors, or (b) are specific to sub-Gaussian (but not necessarily sub-exponential) distributions. This is our main technical contribution (\cref{thm:entrywise-bound}). 

Finally, $\hat{M}$ is in fact not exactly an estimate of $M^*$, but rather a {\em linear scaling} of $M^*$ that depends on $\alpha^*$ and $p_A^*$. Here the entrywise guarantee is again vital, as it enables us in Step 2 to produce accurate estimates $\hat{\alpha}$ and $\hat{p}_A$ using any `standard' parametric estimator (we will specify a concrete estimator in \cref{eq:optimize-F}, but this can largely be viewed as a black box), which then allows us to `undo' the linear scaling. We will describe the entire algorithm in greater detail in the next subsection.

\vspace{1em}
\subsection{Algorithm Details}\label{sec:AlgorithmDetails}
We conclude this section by `filling in' the details of the description of our algorithm.

\vspace{1em}
\noindent
\textsf{\textbf{Step 1: De-noising with an Entrywise Guarantee: }}
Our algorithm is initiated with an SVD-based de-noising of $X_\Omega$: 
\[\hat{M} = \frac{nm}{|\Omega|} \SVD(X_{\Omega})_{r},\] 
where $\SVD(X_{\Omega})_{r} := \arg\min_{\rank(M)\leq r} \norm{M-X'}_{\F}$,\footnote{While phrased here as an optimization problem, the $\SVD(\cdot)_{r}$ of a matrix would in practice be computed by calculating its top $r$ singular vectors and values, a highly-efficient computation.} and $X'$ is obtained from $X_{\Omega}$ by setting entries outside of $\Omega$ to 0. 
As mentioned previously, $\hat{M}$ is not anticipated to be close to $M^*$, but rather a linear scaling of $M^*$ that depends on $\alpha^*$ and $p_A^*$. Indeed, a quick quick calculation shows that 
\[ \E{X_{ij}} = \left(p_A^* g(\alpha^*) + (1-p_A^*)\right) M_{ij}^*,  \]
where $g(\cdot)$ is as defined in \cref{eq:def-g}.
To ease notation, let $\theta = (\alpha,p_A)$ and $\theta^{*} = (\alpha^{*},p_{\A}^{*})$, so that $\theta,\theta^* \in \Theta := \Gamma \times [0,1)$,\footnote{Recall that $\alpha^* \in \Gamma \subset \mathbb{R}^d$, and $p_A^*$ is assumed to be bounded away from one by a constant.} and let $e(\theta)$ denote the linear scaling, i.e.~$e(\theta) := p_{\A}g(\alpha)+(1-p_{\A})$. As a sanity check, $\E{X} = e(\theta^*)M^*$.

While the SVD-based de-noising algorithm used here is standard, the key result that drives the rest of the algorithm (and its analysis) is the following new {\em entrywise} error bound, which is likely to be of independent interest:
\begin{theorem}\label{thm:entrywise-bound}
With probability $1-O(\frac{1}{nm})$, \[\norm{\hat{M} - e(\theta^{*})M^{*}}_{\max} \leq C\kappa^4 \mu r L \sqrt{\frac{\log(m)}{p_{\O}m}}.\]
\end{theorem}
\cref{thm:entrywise-bound} can be viewed as the first entrywise guarantee result for Poisson matrix completion (in fact, the proof also holds valid for sub-exponential noise).  A proof sketch is provided in \cref{sec:proof-sketches}. As a comparison, consider the  recent results for aggregated error  on matrix completion with Poisson noise \citep{mcrae2019low}. Under carefully-selected hyperparameters, their results based on the SVD provide the following Frobenius norm bound: $\|\hat{M} - M^{*}\|_{\F} \lesssim m^{1/2}.$ In contrast, our entrywise guarantee states that $\|{\hat{M} - M^{*}}\|_{\max} \lesssim m^{-1/2} \log^{1/2}(m).$
Therefore, our results show that the SVD approach not only enjoys an aggregated error guarantee, but in fact the entrywise error is evenly distributed among all entries up to a logarithmic factor. 

\vspace{1em}
\noindent
\textsf{\textbf{Step 2: Recovering $(\alpha^*,p_A^*)$ with a Moment Matching Estimator: }}
Step 1 yields an (entrywise) accurate estimator $\hat{M}$ of $e(\theta^*)M^*$, where $\theta^* = (\alpha^*,p_A^*)$. Now in Step 2, we use $\hat{M}$ to accurately estimate $\theta^*$, and therefore $M^*$ itself. Let $\hat{\theta}$ denote our estimator for $\theta^*$. The accuracy we will require on $\hat{\theta}$ (in order for \cref{thm:main-theorem} to hold) is the following:
\begin{equation} \label{eq:step-2-accuracy}
\norm{\hat{\theta} - \theta^{*}} \leq  C (K+L)\kappa^4 \mu r L \sqrt{\frac{\log(m)}{p_{\O}m}}.
\end{equation}
Now given the guarantee on $\hat{M}$ in \cref{thm:entrywise-bound}, there are a variety of `standard' estimators that would suffice, e.g.~a maximum likelihood estimator would be quite natural. For concreteness, here we will specify one estimator that satisfies \cref{eq:step-2-accuracy}, in the case of discrete distributions (such as the Poisson).

Our estimator works by `matching' generalized moments of the cumulative distribution function, at sufficiently many values for identifiability. Specifically, for any matrix $M$ and any $\theta$, consider the observation model fully specified by $M$ and $\theta$ (and technically $p_{\O}$ as well), and let $g_t(\theta,M)$ denote the proportion of entries of $X_\Omega$ expected to be at most $t$: 
$$
g_t(\theta, M) := \frac{\E{|X_{ij}\leq t, (i,j)\in \Omega|}}{\E{|\Omega|}}.
$$ 
These values, namely $g_t(\theta, M)$ for various $t \in \mathbb{N}$, are the generalized moments we are referring to.
Given that $M^{*} \approx \hat{M}/e(\theta^{*})$ by \cref{thm:entrywise-bound}, we choose $\hat{\theta}$ to be the minimizer of the following function which seeks to match a set of $T$ empirical moments to their expectations as closely as possible (in $\ell^2$ distance):
\begin{align} 
    \hat{\theta} := \argmin_{\theta \in \Theta}\sum_{t=0}^{T-1} \left(g_t(\theta, \hat{M}/e(\theta)) - \frac{|X_{ij}\leq t, (i,j)\in \Omega|}{|\Omega|}\right)^2, \label{eq:optimize-F}
\end{align}
where $T$ is a large enough constant for identifiability ($T=d+1$ typically suffices).

This estimator satisfies \cref{eq:step-2-accuracy}. To state this formally, let $F = (F_0, F_1, \dotsc, F_{T-1}): \Theta \rightarrow \R^{T}$ be defined as $F_t(\theta) = g_t(\theta, M^{*}e(\theta^{*})/e(\theta))$, and let \[\delta' = \kappa^4 \mu r L \sqrt{\frac{\log m}{p_{\O}m}} \] be the entrywise bound on $\|{\hat{M}-e(\theta^{*})M^{*}}\|_{\max}.$ 
\begin{lemma} \label{lem:step2}
Assume the following regularity conditions on $F(\theta)$:
\begin{itemize}
\item $F: \Theta \rightarrow \R^{T}$ is  continuously differentiable and injective.
\item Let $\tilde{\delta} =\delta'(K+L)\log m$. We require $B_{\tilde{\delta}}(\theta^{*}) \subset \Theta$, where $B_{r}(\theta^{*}) = \{\theta: \norm{\theta^{*} - \theta} \leq r\}.$
\item For any $\theta \in B_{\tilde{\delta}}(\theta^{*})$, $\norm{J_{F}(\theta) - J_{F}(\theta^{*})}_{2} \leq \frac{C}{\tilde{\delta}} \norm{\theta-\theta^{*}}$, where $J$ is the Jacobian matrix.
\item $\norm{J_{F}(\theta^{*})^{-1}}_2 \leq C$.	
\end{itemize}
Then with probability $1-O(\frac{1}{nm})$, \cref{eq:step-2-accuracy} holds.
\end{lemma}
The regularity conditions in \cref{lem:step2} are among the typical set of conditions for methods involving generalized moments and are well justified in typical applications \citep{newey1994large,imbens1995information,hall2005generalized,hansen1982large}. The net of this is that our moment matching estimator is able to accurately estimate $\theta^*.$

\noindent
{\em{(Aside): Extending to continuous noise models:}} To extend \cref{alg:EW} to general (possibly continuous) sub-exponential noise, the same steps work, except that the estimator for $\hat{\theta}$ in Step 2 needs to be changed. For observation $X$ with continuous values, one can use MLE estimator to solve $\hat{\theta} = \arg\max_{\theta} \Prob{X|\theta, M^{*}}$ by plugging in $M^{*}\approx \hat{M}/e(\theta)$. 

\vspace{1em}
\noindent
\textsf{\textbf{Step 3: A Plug-in Estimator for $f_{ij}^{*}$: }} \label{sec:plug-in-f}
Recall that $f_{ij}^*$, as defined in \cref{eq:fij-star}, is the key quantity used by the `optimal' algorithm. Although $f_{ij}^{*}$ is not known, we can obtain an estimate $\hat{f}_{ij}$ based on the estimates $(\hat{M}, \hat{\alpha}, \hat{p}_{\A})$ derived in the previous steps. Specifically, by independence across entries, for $(i, j) \in \Omega$, we have 
\begin{align}
f_{ij}^{*} 
&:= \Prob{B_{ij} = 0~|~X_{\Omega}} \nonumber \\
&= \Prob{B_{ij} = 0~|~X_{ij}} \nonumber\\
&\overset{(i)}{=} \frac{\Prob{X_{ij}~|~B_{ij}=0} \Prob{B_{ij}=0}}{ \sum_{k=0,1} \Prob{X_{ij}~|~B_{ij}=k}\Prob{B_{ij}=k}} \nonumber\\
&= \frac{(1-p_\A^{*}) \Probx{\Pos(M_{ij}^*)}{X_{ij}}}{p_\A^{*} \Probx{\Ano(\alpha^*,M_{ij}^*)}{X_{ij}} + (1-p_\A^{*}) \Probx{\Pos(M_{ij}^*)}{X_{ij}}},\label{eq:fij-star}
\end{align}
where (i) is due to Bayes' Theorem. Thus, we can re-write $f_{ij}^{*}$ as $f_{ij}^* = y_{ij}^{*}/(x_{ij}^{*} + y_{ij}^{*})$ by defining $x_{ij}^{*}$ and $y_{ij}^{*}$ as follows:  
\begin{align*}
x_{ij}^{*} &:= p_\A^{*} \Probx{\Ano(\alpha^*,M_{ij}^*)}{X_{ij}} \\
y_{ij}^{*} &:=  (1-p_\A^{*}) \Probx{\Pos(M_{ij}^*)}{X_{ij}}.
\end{align*}

We estimate $x_{ij}^{*}$ and $y_{ij}^{*}$ via direct plug in: 
\begin{align*}
\hat{x}_{ij} &:= [\hat{p}_\A \mathbb{P}_{\Ano}(X_{ij}|\hat{\alpha}, \hat{M}_{ij}/e(\hat{\theta}))],\\
\hat{y}_{ij} &:= [(1-\hat{p}_\A)\mathbb{P}_{\Pos}(X_{ij}|\hat{M}_{ij}/e(\hat{\theta}))],	
\end{align*}
where $[x]$ denotes $x$ `truncated' to its nearest value in $[0,1]$, i.e.~$[x]=\max(\min(x,1),0)$. Naturally, we then estimate $f^*_{ij}$ as 
\begin{align} \label{eq:plug-in-f}
\hat{f}_{ij} := \frac{\hat{y}_{ij}}{\hat{x}_{ij} + \hat{y}_{ij}}.
\end{align}

\vspace{1em}
\noindent
\textsf{\textbf{Step 4: Near-optimal Decision Rule: }}
The final step is exactly as stated previously. The optimal decision rule is applied with $\hat{f}_{ij}$ as proxy for $f^*_{ij}$:
\begin{align*} A_{ij}^{\pi^{\mathrm{EW}}} =  \1{ \frac{c_{ij}^{(1)}}{c_{ij}^{(0)} + c_{ij}^{(1)}}  \geq \hat{f}_{ij}}.   
\end{align*}

\vspace{1em}

\section{Experiments} \label{sec:experiments}

To evaluate the empirical performance of our EW algorithm, we compare it against various state-of-the-art approaches. We first consider a synthetic setting, which has the advantage of an exact ground truth, and then we  measure performance on real-world data from a large retailer. The results show that EW algorithm outperforms existing methods in both settings.

%

\vspace{1em}
\subsection{Synthetic Data} 

\textsf{\textbf{Data Generation Process:}} We generated an ensemble of matrices $M^* \in \R^{n\times m}$. The varying parameters of the ensemble include (i) $r$: the rank of the matrix; (ii) $\bar{M}^{*} = \frac{1}{nm}\sum_{ij} M_{ij}^{*}:$ the average value of all entries; (iii) $p_{\O}:$ the probability of an entry being observed; (iv)  $p_{\A}^{*}$: the probability of an entry where an anomaly occurs; and (v) $\alpha^{*}$: the anomaly parameter. When an anomaly occurs, $\E{\Ano(\alpha^{*}, M)} = \alpha^{*} M.$

The parameters were sampled uniformly: $r \in [1,10], \bar M^* \in [1,10], p_\O \in [0.5,1], p_{\A}^{*} \in [0,0.3]$ and $\alpha^{*} \in [0,1]$. Each instance was generated in the following steps: (i) Generate $M^{*}$: for a given choice of $r$ and entrywise mean $\bar M^*$, we set $M^{*} = k U V^{T}$. $U, V \in \R^{n\times r}$ are random with independent ${\rm Gamma}(1,2)$ entries and $k$ is picked so that $\bar{M}^{*} = \frac{1}{nm}\sum_{ij} M_{ij}^{*}$. This is a typical way of generating $M^{*}$ with rank $r$ and non-negative entries \citep{cemgil2008bayesian}. (ii) Observation: If $(i,j)$ is observed, then with probability $1-p_{\A}^{*}$, $X_{ij} \sim \Pos(M_{ij})$; otherwise, $X_{ij} \sim  \Pos (\mathrm{Exp}(\alpha^{*}) M_{ij})$. Here $\mathrm{Exp}(\alpha^{*})$ models the occurring time of the anomalous event. (iii) Costs: we generate each $c_{ij}^{(0)} \in [0, 10], c_{ij}^{(1)} \in [0, 10]$ uniformly. Here $c_{ij}^{(0)}$ and $c_{ij}^{(1)}$ model the heterogeneous false negative costs and false positive costs respectively. 

\textsf{\textbf{Existing Methods and Implementations:}}
For practical considerations, we implemented a slight variant of the EW algorithm where (i) the matrix completion step uses the typical soft-impute algorithm~\citep{mazumder2010spectral}, and (ii) the anomaly model estimation is done via maximum likelihood estimation.

%

We compared our EW algorithm with three existing algorithms: (i) Stable-PCP \citep{zhou2010stable,chen2020bridging}, (ii) Robust Matrix Completion (RMC) \citep{klopp2017robust}, and (iii) Direct Robust Matrix Factorization (DRMF) \citep{xiong2011direct}. These three algorithms all recover the matrices by decomposing $X$ as $X=\hat{M}+\hat{A}+\hat{E}$, and minimizing some objective $f(\hat{M})+\lambda_1 g(\hat{A})+ \lambda_2 h(\hat{E})$, where $f,g,h$ are penalty functions with Lagrange multipliers $\lambda_1,\lambda_2$.
For all algorithms, we tuned the Lagrange multipliers corresponding to rank using knowledge of the true rank, and optimized for $\cost^{\pi}.$

\textsf{\textbf{Results for Various Metrics:}} We generated 1000 instances with $n=m=100$. The results are summarized in \cref{tb:comparison}. \cref{tb:comparison} reports the regret (i.e. cost above that of $\pi^*$), along with $\|\hat{M}-M\|_{\F}$ and $\|\hat{M}-M\|_{\max}$. all averaged over the $1000$ instances ($\hat{M}$ of EW is obtained after recovering from the estimated scaling). The results show that EW outperforms all other algorithms significantly along all metrics. The reduction in cost brought upon by EW is promising, suggesting the usefulness of incorporating cost information and underlying entry-wise anomaly models. 

\begin{table}[h]
\begin{center}
\begin{tabular}{@{}lccc@{}}
\toprule
 Algorithm & $\cost^{\pi}-\cost^{\pi^{*}}$ & $\|\hat{M}-M\|_{\text{F}}$ & $\|\hat{M}-M\|_{\text{max}}$ \\ 
\midrule 
EW & 0.06& 237.1 & 27.4 \\
Stable PCP& 0.70 & 314.3  & 43.6\\  
 DRMF &0.69& 391.2 &  60.4\\
 RMC &0.90 & 1099.0   & 123.1\\
 \bottomrule 
 \end{tabular}
 \vspace{10pt}
 \caption{Summary of results on synthetic data. Costs (relative to the idealized algorithm $\pi^*$, along with $\|\hat{M}-M\|_{\F}$ and $\|\hat{M}-M\|_{\max}$, are averaged over 1000 instances. The evaluated algorithms include our algorithm (EW), and three existing benchmarks.
}
 \label{tb:comparison}
\end{center}
\end{table}

\begin{figure}[h]
\centering
    \includegraphics[scale=0.6]{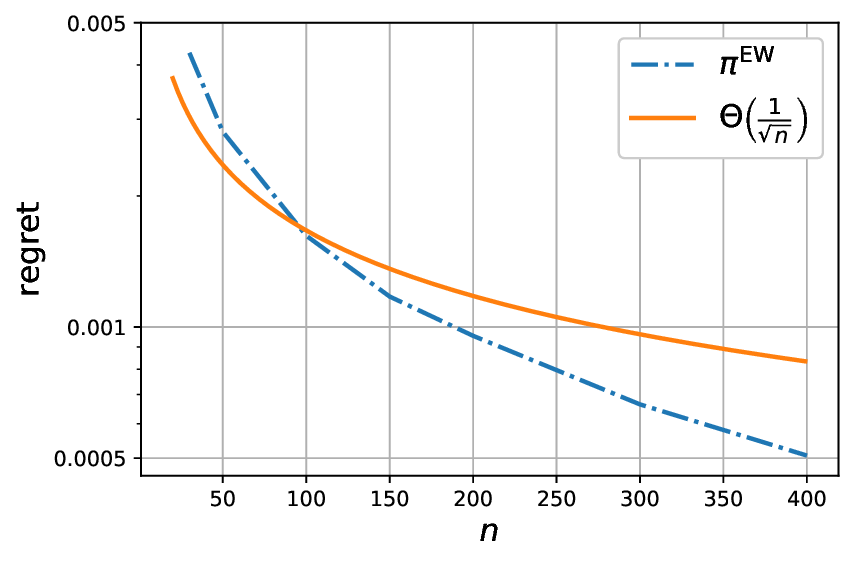}
     \caption{{\bf Synthetic data.} The regret of $\cost^{\pi^{\EW}} - \cost^{\pi^{*}}$ corresponds to $n$ in a representative setting with $n=m, r = 3, \bar{M}^{*} = 5, p_{\O}=0.8, p_{\A}^{*}=0.04, \alpha^{*}=0.2.$}
  \label{fig:regret}
\end{figure}

\textsf{\textbf{Vanishing Regret When $n$ Scales:}} To study the trend of the regret of EW when $n$ increases, we consider a representative setting with $n=m, r = 3, \bar{M}^{*} = 5, p_{\O}=0.8, p_{\A}^{*}=0.04, \alpha^{*}=0.2.$  \cref{fig:regret} shows how the regret of EW scales with $n$. The result confirms \cref{thm:main-theorem}: the regret will vanish when $n$ grows (in fact, the rate in this example is slightly faster than the upper bound $\Theta(1/\sqrt{n})$). This illustrates the power of using cross-sectional data with more available stores and products. 

\textsf{\textbf{Evaluation of Anomaly Detection as a Classification Task:}} Another interesting metric related to anomaly detection, besides the average cost/benefits, is the rate of successfully detecting anomalies. To make this precise, suppose the `goal' of an algorithm $\pi$ were to correctly {\em classify} the entries into an `anomaly set' and a `non-anomaly set'. This is precisely a {\em classification} task, in the statistical learning sense, and as such we can measure performance via the standard true positive rates (TPR) and false positive rates (FPR).

Our EW algorithm can easily be generalized to obtain (near) optimal TPR with a given constraint on FPR (see Appendix \ref{sec:cost-model} for details). To compare the performance, we measure the area under receiver operating characteristic (ROC) curves, i.e., AUCs. For existing methods, we generate ROC curves by varying the Lagrange multipliers.  
We also consider the idealized algorithm $\pi^*$ that knows $M^*$ and the anomaly model.

\begin{figure*}
  \begin{subfigure}[t]{.5\textwidth}
  \centering
    \includegraphics[scale=0.6]{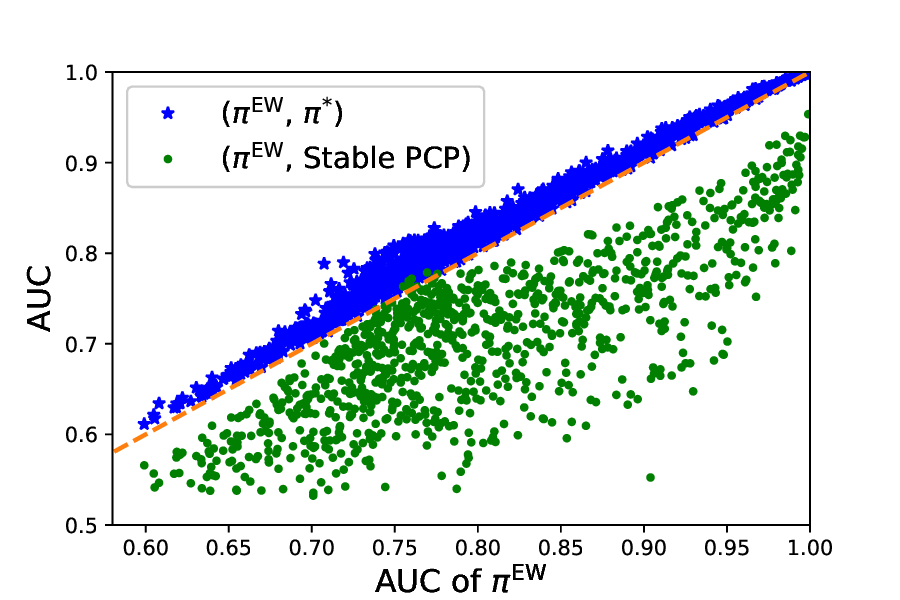}
    \label{fig:synthetic-two}
\end{subfigure}    
    \hfill
  \begin{subfigure}[t]{0.5\textwidth}
    \centering
    \includegraphics[scale=0.6]{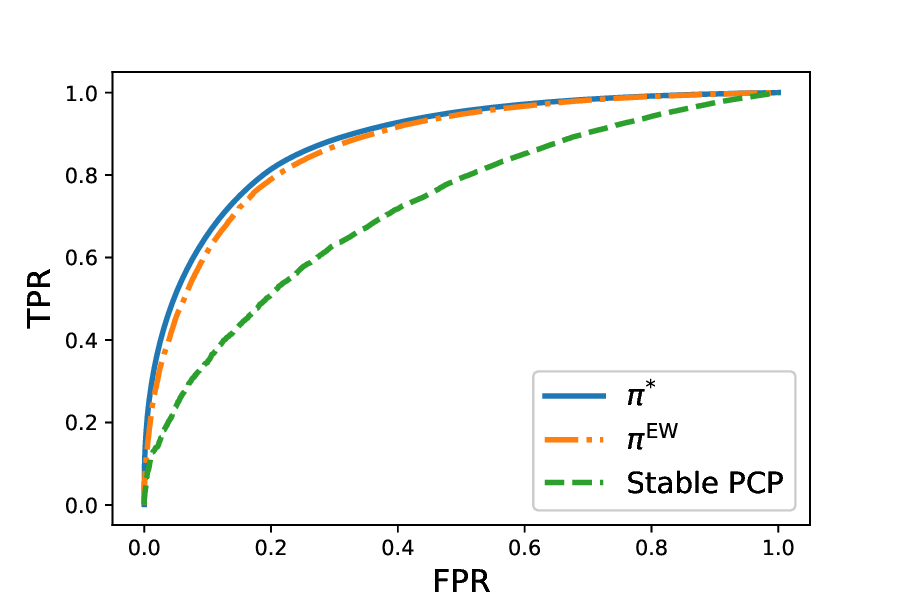}
    \label{fig:synthetic-three}
  \end{subfigure}
  \caption{{\bf Synthetic data.} (Left) Scatter plot showing AUC of (unachievable) ideal algorithm vs. that of EW (blue points, above 45-degree line); and AUC of Stable PCP vs EW (green, mostly below 45 degree line). (Right) ROC curve in a representative setting with $n=m=100, r = 3, \bar{M}^{*} = 5, p_{\O}=0.8, p_{\A}^{*}=0.04, \alpha^{*}=0.2.$}
  \label{fig:synthetic}
\end{figure*}

We generated 1000 ensembles with $n=m=100.$ Figure~\ref{fig:synthetic} (Left) shows the scatter plot of AUCs of $\pi^{*}, \pi^{\EW}$ and Stable-PCP. It confirms that our algorithm $\pi^{\EW}$ achieves  similar AUC to $\pi^{*}$, confirming its the near-optimality. The results also show that we outperform existing methods uniformly over the ensemble (we show our results vs. Stable-PCP, but the same holds true for the other two existing algorithms in the experiments). Figure~\ref{fig:synthetic} (Right) shows the explicit ROC curve for a representative setting. 
%

\begin{figure}
\centering
\includegraphics[scale=0.8]{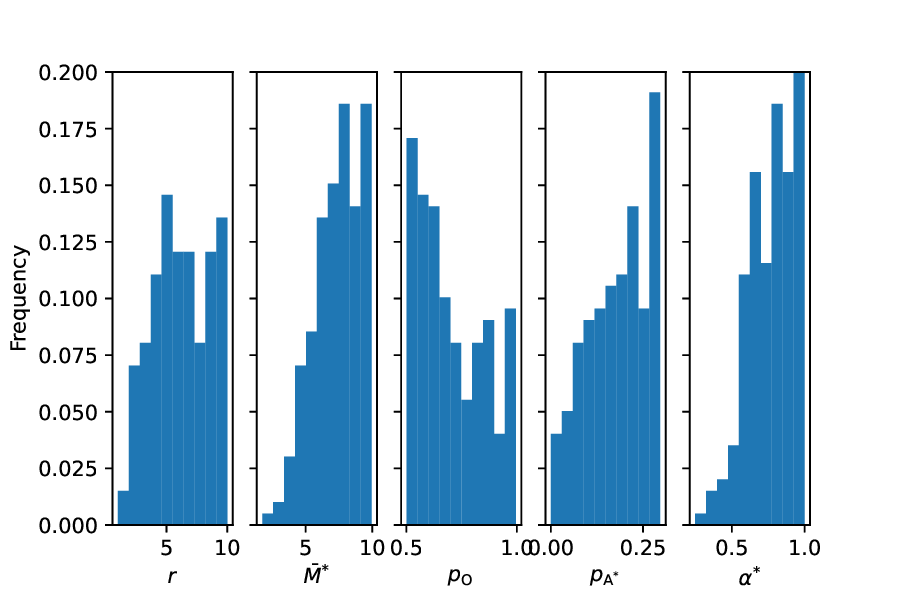}
\caption{Synthetic data. Histograms shows problem characteristics where EW performs worst relative to ideal algorithm (20th percentile).}
\label{fig:synthetic-ROC}
\end{figure}

We also studied the limitation of our algorithm, in which the performance starts to degrade. Figure~\ref{fig:synthetic-ROC} shows that the problem instances (in the experiment of the synthetic data) where the AUC of EW was furthest away from the ideal AUC (20th percentile). The results show largely intuitive characteristics: higher $\alpha^{*}$ (so anomalies look similar to non-anomalous entires), lower $p_\O$, higher $p_\A^{*}$ and higher $r$ (so that $M^*$ is harder to estimate). The behavior with respect to $\bar M^*$ is surprising but was consistently observed across other ensembles as well. 

\vspace{1em}

\subsection{Real Data} \label{sec:real-data}

We collected data $X_{\Omega}$, from a national retailer, consisting of weekly sales of $m=300$ SKUs across $n=30$ stores with $p_{\O} \sim 0.23$. Since there is no ground-truth for anomalies, we backed out the ``real'' anomalies through the following process:
\begin{itemize}
\item Let $\bar{S}_{ij}$ be the unit sales of $(i, j)$ averaged over the 10 weeks surrounding the current week (including the past 5 weeks and future 5 weeks).
\item Treat $(i, j)$ as anomalous if sales at the current week are less than one half of $\bar{S}_{ij}$, i.e., $B_{ij} = 1$ if and only if $X_{ij} \leq \bar{S}_{ij} / 2.$
\end{itemize}

Such anomalies are of practical interest, e.g., a sudden drop of sale might be due to the errors in the inventory records. Even if no phantom inventory event occurs, these products may still warrant a careful examination to understand why the drop in sales occurred. We  simulated the cost model by generating each $c_{ij}^{(0)} \in [0, 500], c_{ij}^{(1)} \in [0, 500]$ uniformly.

\begin{table}[h]
\begin{center}
\begin{tabular}{@{}lcc@{}}
\toprule
 Algorithm & $\cost^{\pi}$ & \\ 
\midrule 
EW & 0.860 ($\pm$ 0.302)\\
 DRMF &2.294 ($\pm$ 0.185)\\
 Ignore & 2.631 ($\pm$ 0.198) \\
 \bottomrule 
 \end{tabular}
 \vspace{10pt}
 \caption{Real Data. Regret of costs are averaged over 1000 instances.
}
 \label{tb:real-regret}
\end{center}
\end{table}

We generated an ensemble of $1000$ such instances (randomness is due to the cost model). \cref{tb:real-regret} reports the results (we show EW vs. DRMF, the other two existing algorithms are significantly worse and substantially slower in this experiment). The ``Ignore'' algorithm serves as the benchmark that simply ignores the inconsistent inventory records (which is not uncommon in practice). The cost of EW is roughly one third that of DRMF, suggesting its practical usefulness. It is also interesting to note that here we are able to discover the sales drop using only the cross-sectional data. Furthermore, the $\Omega$ and $B$ in this experiment are in fact highly non-uniform due to the various correlations over missingness and sales (such as store-correlation, SKU-correlation, and time-correlation) in the real data. The results thus suggested the practical robustness of EW algorithm in the presence of non-uniform missing patterns and anomaly patterns. More details about experiments can be found in Appendix \ref{sec:experiments-appendix}.

\vspace{1em}
\noindent
\textsf{\textbf{Scalability:}} EW is also much faster than the competing algorithms, since our main computational cost is a typical matrix completion procedure. Concretely, we can expect to solve a $70000 \times 10000$ matrix with $10^7$ observed entries within minutes \citep{yao2018accelerated}.

\vspace{1em}

\section{Proof Sketches} \label{sec:proof-sketches}
Before concluding, we sketch the proofs of \cref{thm:main-theorem,thm:entrywise-bound}. Complete proofs can be found in Appendices \ref{sec:main-theorem-appendix} and \ref{sec:entry-wise-appendix}.
\vspace{1em}
\subsection{Proof Sketch of \cref{thm:main-theorem}} \label{sec:proof-main-theorem}

Mirroring the algorithm itself, the following sketch is given in four parts: (i) an entrywise guarantee for $\hat{M}$; (ii) a moment matching estimator for $(\hat{p}_{A}, \hat{\alpha})$; (iii) a plug-in estimator for $f_{ij}^*$ ; (iv) an analysis of the cost incurred by the decision $A_{ij}^{\pi^{\mathrm{EW}}}$.

\vspace{1em}
\noindent
\textsf{\textbf{Step 1:}} The entrywise guarantee for $\hat{M}$, namely \cref{thm:entrywise-bound}, is our key result, and we will sketch its proof in the next subsection.

\vspace{1em}
\noindent
\textsf{\textbf{Step 2:}} The recovery guarantee for the moment matching estimator, namely \cref{lem:step2}, relies on an intermediate step which is to show that $F_{t}(\hat{\theta}) \approx F_{t}(\theta^{*})$ by solving $\hat{\theta}$ from \cref{eq:optimize-F}. In fact, we have the following result: 
\begin{lemma}\label{lem:step2-F}
With probability $1-O(\frac{1}{nm})$, \[\norm{F(\hat{\theta}) - F(\theta^{*})} \leq C (K+L)\kappa^4 \mu r L \sqrt{\frac{\log(m)}{p_{\O}m}}.\]
\end{lemma}
The additional regularity conditions then allow us to establish that $\hat{\theta} \approx \theta^{*}$ from $F(\hat{\theta}) \approx F(\theta^{*})$.

\vspace{1em}
\noindent
\textsf{\textbf{Step 3:}}
The following lemma translates the previously-established entrywise guarantees on $\|\hat{M}_{ij} - e(\theta^{*})M^{*}_{ij}\|$ (\cref{thm:entrywise-bound}), and $\theta^{*} \approx \hat{\theta}$ (\cref{lem:step2}), into closeness between $\hat{f}_{ij}$ and $f_{ij}^{*}$:
\begin{lemma}\label{lem:point-estimate}
Let \[\delta =  (K+L)^3\kappa^4 \mu r L^2 \sqrt{\frac{\log m}{p_{\O}m}}.\] Then with probability $1-O(\frac{1}{nm})$, for every $(i, j) \in \Omega$, we have 
$$
\left|f_{ij}^{*} - \hat{f}_{ij}\right| \leq \min\left(\frac{2\delta}{x_{ij}^{*}+y_{ij}^{*}}, 1\right).
$$
\end{lemma}

\vspace{1em}
\noindent
\textsf{\textbf{Step 4:}}
The final step is to analyze the cost incurred by using $\hat{f}_{ij}$ to replace $f_{ij}^{*}$ in the optimal decision rules. To simplify the notation, define auxiliary variables $a_{ij}$ and $b_{ij}$: 
\begin{align*}
a_{ij} &:= -c_{ij}^{(1)} + \left(c_{ij}^{(0)} + c_{ij}^{(1)} \right)\hat{f}_{ij}\\
b_{ij} &:= -c_{ij}^{(1)} + \left(c_{ij}^{(0)} + c_{ij}^{(1)} \right)f^{*}_{ij}.
\end{align*}
Then, we can write the (excess) cost explicitly as the following:
\begin{align*}
\cost^{\pi^{\EW}}(X_{\Omega}) - \cost^{\pi^{*}}(X_{\Omega}) 
&= \frac{1}{|\Omega|}\sum_{(i,j)\in \Omega} (\1{a_{ij}\leq 0} - \1{b_{ij} \leq 0})b_{ij} \\
&\leq \frac{1}{|\Omega|}\sum_{(i,j)\in \Omega}  |a_{ij} - b_{ij}|.
\end{align*}
Here, the last inequality is simply due to  algebra (one can check the four cases (i) $a_{ij} \geq 0, b_{ij} \geq 0$ (ii) $a_{ij} \leq 0, b_{ij} \leq 0$ (iii) $a_{ij} \geq 0, b_{ij} \leq 0$ (iv) $a_{ij} \leq 0, b_{ij} \geq 0$  separately).  This implies that
\begin{align*}
\cost^{\pi^{\EW}}(X_{\Omega}) - \cost^{\pi^{*}}(X_{\Omega}) 
&\leq \frac{1}{|\Omega|}\sum_{(i,j) \in \Omega}  \left|c_{ij}^{(0)} + c_{ij}^{(1)} \right||\hat{f}_{ij} - f^{*}_{ij}| \\
&\overset{(i)}{\leq} \frac{1}{|\Omega|}\sum_{(i,j) \in \Omega} C \min\left(\frac{2\delta}{x_{ij}^{*}+y_{ij}^{*}}, 1\right),
\end{align*}
where (i) is due to \cref{lem:point-estimate} and taking $C := \max (|c_{ij}^{(0)}+c_{ij}^{(1)}|).$

We now construct random variables \[z_{ij} := \min\left(\frac{2\delta}{x_{ij}^{*}+y_{ij}^{*}}, 1\right) \1{(i,j) \in \Omega}.\] The problem boils down to bound $\sum_{ij} z_{ij}.$ Note that $z_{ij} \in [0, 1].$ Furthermore, 
\begin{align*}
\E{z_{ij}} 
&= p_{\O} \sum_{t= 0,1, 2\dotsc } \min\left(\frac{2\delta}{\Prob{X_{ij}=t}}, 1\right) \Prob{X_{ij}=t} \\
&\leq p_{\O}\sum_{t=0}^{k} 2\delta + p_{\O} \sum_{t=k+1}^{\infty} \Prob{X_{ij}=t}\\
&\leq p_{\O}(2k\delta + \Prob{X_{ij} > k}).
\end{align*}
We choose $k = \Theta(L\log(1/\delta))$. By the sub-exponentiality of $X_{ij}$, we have $\Prob{X_{ij} > k} = O(\delta).$ Then
$$
\E{z_{ij}} = O(p_{\O}\delta \log(1/\delta)).
$$
Finally, by the Chernoff bound ($z_{ij}$ are independent from each other), with probability $1-O(1/(nm)), $
\begin{align*}
\cost^{\pi^{\EW}}(X_{\Omega}) - \cost^{\pi^{*}}(X_{\Omega}) \overset{(i)}{\leq} \frac{C}{2 nmp_{\O}}\sum_{i,j} z_{ij} = O\left(\frac{\log^{1.5}(m)}{\sqrt{m}}\right),
\end{align*}
where (i) uses that $|\Omega| \leq 2 nmp_{\O}$ with high probability and $\delta = O(\log^{0.5}(m)/\sqrt{m})$. This completes the proof. 

\vspace{1em}

\vspace{1em}

\subsection{Proof Sketch of \cref{thm:entrywise-bound}: \\Entrywise Guarantee for Sub-Exponential Random Matrices}\label{sec:entry-wise}
In this subsection, we provide a proof sketch for \cref{thm:entrywise-bound} (the full proof can be found in Appendix \ref{sec:entry-wise-appendix}). Our idea combines recently-developed techniques for entrywise analysis for random matrices \citep{abbe2017entrywise} and recent matrix completion result for Poisson observations \citep{mcrae2019low}. 

The key difficulty is to generalize \cite{abbe2017entrywise}, which provided entrywise results for sub-Gaussian noise to the scenario with sub-exponential noise. Although rich results for sub-Gaussian noise are known, the bound in sub-exponential matrices that we required for generalizing \cite{abbe2017entrywise} is missing. Then we find that a lemma developed in \cite{mcrae2019low} that originally provided the result on aggregated error for Poisson matrix completion can be effectively used in our proof. This observation with considerably more fine-tuned analysis leads us to show the entry-wise guarantee result for matrix completion with Poisson noise (or sub-exponential noise, more generally). 

Since the main result in \cite{abbe2017entrywise} is for the symmetric scenario, we consider $\bar{X}, \bar{M}^{*}$ that are symmetrical under our anomaly model. In particular, let $\bar{M}^{*} \in \R_{+}^{n\times n}$ be a symmetric matrix. For $1\leq i \leq j \leq n$, let 
\begin{align*}
\begin{cases}
\bar{X}_{ij} \sim \Pos(\bar{M}_{ij}^{*}) & \text{with prob. } (1-p_\A^{*})p_\O\\
\bar{X}_{ij} \sim \Ano(\alpha^{*}, \bar{M}_{ij}^{*}) & \text{with prob. } p_\A^{*}  p_\O\\
\bar{X}_{ij} = 0 & \text{with prob. } 1-p_\O.
\end{cases}
\end{align*}
Let $\bar{X}_{ji} = \bar{X}_{ij}$ for $1\leq i\leq j \leq n.$ Let $t = g(\alpha^{*})p_{\A}^{*}p_\O + (1-p_{\A}^{*})p_\O.$ It is easy to verify $\E{\bar{X}/t} = \bar{M}^{*}.$ Furthermore, suppose $\max\left(\bar{M}_{ij}^{*} + 1, \norm{ \Ano(\alpha^{*}, \bar{M}_{ij}^{*})}_{\psi_1}\right) \leq L$ for $(i,j) \in [n]\times [n].$ Denote the eigenvalues of $\bar{M}^{*}$ by $\lambda_1^{*} \geq \lambda_2^{*} \geq  \dotsc \geq \lambda^{*}_{n}$ with their associated eigenvectors by $\{\bar{u}_j^{*}\}_{j=1}^{n}.$ Denote the eigenvalues of $\bar{X}$ by $\lambda_1 \geq \lambda_2 \geq \dotsc \geq \lambda_{n}$ with their associated eigenvectors by $\{\bar{u}_{j}\}_{j=1}^{n}.$ 

Suppose $r$ is an integer such that $1\leq r < n$. Assume $\bar{M}^{*}$ satisfies $\lambda_1^{*} \geq \lambda_2^{*} \geq \dotsc \geq \lambda_{r}^{*} \geq 0$ and $\lambda_{r+1}^{*} \leq 0.$ Let $    \bar{U}^{*} = (u_{1}^{*}, u_{2}^{*}, \dotsc, u_{r}^{*}) \in \R^{n\times r}, \bar{U} = (u_1, u_2, \dotsc, u_{r}) \in \R^{n\times r}.$ We aim to show that $\bar{U}$ is a good estimation of $\bar{U}^{*}$ in the entry-wise manner under some proper rotation. In particular, let $\bar{H} := \bar{U}^{T}\bar{U}^{*} \in \R^{r\times r}.$ Suppose the SVD decomposition of $\bar{H}$ is $\bar{H} = U'\Sigma'V'^{T}$. The matrix sign function of $\bar{H}$ is denoted by $\sgn(\bar{H}):= U'V'^{T}.$ In fact, $\sgn(\bar{H}) = \arg\min_{O} \norm{\bar{U}O-\bar{U}^{*}}_{\F}$ subject to $OO^{T}=I$.\footnote{See \cite{gross2011recovering} for more details about the matrix sign function.} We aim to show an upper bound on $\norm{\bar{U}\sgn(\bar{H}) - \bar{U}^{*}}_{2,\infty}.$ Let $\Delta^{*} := t\lambda_r^{*}, \kappa := \frac{\lambda_1^{*}}{\lambda_r^{*}}.$ We rephrase the Theorem 2.1 in \cite{abbe2017entrywise} for the above scenario and rewrite it as the following lemma.
\begin{lemma}[Theorem 2.1 \cite{abbe2017entrywise}]\label{lem:generalize-entrywise-bound}
Suppose $\gamma \in \R_{\geq 0}$. Let $\phi(x): \R_{\geq 0} \rightarrow \R_{\geq 0}$ be a continuous and non-decreasing function with $\phi(0) = 0$ and $\phi(x)/x$ non-increasing in $\R_{>0}$. Let $\delta_0, \delta_1 \in (0, 1)$. With the above quantities, consider the following four assumptions:
\begin{enumerate}
    \item[\textbf{A1.}] \label{assum:A1} $\norm{t\bar{M}^{*}}_{2,\infty} \leq \gamma \Delta^{*}.$
    \item[\textbf{A2.}] \label{assum:A2} For any $m \in [n]$, the entries in the $m$th row and column of $\bar{X}$ are independent with others.
    \item[\textbf{A3.}] \label{assum:A3} 
	$\Prob{\norm{\bar{X}-t\bar{M}^{*}}_2 \leq \gamma \Delta^{*}} \geq 1 - \delta_0.$
    \item[\textbf{A4.}] \label{assum:A4} For any $m \in [n]$ and any $W \in \R^{n\times r}$,
    \begin{align*}
        \Prob{\norm{\left(\bar{X}-t\bar{M}^{*}\right)_{m,\cdot}W}_2 \leq \Delta^{*}\norm{W}_{2,\infty}\phi\left(\frac{\norm{W}_{\F}}{\sqrt{n}\norm{W}_{2,\infty}}\right)} \geq 1 - \frac{\delta_1}{n}
    \end{align*}
\end{enumerate}
If $32\kappa \max(\gamma, \phi(\gamma))\leq 1$, under above Assumptions \hyperref[assum:A1]{A1}--\hyperref[assum:A4]{A4}, with probability $1-\delta_0-2\delta_1$, the following hold:
\begin{align*}
    \norm{\bar{U}}_{2,\infty} &\lesssim (\kappa+\phi(1)) \norm{\bar{U}^{*}}_{2,\infty} + \gamma \norm{t\bar{M}^{*}}_{2,\infty}/\Delta^{*}\\
    \norm{\bar{U}\sgn(\bar{H}) - \bar{U}^{*}}_{2,\infty} &\lesssim (\kappa(\kappa+\phi(1))(\gamma+\phi(\gamma)) + \phi(1)) \norm{\bar{U}^{*}}_{2,\infty} + \gamma \norm{t\bar{M}^{*}}_{2,\infty}/\Delta^{*}.
\end{align*}
\end{lemma}

To obtain useful results from \cref{lem:generalize-entrywise-bound}, one need to find proper $\gamma$ and $\phi(x)$ and show that the Assumptions \hyperref[assum:A1]{A1}--\hyperref[assum:A4]{A4} hold. For sub-gaussian noise, these assumptions can be easily verified. We investigate the scenario for sub-exponential noise. We define $\bar{\gamma}$ and $\bar{\phi}(x)$ as the proper form for $\gamma$ and $\phi(x)$ respectively in the following.
\begin{definition} 
Let $\bar{\gamma} := \frac{\sqrt{n}}{\Delta^{*}}L, \bar{\phi}(x) :=  \frac{\sqrt{n}}{\Delta^{*}}L \log (2n^3r)  x$.
\end{definition}
Under $\bar{\gamma}$ and $\bar{\phi}(x)$, we will show that Assumption \hyperlink{assum:A3}{A3} holds based on \cref{lem:A2}, Assumption \hyperlink{assum:A4}{A4} holds based on \cref{lem:A4}. Note that Assumption \hyperlink{assum:A2}{A2} naturally holds since each element of $\bar{X}$ is independent of each other. Assumption \hyperlink{assum:A1}{A1} holds due to that $\norm{t\bar{M}^{*}}_{2,\infty} = \max_{i} \sqrt{\sum_{j} t^2\bar{M}^{*2}_{ij}} \leq t\sqrt{n}L \leq \bar{\gamma}\Delta^{*}.$ First, we observe that $\bar{X}_{ij} - t\bar{M}_{ij}^{*}$ is a sub-exponential random matrix. 
\begin{lemma}\label{lem:element-tail}
For any $(i,j) \in [n]\times[n]$, $\norm{\bar{X}_{ij} - t\bar{M}_{ij}^{*}}_{\psi_1} \leq 6L$.
\end{lemma}

To show that Assumption \hyperlink{assum:A3}{A3} holds, we introduce a result in \cite{mcrae2019low} that helps to control the operator norm of $\bar{X} - t\bar{M}^{*}$.
\begin{lemma}[Lemma 4 in \cite{mcrae2019low}] \label{lem:operator-norm-bound}
Let $Y$ be a random $n_1\times n_2$ matrix whose entries are independent and centered, and suppose that for some $v, t_0 > 0$, we have, for all $t_1\geq t_0$, $\Prob{|Y_{ij}| \geq t_1} \leq 2e^{-t_1/v}.$
Let $\epsilon \in (0, 1/2),$ and let $K = \max\{t_0, v\log (2mn/\epsilon)\}.$ Then,
\begin{align*}
\Prob{\norm{Y}_2 \geq 2\sigma + \frac{\epsilon v}{\sqrt{n_1 n_2}} + t_1} \leq \max(n_1,n_2) \exp(-t_1^2/(C_0(2K)^2)) + \epsilon,
\end{align*}
where $C_0$ is a constant and $\sigma = \max_{i} \sqrt{\sum_{j} \E{Y_{ij}^2}} + \max_{j} \sqrt{\sum_{i} \E{Y_{ij}^2}}.$
\end{lemma}
In order to use \cref{lem:operator-norm-bound} for Assumption \hyperlink{assum:A3}{A3}, we need to convert the asymmetrical results into symmetric scenario. In particular, we have
\begin{lemma}\label{lem:A2}
$\Prob{\norm{\bar{X} - t\bar{M}^{*}}_2 \leq C\bar{\gamma} \Delta^{*}} \geq 1-\frac{1}{n^2}$ for some constant $C$.  
\end{lemma}
\begin{proof}{Proof.}
Denote $Y \in \R^{n\times n}$ by
\begin{align*}
Y_{ij} = \begin{cases}
2(\bar{X}_{ij} - t\bar{M}^{*}_{ij}) & i < j\\
(\bar{X}_{ij} - t\bar{M}^{*}_{ij}) & i = j\\
0 & i>j
\end{cases}.
\end{align*} 

Note that $\norm{Y_{ij}}_{\psi_1} \leq 2\norm{\bar{X}_{ij} - t\bar{M}^{*}_{ij}}_{\psi_1} \leq 12L$ by \cref{lem:element-tail}. By the property of subexponential random variable, we have $\E{Y_{ij}^2} \leq C_1L^2$ and for all $t'\geq 0$, $\Prob{|Y_{ij}| \geq t'} \leq 2\exp(-t'/C_2L)$ where $C_1, C_2$ are two constants.

Consider applying \cref{lem:operator-norm-bound} to $X$ with $n_1=n_2=n$. Let $\epsilon = \frac{1}{2n^2}.$ Then $K = C_2 L \log (4n^2).$ Take $t' = \sqrt{C_0 3\ln n} 2K + \ln 2$. Then $\max(n_1,n_2) \exp(-t'^2/(C_0(2K)^2)) + \epsilon = \frac{1}{n^2}.$ Furthermore
$2\sigma + \frac{\epsilon v}{\sqrt{n_1n_2}} + t' \leq  C_3 \sqrt{n}L$
for some constant $C_3$. 

Therefore, $\norm{Y}_2 \leq C_3 \sqrt{n}L$ with probability $1-\frac{1}{n^2}.$ Note that $\bar{X} - t\bar{M}^{*} = (Y + Y^{T})/2.$ Hence, with probability $1-\frac{1}{n^2},$ $\norm{\bar{X} - t\bar{M}^{*}}_2 \leq (\norm{Y}_2 + \norm{Y^{T}}_2) / 2 \leq C_3\sqrt{n}L \leq C_3\bar{\gamma} \Delta^{*}.$
\end{proof}

Next, we will show that Assumption \hyperlink{assum:A4}{A4} holds based on the Bernstein-type  inequalities to control the tail bound of sum of sub-exponential random variables. 
\begin{lemma}\label{lem:A4}
For any $m \in [n]$ and any $W \in \R^{n \times r}$, the following holds
\begin{align*}
\Prob{\norm{(\bar{X}-t\bar{M}^{*})_{m,\cdot}W}_2 \leq C\Delta^{*}\norm{W}_{2,\infty}\bar{\phi}\left(\frac{\norm{W}_{\F}}{\sqrt{n}\norm{W}_{2,\infty}}\right)} \geq 1 - \frac{1}{n^3},
\end{align*}
where $C$ is a constant.
\end{lemma}
\begin{proof}{Proof.}
First, consider a special case with $r=1, w \in \R^{n\times 1}.$ Let $Y_{j} = \bar{X}_{ij}-t\bar{M}_{ij}^{*}$. By \cref{lem:element-tail}, we have $\max_{j\in[n]} \norm{Y_{j}}_{\phi_1} \leq 6L =: K.$ Then, by Bernstein's inequality, 
$\Prob{|\sum_{i=1}^{N} w_{i} Y_{i}| \geq t} \leq 2\exp\left\{-C_2 \left(\frac{t^2}{K^2\norm{w}_2^2} \vee \frac{t}{K\norm{w}_{\infty}}\right)\right\}.$

Take $t = (\frac{1}{C_2}+1)\norm{w}_2 K \log (2n^3r)$, then $C_2\frac{t^2}{K^2\norm{w}_2^2} = (\frac{1}{C_2}+1)(1+C_2) \log(2n^3 r) \log(2n^3 r) \geq \log(2n^3 r)$ and $
   C_2\frac{t}{K\norm{w}_{\infty}} = (1+C_2)\frac{\norm{w}_2}{\norm{w}_{\infty}} \log (2n^3 r) \geq \log(2n^3 r).$ Therefore,  
   \begin{align*}
   \Prob{|\sum_{i=1}^{N} w_{i} Y_{i}| \geq (\frac{1}{C_2}+1)\norm{w}_2 K \log (2n^3r)} \leq \frac{1}{n^3r}.
   \end{align*}
This idea can be generalized to the scenario with $r>1$.
\end{proof}


After showing that Assumptions \hyperref[assum:A1]{A1}--\hyperref[assum:A4]{A4} hold, we can prove the following result.
\begin{proposition}\label{thm:symmetric} Let $t := (g(\alpha^{*})p_{\A}^{*}+(1-p_{\A}^{*}))p_\O.$
Suppose $\sqrt{n}L\log (n) \kappa^2 \leq Ct\lambda_1^{*}$ for some known constant $C$. Then, with probability $1-O(n^{-2})$, the following hold:
\begin{align*}
    \norm{\bar{U}}_{2,\infty} &\lesssim \kappa (\norm{\bar{U}^{*}}_{2,\infty} + \norm{\bar{M}^{*}}_{2,\infty}/\lambda_1^{*})\\
    \norm{\bar{U}\sgn(H) - \bar{U}^{*}}_{2,\infty} &\lesssim \frac{\sqrt{n}\log (n)}{t\lambda_1^{*}} \kappa^3L\norm{\bar{U}^{*}}_{2,\infty} + \kappa \norm{\bar{M}^{*}}_{2,\infty} /\lambda_1^{*}.
\end{align*}
\end{proposition}

Next we need to convert \cref{thm:symmetric} back to our asymmetric scenario. We use the ``symmetric dilation'' technique \citep{paulsen2002completely,abbe2017entrywise,ma2019implicit,chen2020nonconvex} to generalize the result to accommodate our asymmetric model. Let $M^{*}=U^{*}\Sigma^{*}V^{*}, \SVD(X_{\Omega})_{r} = U \Sigma V$ be the SVD decomposition of $M^{*}$ and $\SVD(X_{\Omega})_{r}$ respectively. Some fine-tuned analysis leads to the following result. 

\begin{proposition} \label{prop:entry-wise-bound}
Let $H = \frac{1}{2}(U^{T}U^{*} + V^{T}V^{*}), \mu = \max\left(n\norm{U^{*}}_{2,\infty}^2, m \norm{V^{*}}_{2,\infty}^2\right)/r, \kappa = \sigma_1^{*}/\sigma_r^{*}, t = (p_{\A}^{*}g(\alpha^{*}) + 1-p_{\A}^{*})p_\O$. Suppose $\sqrt{m} L \log (m)\kappa^2 \leq C t \sigma_1^{*}$ for some known constant $C$. Then, with probability $1-O((nm)^{-1})$, the following hold:
\begin{align*}
(\norm{U}_{2,\infty} \vee \norm{V}_{2, \infty}) &\lesssim  \kappa\sqrt{\frac{\mu r}{n}} \\
    (\norm{U\sgn(H) - U^{*}}_{2,\infty} \vee \norm{V\sgn(H) - V^{*}}_{2, \infty}) &\lesssim  \frac{\sqrt{m}\log(m)\kappa^3L\sqrt{\mu r}}{t\sigma_1^{*}\sqrt{n}} \\
    \norm{\SVD(X_{\Omega})_{r} - t M^{*}}_{\max} &\lesssim \kappa^4 \mu r \log(m) L \frac{\sqrt{m}}{n}.
\end{align*}
\end{proposition}
 Finally, a concentration bound implying that $|\Omega| \approx nmp_{\O}$ provides us
 \begin{align*}
 \norm{\SVD(X_{\Omega})_{r} \frac{nm}{|\Omega|} - e(\theta^{*})M^{*}}_{\max} & \lesssim \frac{(\kappa^4\mu r)L}{p_\O} \frac{\log m \sqrt{m}}{n},
 \end{align*}
 which completes the proof (sketch).

\vspace{1em}

\section{Conclusion} \label{sec:conclusion}

We proposed a simple statistical model for anomaly detection in low-rank matrices that is motivated by fixing inventory inconsistency in retail. We proved a new entrywise bound for matrix completion with sub-exponential noise, and used this to motivate a simple policy for the anomaly detection problem. We proved matching upper and lower bounds on the anomaly detection costs of our algorithm, and demonstrated in experiments that our approach provides substantial improvements over existing approaches. 

While our results are somewhat encouraging, they by no means cover the most general settings of practical interest. There are many possible extensions that merit future investigation, to name a few, 
\begin{itemize}

\item \textbf{Dependency on $K$ and $L$.} Our current regret likely scales sub-optimal with $K$ amd $L$. A more refined analysis may lead to the improvement for such dependency. 

\item \textbf{Incorporating Longitudinal Information.} In this paper, we focus on exploring the benefits brought by utilizing the cross-sectional data in a dynamic environment. But if the environment is relatively stable, one can expect that both longitudinal information and cross-sectional shall help. The combination of such tensor-type information is a promising future direction to study.   
\end{itemize}



\vspace{1em}

\bibliographystyle{informs2014}
\bibliography{reference} 
%
%

%
%


\newpage
\begin{appendices}
\crefalias{section}{appendix}
\setcounter{lemma}{5}
\setcounter{proposition}{1}
\setcounter{definition}{1}

\newpage




\vspace{1em}
\section{Entry-wise Bound and Proof of Theorem \ref{thm:entrywise-bound}}\label{sec:entry-wise-appendix}
In this section, we will prove  \cref{thm:entrywise-bound} based on recent results on entry-wise analysis for random matrices \citep{abbe2017entrywise} and matrix completion with Poisson observation \citep{mcrae2019low}. The proof idea can be viewed as a generalization from Gaussian noise in  Theorem 3.4 of \cite{abbe2017entrywise} to subexponential noise. In particular, we will proceed with the proof in two steps: (i) consider the symmetric scenario where $M^{*}$, noises, and anomalies have symmetries; (ii) generalize the results to the asymmetric scenario.  

\vspace{1em}
\subsection{Symmetric Case}
Consider a symmetric scenario. Let $\bar{M}^{*} \in \R_{+}^{n\times n}$ be a symmetric matrix. For $1\leq i \leq j \leq n$, let 
\begin{align}
\begin{cases}
\bar{X}_{ij} \sim \Pos(\bar{M}_{ij}^{*}) & \text{with prob. } (1-p_\A^{*})p_\O\\
\bar{X}_{ij} \sim \Ano(\alpha^{*}, \bar{M}_{ij}^{*}) & \text{with prob. } p_\A^{*}  p_\O\\
\bar{X}_{ij} = 0 & \text{with prob. } 1-p_\O.
\end{cases}\label{eq:defintion-Xbar}
\end{align}
Let $\bar{X}_{ji} = \bar{X}_{ij}$ for $1\leq i\leq j \leq n.$ Let $t = g(\alpha^{*})p_{\A}^{*}p_\O + (1-p_{\A}^{*})p_\O.$ It is easy to verify $\E{\bar{X}/t} = \bar{M}^{*}.$ Furthermore, suppose $\max\left(\bar{M}_{ij}^{*} + 1, \norm{ \Ano(\alpha^{*}, \bar{M}_{ij}^{*})}_{\psi_1}\right) \leq L$ for $(i,j) \in [n]\times [n].$

Denote the eigenvalues of $\bar{M}^{*}$ by $\lambda_1^{*} \geq \lambda_2^{*} \geq  \dotsc \geq \lambda^{*}_{n}$ with their associated eigenvectors by $\{\bar{u}_j^{*}\}_{j=1}^{n}.$ Denote the eigenvalues of $\bar{X}$ by $\lambda_1 \geq \lambda_2 \geq \dotsc \geq \lambda_{n}$ with their associated eigenvectors by $\{\bar{u}_{j}\}_{j=1}^{n}.$ 

Suppose $r$ is an integer such that $1\leq r < n$. Assume $\bar{M}^{*}$ satisfies $\lambda_1^{*} \geq \lambda_2^{*} \geq \dotsc \geq \lambda_{r}^{*} \geq 0$ and $\lambda_{r+1}^{*} \leq 0.$ Let $    \bar{U}^{*} = (u_{1}^{*}, u_{2}^{*}, \dotsc, u_{r}^{*}) \in \R^{n\times r}, \bar{U} = (u_1, u_2, \dotsc, u_{r}) \in \R^{n\times r}.$ We aim to show that $\bar{U}$ is a good estimation of $\bar{U}^{*}$ in the entry-wise manner under some proper rotation. In particular, let $\bar{H} := \bar{U}^{T}\bar{U}^{*} \in \R^{r\times r}.$ Suppose the SVD decomposition of $\bar{H}$ is $\bar{H} = U'\Sigma'V'^{T}$ where $U', V' \in \R^{r \times r}$ are orthonormal matrices and $\Sigma' \in \R^{r\times r}$ is a diagonal matrix. The matrix sign function of $\bar{H}$ is denoted by $\sgn(\bar{H}):= U'V'^{T}.$ In fact, $\sgn(\bar{H}) = \arg\min_{O} \norm{\bar{U}O-\bar{U}^{*}}_{\F}$ subject to $OO^{T}=I$.\footnote{See \cite{gross2011recovering} for more details about the matrix sign function.} We aim to show an upper bound on $\norm{\bar{U}\sgn(\bar{H}) - \bar{U}^{*}}_{2,\infty}.$ 

Let $\Delta^{*} := t\lambda_r^{*}, \kappa := \frac{\lambda_1^{*}}{\lambda_r^{*}}.$ We rephrase the Theorem 2.1 in \cite{abbe2017entrywise} for the above scenario and rewrite it as the following lemma.
\begin{lemma}[Theorem 2.1 \cite{abbe2017entrywise}]\label{lem:generalize-entrywise-bound}
Suppose $\gamma \in \R_{\geq 0}$. Let $\phi(x): \R_{\geq 0} \rightarrow \R_{\geq 0}$ be a continuous and non-decreasing function with $\phi(0) = 0$ and $\phi(x)/x$ non-increasing in $\R_{>0}$. Let $\delta_0, \delta_1 \in (0, 1)$. With the above quantities, consider the following four assumptions:
\begin{enumerate}
    \item[\textbf{A1.}] \label{assum:A1} $\norm{t\bar{M}^{*}}_{2,\infty} \leq \gamma \Delta^{*}.$
    \item[\textbf{A2.}] \label{assum:A2} For any $m \in [n]$, the entries in the $m$th row and column of $\bar{X}$ are independent with others.
    \item[\textbf{A3.}] \label{assum:A3} 
	$\Prob{\norm{\bar{X}-t\bar{M}^{*}}_2 \leq \gamma \Delta^{*}} \geq 1 - \delta_0.$
    \item[\textbf{A4.}] \label{assum:A4} For any $m \in [n]$ and any $W \in \R^{n\times r}$,
    \begin{align*}
        \Prob{\norm{\left(\bar{X}-t\bar{M}^{*}\right)_{m,\cdot}W}_2 \leq \Delta^{*}\norm{W}_{2,\infty}\phi\left(\frac{\norm{W}_{\F}}{\sqrt{n}\norm{W}_{2,\infty}}\right)} \geq 1 - \frac{\delta_1}{n}
    \end{align*}
\end{enumerate}
If $32\kappa \max(\gamma, \phi(\gamma))\leq 1$, under above Assumptions \hyperref[assum:A1]{A1}--\hyperref[assum:A4]{A4}, with probability $1-\delta_0-2\delta_1$, the followings hold,
\begin{align*}
    \norm{\bar{U}}_{2,\infty} &\lesssim (\kappa+\phi(1)) \norm{\bar{U}^{*}}_{2,\infty} + \gamma \norm{t\bar{M}^{*}}_{2,\infty}/\Delta^{*}\\
    \norm{\bar{U}\sgn(\bar{H}) - \bar{U}^{*}}_{2,\infty} &\lesssim (\kappa(\kappa+\phi(1))(\gamma+\phi(\gamma)) + \phi(1)) \norm{\bar{U}^{*}}_{2,\infty} + \gamma \norm{t\bar{M}^{*}}_{2,\infty}/\Delta^{*}.
\end{align*}
\end{lemma}

To obtain useful results from \cref{lem:generalize-entrywise-bound}, one need to find proper $\gamma$ and $\phi(x)$ and show that the Assumptions \hyperref[assum:A1]{A1}--\hyperref[assum:A4]{A4} hold. We define $\bar{\gamma}$ and $\bar{\phi}(x)$ as the proper form for $\gamma$ and $\phi(x)$ respectively in the following.
\begin{definition} 
Let $\bar{\gamma} := \frac{\sqrt{n p_{\O}}}{\Delta^{*}}L, \bar{\phi}(x) :=  \frac{\sqrt{n}}{\Delta^{*}}L \log (2n^3r)  x$.
\end{definition}
Under $\bar{\gamma}$ and $\bar{\phi}(x)$, we will show that Assumption \hyperlink{assum:A3}{A3} holds based on \cref{lem:A2}, Assumption \hyperlink{assum:A4}{A4} holds based on \cref{lem:A4}. Note that Assumption \hyperlink{assum:A2}{A2} naturally holds since each element of $\bar{X}$ is independent of each other. Assumption \hyperlink{assum:A1}{A1} holds due to that $\norm{t\bar{M}^{*}}_{2,\infty} = \max_{i} \sqrt{\sum_{j} t^2\bar{M}^{*2}_{ij}} \leq t\sqrt{n}L \leq \bar{\gamma}\Delta^{*}.$

To show that Assumption \hyperlink{assum:A3}{A3} holds, we introduce a result in \cite{mcrae2019low} that helps to control the operator norm of $\bar{X} - t\bar{M}^{*}$.
\begin{lemma}[Lemma 4 in \cite{mcrae2019low}] \label{lem:operator-norm-bound}
Let $Y$ be a random $n_1\times n_2$ matrix whose entries are independent and centered, and suppose that for some $v, t_0 > 0$, we have, for all $t_1\geq t_0$, $\Prob{|Y_{ij}| \geq t_1} \leq 2e^{-t_1/v}.$
Let $\epsilon \in (0, 1/2),$ and let $K = \max\{t_0, v\log (2mn/\epsilon)\}.$ Then,
\begin{align*}
\Prob{\norm{Y}_2 \geq 2\sigma + \frac{\epsilon v}{\sqrt{n_1 n_2}} + t_1} \leq \max(n_1,n_2) \exp(-t_1^2/(C_0(2K)^2)) + \epsilon,
\end{align*}
where $C_0$ is a constant and $\sigma = \max_{i} \sqrt{\sum_{j} \E{Y_{ij}^2}} + \max_{j} \sqrt{\sum_{i} \E{Y_{ij}^2}}.$
\end{lemma}

In order to use \cref{lem:operator-norm-bound}, we show that every entry of $\bar{X} - t\bar{M}^{*}$ is a sub-exponential random variable based on \cref{lem:poisson-bound,lem:combination-bound,lem:element-tail}.

\begin{lemma}\label{lem:poisson-bound}
Let $Y\sim \Pos(\lambda)$. Then $\norm{Y}_{\psi_1} \leq 4\lambda + 1.$
\end{lemma}
\begin{proof}
Note that for any $t_1>0$, 
\begin{align*}
\E{e^{|Y|/t_1}} = \E{e^{Y/t_1}} = \sum_{k=0}^{\infty} e^{k/t_1} \frac{\lambda^{k}}{k!}e^{-\lambda}
= e^{-\lambda}\sum_{k=0}^{\infty} \frac{(e^{1/t_1}\lambda)^{k}}{k!}= e^{-\lambda} e^{e^{1/t_1}\lambda} =  e^{\lambda(e^{1/t_{1}}-1)}.
\end{align*}
Note that $1/(4\lambda+1) \leq 1$, hence $e^{1/(4\lambda+1)}-1 = \frac{1}{4\lambda + 1}e^{s} \leq \frac{1}{4\lambda + 1} e$ where $s\in [0, 1/(4\lambda+1)]$ by Taylor expansion. Therefore
\begin{align*}
   \E{e^{|Y|/(4\lambda+1)}} \leq e^{\frac{\lambda}{4\lambda + 1} e} \leq e^{e/4} \approx 1.973 < 2.
\end{align*}
By the definition of $\norm{\cdot}_{\psi_1}$, we have $\norm{Y}_{\psi_1} \leq 4\lambda+1.$
\end{proof}

\begin{lemma}\label{lem:combination-bound}
Let $Y_1, Y_2 \dotsc Y_q$ be $q$ subexponential random variables with $\norm{Y_i}_{\psi_1} \leq L_{\max}.$ Let $c \in \{1, 2, \dotsc, q\}$ be a random variable. Then $\norm{Y_c}_{\psi_1} \leq L_{\max}.$
\end{lemma}
\begin{proof}{Proof.}
This is because $\E{e^{|Y_c|/L_{\max}}} = \sum_{1\leq i\leq q} \Prob{c=i} \E{e^{|Y_i|/L_{\max}}} \leq \sum_{1\leq i\leq q} 2\Prob{c=i}  = 2.$
\end{proof}
\begin{lemma}\label{lem:element-tail}
For any $(i,j) \in [n]\times[n]$, $\norm{\bar{X}_{ij} - t\bar{M}_{ij}^{*}}_{\psi_1} \leq 6L$.
\end{lemma}
\begin{proof}{Proof.}
Note that $\norm{\Pos(\bar{M}_{ij}^{*})}_{\psi_1} \leq 4L$ by \cref{lem:poisson-bound} and $\norm{\Ano(\alpha^{*}, \bar{M}_{ij}^{*})} \leq L$ by the definition of $L$. We have $\norm{\bar{X}_{ij}}_{\psi_1} \leq 4L$ by \cref{eq:defintion-Xbar} and \cref{lem:combination-bound}. Then, by the triangle inequality, $\norm{\bar{X}_{ij} - t\bar{M}_{ij}^{*}}_{\psi_1} \leq \norm{\bar{X}_{ij}}_{\psi_1}  + \norm{\bar{M}_{ij}^{*}}_{\psi_1} \leq 4L + 2L = 6L$.
\end{proof}

 Next we show that Assumption \hyperlink{assum:A3}{A3} holds.
\begin{lemma}\label{lem:A2}
Suppose $p_{\O} \geq \frac{\log^3 n}{n}$. $\Prob{\norm{\bar{X} - t\bar{M}^{*}}_2 \leq C\bar{\gamma} \Delta^{*}} \geq 1-\frac{1}{n^2}$ for some constant $C$.  
\end{lemma}
\begin{proof}{Proof.}
Denote $Y \in \R^{n\times n}$ by
\begin{align*}
Y_{ij} = \begin{cases}
2(\bar{X}_{ij} - t\bar{M}^{*}_{ij}) & i < j\\
(\bar{X}_{ij} - t\bar{M}^{*}_{ij}) & i = j\\
0 & i>j
\end{cases}.
\end{align*} 

Note that $\norm{Y_{ij}}_{\psi_1} \leq 2\norm{\bar{X}_{ij} - t\bar{M}^{*}_{ij}}_{\psi_1} \leq 12L$ by \cref{lem:element-tail}. By the property of subexponential random variable, for all $t'\geq 0$, $\Prob{|Y_{ij}| \geq t'} \leq 2\exp(-t'/C_1L)$ where $C_1$ is a constant. By the construction of $Y$, we also have
\begin{align}\label{eq:Y-variance}
\E{Y_{ij}} = 0\text{\quad and \quad}\E{Y_{ij}^2} \leq 2E[\bar{X}_{ij}^2] \leq C_2 p_{\O} L^2
\end{align}
for some constant $C_2$.

Consider applying \cref{lem:operator-norm-bound} to $X$ with $n_1=n_2=n$. Let $\epsilon = \frac{1}{2n^2}.$ Then $K = C_1 L \log (4n^2).$ Take $t' = \sqrt{C_0 3\ln n} 2K + \ln 2$. Then $\max(n_1,n_2) \exp(-t'^2/(C_0(2K)^2)) + \epsilon = \frac{1}{n^2}.$ Furthermore, by \cref{eq:Y-variance} and $np_{\O} \geq \log^3(n)$, one can verify that
\begin{align*}
2\sigma + \frac{\epsilon v}{\sqrt{n_1n_2}} + t' \leq  C_3 \sqrt{n p_{\O}}L
\end{align*}
for some constant $C_3$. 

Therefore, $\norm{Y}_2 \leq C_3 \sqrt{n p_{\O}}L$ with probability $1-\frac{1}{n^2}.$ Note that $\bar{X} - t\bar{M}^{*} = (Y + Y^{T})/2.$ Hence, with probability $1-\frac{1}{n^2},$ $\norm{\bar{X} - t\bar{M}^{*}}_2 \leq (\norm{Y}_2 + \norm{Y^{T}}_2) / 2 \leq C_3\sqrt{n p_{\O}}L \leq C_3\bar{\gamma} \Delta^{*}.$
\end{proof}

Next, we will show that Assumption \hyperlink{assum:A4}{A4} holds based on the matrix Bernstein's inequality to control the tail bound of sum of sub-exponential random variables. 
\begin{lemma}[Matrix Bernstein's inequality] \label{lem:bernstein}
Given $n$ independent random $m_1 \times m_2$ matrices $X_1, X_2, \dotsc, X_{n}$ with $E[X_i] = 0.$ Let 
\begin{align}
    V \triangleq \max\left(\norm{\sum_{i=1}^{n} E[X_i X_i^{T}]}, \norm{\sum_{i=1}^{n} E[X_i^{T} X_i]}\right).
\end{align}
Suppose $\norm{\norm{X_i}}_{\psi_1} \leq L$ for $i \in [n].$ Then, 
\begin{align}
    \norm{X_1+X_2+\dotsc + X_{n}}\lesssim  \sqrt{V\log(n(m_1+m_2))} + L\log(n(m_1+m_2))\log(n)
\end{align}
with probability $1-O(n^{-c})$ for any constant $c$.
\end{lemma}

\begin{proof}{Proof.}
Let $Y_i = X_i \1{\norm{X_i} \leq B}$ be the truncated version of $X_i$. We have, 
\begin{align}
    \norm{\E{Y_i}} 
    &\leq \norm{\int X_i \1{\norm{X_i} > B} df(X_i)} \nonumber\\
    &\overset{(i)}{\leq} \int \norm{X_i} \1{\norm{X_i} > B} df(X_i) \nonumber\\
    &\leq B P(\norm{X_i} > B) + \int_{B}^{\infty} P(\norm{X_i} > t) dt \nonumber\\
    &\overset{(ii)}{\leq} Be^{-B/CL} + CL e^{-B/CL} \label{eq:Ex-Yi-bound}
\end{align}
where (i) is due to the convexity of $\norm{\cdot}$ and (ii) is due to the subexponential property of $\norm{X_i}$ and $C$ is a constant. Meanwhile, we have
\begin{align*}
    \norm{\sum_{i=1}^{n} \E{(Y_i-\E{Y_i}) (Y_i-\E{Y_i})^{T}}} 
    &= \norm{\sum_{i=1}^{n} \E{Y_iY_i^{T}} - \E{Y_i}\E{Y_i}^{T}}\\
    &\overset{(i)}{\leq} \norm{\sum_{i=1}^{n} \E{Y_iY_i^{T}}}\\
    &= \norm{\sum_{i=1}^{n} \E{X_iX_i^{T}} - \E{X_iX_i^{T}\1{\norm{X_i}>B}}}\\
    &\overset{(ii)}{\leq} \norm{\sum_{i=1}^{n} \E{X_iX_i^{T}}} \leq V
\end{align*}
where (i) is due to the positive-semidefinite property of $\E{Y_i}\E{Y_i}^{T}$ and $\E{Y_iY_i^{T}} - \E{Y_i}\E{Y_i}^{T}$, (ii) is due to the positive-semidefinite property of $\E{X_iX_i^{T}\1{\norm{X_i}>B}}$ and $\E{Y_iY_i^{T}}.$ Similarly, $\norm{\sum_{i=1}^{n} \E{(Y_i-\E{Y_i})^{T} (Y_i-\E{Y_i})}} \leq V.$

Then, by Theorem 6.1.1 \cite{tropp2015introduction}, we have
\begin{align*}
    \Prob{\norm{\sum_{i=1}^{N}(Y_i - \E{Y_i})} \geq t} \leq 2\exp\left(-\frac{t^2/2}{V + 2Bt/3}\right).
\end{align*}
Then, with probability $1-O(n^{-c})$ for some constant $c$, 
\begin{align*}
    \norm{\sum_{i=1}^{N}(Y_i - \E{Y_i})} \lesssim \sqrt{V\log(n(m_1+m_2))} + B\log(n(m_1+m_2)).
\end{align*}
Take $B = L\log(n)C'$ for a proper constant $C'$, by \cref{eq:Ex-Yi-bound}, we have
\begin{align*}
    \norm{\sum_{i=1}^{N}Y_i} 
    &\lesssim \sqrt{V\log(n)} + L\log^2(n) + nL\log(n)O(n^{-C'/C})\\ 
    &\lesssim \sqrt{V\log(n(m_1+m_2))} + L\log(n(m_1+m_2))\log(n).
\end{align*}
By the union bound on the event $\norm{X_i} \leq B$ for all $i$, we can conclude that, with probability $1-O(n^{-c'})$ for some constant $c'$, 
\begin{align*}
    \norm{\sum_{i=1}^{N}X_i} \lesssim  \sqrt{V\log(n(m_1+m_2))} + L\log(n(m_1+m_2))\log(n).
\end{align*}
\end{proof}
%
%
%
Consider the Assumption \hyperlink{assum:A4}{A4}.
\begin{lemma}\label{lem:A4}
For any $m \in [n]$ and any $W \in \R^{n \times r}$, the following holds
\begin{align*}
\Prob{\norm{(\bar{X}-t\bar{M}^{*})_{m,\cdot}W}_2 \leq C\Delta^{*}\norm{W}_{2,\infty}\bar{\phi}\left(\frac{\norm{W}_{\F}}{\sqrt{n}\norm{W}_{2,\infty}}\right)} \geq 1 - O(n^{-3})
\end{align*}
where $C$ is a constant.
\end{lemma}

\begin{proof}{Proof.}
Let $Y_{j} = \bar{X}_{ij}-t\bar{M}_{ij}^{*}$ and $Z_j = Y_j W_{j,\cdot} \in \R^{1\times r}$. Note that 
\begin{align*}
    \norm{\left(\bar{X}-t\bar{M}^{*}\right)_{m,\cdot}W}_2 = \norm{\sum_{j=1}^{n} Z_j}_2. 
\end{align*}
We aim to invoke the \cref{lem:bernstein} for $Z_1, Z_2, \dotsc, Z_{n}$. Note that $\E{Z_j}=0$ since $\E{Y_j} = 0$ and $Z_j$ are independent since $Y_{j}$ are independent. Also, for the subexponential norm, we have
\begin{align*}
\norm{\norm{Z_j}_2}_{\psi_1} 
&\leq \norm{|Y_j|}_{\psi_1} \norm{W_{j,\cdot}}_2\\
&\leq \norm{|Y_j|}_{\psi_1} \norm{W}_{2,\infty}\\
&\lesssim L \norm{W}_{2,\infty}
\end{align*}
where the last inequality is due to \cref{lem:element-tail}. Then, one can check
\begin{align*}
\norm{\sum_{j=1}^{n}\E{Z_j^{T}Z_j}}
&\leq \sum_{j=1}^{n} \norm{\E{Z_j^{T}Z_j}}\\
&\leq \sum_{j=1}^{n} \E{Y_j^2} \norm{W_{j,\cdot}}_2^{2}\\
&\lesssim \sum_{j=1}^{n} L^2 p_{\O}\norm{W_{j,\cdot}}_2^{2}\\
&\lesssim L^2 p_{\O} \norm{W}_{\F}^2.
\end{align*}
Similarly, one can show that $\norm{\sum_{j=1}^{n}\E{Z_j^{T}Z_j}} \lesssim L^2 p_{\O} \norm{W}_{\F}^2.$ Then, we can invoke \cref{lem:bernstein} and obtain, with probability $1-O(n^{-3})$,
\begin{align*}
 \norm{\left(\bar{X}-t\bar{M}^{*}\right)_{m,\cdot}W}_2 \lesssim L\sqrt{p_{\O}} \norm{W}_{\F} \sqrt{\log(n)} + L\norm{W}_{2,\infty}\log^2(n).
\end{align*}
Since $\bar{\phi}(x) = \sqrt{\log(n)}L \frac{\sqrt{n p_{\O}}}{\Delta^{*}} x + L\frac{\log^2 n}{\Delta^{*}}$, we have
\begin{align*}
	L\sqrt{p_{\O}} \norm{W}_{\F} \sqrt{\log(n)} + L\norm{W}_{2,\infty}\log^2(n) \lesssim \Delta^{*}\norm{W}_{2,\infty}\bar{\phi}\left(\frac{\norm{W}_{\F}}{\sqrt{n}\norm{W}_{2,\infty}}\right). 
\end{align*}
This finishes the proof. 
\end{proof}

After showing that Assumptions \hyperref[assum:A1]{A1}--\hyperref[assum:A4]{A4} hold, we can prove the following theorem.
\begin{proposition}\label{thm:symmetric} Let $t := (g(\alpha^{*})p_{\A}^{*}+(1-p_{\A}^{*}))p_\O.$
Suppose $p_{\O} \geq \frac{\log^3 n}{n}$ and $\sqrt{n p_{\O} \log(n)}L \kappa^2 \leq Ct\lambda_1^{*}$ for some known constant $C$. Then, with probability $1-O(n^{-2})$, the following holds
\begin{align*}
    \norm{\bar{U}}_{2,\infty} &\lesssim \kappa \norm{\bar{U}^{*}}_{2,\infty} + \frac{\sqrt{np_{\O}}\kappa L}{\lambda_1^{*}t} \norm{\bar{M}^{*}}_{2,\infty}/\lambda_1^{*}\\
    \norm{\bar{U}\sgn(H) - \bar{U}^{*}}_{2,\infty} &\lesssim \frac{\sqrt{n p_{\O} \log(n)}}{t\lambda_1^{*}} \kappa^3L\norm{\bar{U}^{*}}_{2,\infty} + \frac{\sqrt{np_{\O}}\kappa L}{\lambda_1^{*}t} \norm{\bar{M}^{*}}_{2,\infty} /\lambda_1^{*}.
\end{align*}
\end{proposition}
\begin{proof}{Proof.}
Let $\gamma = (C_1+1)\bar{\gamma}, \phi(x) = C_2\bar{\phi}(x)$ where $C_1, C_2$ are constants defined in \cref{lem:A2,lem:A4} respectively. One can verify that $\gamma = (C_1+1)\frac{\sqrt{np_{\O}}\kappa L}{\lambda_1^{*}t}, \phi(x)= C_2 \frac{(\sqrt{np_{\O}\log(n)} x + \log^2 n)\kappa L}{\lambda_1^{*}t}$. In order to apply \cref{lem:generalize-entrywise-bound}, we still need to show that $32\kappa \max(\gamma, \phi(\gamma)) \leq 1.$ Because $p_{\O} \geq \frac{\log^3 n}{n}$ and $\sqrt{n p_{\O} \log(n)}L \kappa^2 \leq Ct\lambda_1^{*}$, one can verify that $32\kappa \max(\gamma, \phi(\gamma)) \leq 1$ by choosing a sufficient small $C$. Based on \cref{lem:A2}, \cref{lem:A4}, we can apply \cref{lem:generalize-entrywise-bound} and obtain that, with probability $1-O(n^{-2})$,
\begin{align*}
    \norm{\bar{U}}_{2,\infty} &\lesssim (\kappa+\phi(1)) \norm{\bar{U}^{*}}_{2,\infty} + \gamma \norm{t\bar{M}^{*}}_{2,\infty}/\Delta^{*}\\
    \norm{\bar{U}\sgn(\bar{H}) - \bar{U}^{*}}_{2,\infty} &\lesssim (\kappa(\kappa+\phi(1))(\gamma+\phi(\gamma)) + \phi(1)) \norm{\bar{U}^{*}}_{2,\infty} + \gamma \norm{t\bar{M}^{*}}_{2,\infty}/\Delta^{*}.
\end{align*}
Using the fact $\Delta^{*} = t\lambda_1^{*}/\kappa, \phi(1) \leq 1 \leq \kappa$, we have
\begin{align*}
    \norm{\bar{U}}_{2,\infty} &\lesssim  \kappa \norm{\bar{U}^{*}}_{2,\infty} + \gamma \norm{\bar{M}^{*}}_{2,\infty} / \lambda_1^{*}\\
    \norm{\bar{U}\sgn(\bar{H}) - \bar{U}^{*}}_{2,\infty} &\lesssim (\kappa^2(\gamma+\phi(\gamma)) + \phi(1)) \norm{\bar{U}^{*}}_{2,\infty} + \gamma \norm{\bar{M}^{*}}_{2,\infty} / \lambda_1^{*}.
    \end{align*}
Plug in the definition of $\gamma$ and $\phi$, we complete the proof. 
\end{proof}

\vspace{1em}
\subsection{Asymmetric Case}
Let $X_{\Omega}$ associated with $M^{*}, p_{\A}^{*}, \alpha^{*}, p_{\O}$ be the observation generated by the model described in the \cref{sec:model}. Let $t = (p_{\A}^{*} g(\alpha^{*}) + (1-p_{\A}^{*}))p_\O.$ Let $M^{*} = U^{*}\Sigma^{*}V^{*T}, M = U \Sigma V^{T}$ be the singular decomposition of $M^{*}$ and $M$, where $M = \arg\min_{{\rm rank}(M')\leq r} \norm{X'-M'}_{\F}$ and $X'$ is obtained from $X_{\Omega}$ by setting unobserved entries to 0. We construct the following: $\bar{M}^{*} := \begin{pmatrix} 0_{n\times n} & M^{*}\\ M^{*T} & 0_{m\times m} \end{pmatrix}.$
One can verify that the spectral decomposition of $\bar{M}^{*}$ is 
\begin{align*}
    \bar{M}^{*} = \frac{1}{\sqrt{2}} \begin{pmatrix} U^{*} & U^{*} \\ V^{*} & -V^{*} \end{pmatrix} \cdot \begin{pmatrix} \Sigma^{*} & \\ & -\Sigma^{*} \end{pmatrix} \cdot \frac{1}{\sqrt{2}} 
   \begin{pmatrix} U^{*} & U^{*} \\ V^{*} & -V^{*} \end{pmatrix}^{T}.
\end{align*}
Note that the largest $r$ singular values, $\sigma_1^{*}\geq \sigma_2^{*} \geq \dotsc \geq \sigma_{r}^{*}$, of $M^{*}$ are the same as the largest $r$ eigenvalues of $\bar{M}^{*}$. The $(r+1)$-th eigenvalue of $\bar{M}^{*}$ is non-positive. 
Let $\bar{U}^{*} = \frac{1}{\sqrt{2}} \begin{pmatrix} U^{*} \\ V^{*} \end{pmatrix}$ be the eigenvectors associated with the largest $r$ singular values of $\bar{M}^{*}.$ 
Similarly, let $\bar{X} := \begin{pmatrix} 0_{n\times n} & X\\ X^{T} & 0_{m\times m} \end{pmatrix}.$ Let $\bar{U} = \frac{1}{\sqrt{2}} \begin{pmatrix} U \\ V \end{pmatrix}$ be the eigenvectors associated with the largest $r$ singular values of $\bar{X}.$

We can apply \cref{thm:symmetric} to the $\bar{M}^{*}$ and $\bar{X}$ constructed in this subsection. This gives us the following result. 
\begin{proposition} \label{prop:entry-wise-bound}
Let $H = \frac{1}{2}(U^{T}U^{*} + V^{T}V^{*}), N = n + m, \mu = \max\left(N\norm{U^{*}}_{2,\infty}^2, N\norm{V^{*}}_{2,\infty}^2\right)/r, \kappa = \sigma_1^{*}/\sigma_r^{*}, t = (p_{\A}^{*}g(\alpha^{*}) + 1-p_{\A}^{*})p_\O$. Suppose  $p_{\O} \geq \frac{\log^3 m}{m}$ and $\sqrt{m p_{\O} \log(m)}L \kappa^2 \leq Ct\sigma_1^{*}$ for some known small constant $C$. Then, with probability $1-O((nm)^{-1})$, the followings hold
\begin{align}
(\norm{U}_{2,\infty} \vee \norm{V}_{2, \infty}) &\lesssim  \kappa\sqrt{\frac{\mu r}{m}} \label{eq:bound1}\\
    (\norm{U\sgn(H) - U^{*}}_{2,\infty} \vee \norm{V\sgn(H) - V^{*}}_{2, \infty}) &\lesssim   \frac{\sqrt{p_{\O} \log(m)}\kappa^3 L \sqrt{\mu r}}{t\sigma_1^{*}}\label{eq:bound2} \\
    \norm{M' - t M^{*}}_{\max} &\lesssim \kappa^4 \mu r L \sqrt{\frac{p_{\O}\log(m)}{m}}\label{eq:bound3}.
\end{align}
\end{proposition}
\begin{proof}{Proof.}
%
Note that $\sqrt{2} \norm{\bar{U}^{*}}_{2,\infty} = (\norm{U^{*}}_{2,\infty} \vee \norm{V^{*}}_{2, \infty}) \leq \sqrt{\mu r / N}.$  Furthermore, we have
\begin{align*}
    \norm{\bar{M}^{*}}_{2,\infty} &= \norm{M^{*}}_{2,\infty} \leq \norm{U^{*}}_{2,\infty}\norm{\Sigma^{*}V^{*}}_2 \leq \norm{U^{*}}_{2,\infty} \sigma_1^{*} \leq \sqrt{\mu r / N}\sigma_1^{*}.
\end{align*}
Apply \cref{thm:symmetric} on $\bar{M}^{*}$ and $\bar{X}$ along with the bound on $ \norm{\bar{U}^{*}}_{2,\infty},  \norm{\bar{M}^{*}}_{2,\infty}$, we can obtain, with probability $1-O(N^{-2})$,
\begin{align}
    \norm{\bar{U}}_{2,\infty} &\lesssim \kappa \sqrt{\frac{\mu r}{N}} \label{eq:bound1-in-proof}\\
    \norm{\bar{U}\sgn(H) - \bar{U}^{*}}_{2,\infty} &\lesssim \frac{\sqrt{N p_{\O} \log(N)}}{t\sigma_1^{*}} \kappa^3L \sqrt{\mu r/N} = \frac{\sqrt{p_{\O} \log(N)}\kappa^3 L \sqrt{\mu r}}{t\sigma_1^{*}}.
   \label{eq:bound2-in-proof}
\end{align}
This completes the proof of \cref{eq:bound1} and \cref{eq:bound2}. Next we proceed to the proof of \cref{eq:bound3}.

Let $\tilde{U} = U\sgn(H), \tilde{V} = V\sgn(H), \tilde{\Sigma} = \sgn(H)^{T} \Sigma \sgn(H).$ Note that $M'_{ij} = U_{i,\cdot} \Sigma V_{j,\cdot}^{T} = \tilde{U}_{i, \cdot} \tilde{\Sigma} \tilde{V}_{j,\cdot}^{T}$ and $M^{*}_{ij} = U^{*}_{i,\cdot} \Sigma^{*} V^{*T}_{j,\cdot}.$ Then, 
\begin{align}
    |M'_{ij} - t M^{*}_{ij}| 
    &= |\tr(\tilde{U}_{i, \cdot} \tilde{\Sigma} \tilde{V}_{j,\cdot}^{T}) - t \tr(U^{*}_{i,\cdot} \Sigma^{*} V^{*T}_{j,\cdot})| \nonumber\\
    &= |\tr(\tilde{\Sigma} \tilde{V}_{j,\cdot}^{T} \tilde{U}_{i, \cdot} ) - t \tr(\Sigma^{*} V^{*T}_{j,\cdot}U^{*}_{i,\cdot})| \nonumber\\
    &= |\tr((\tilde{\Sigma} - t\Sigma^{*})( \tilde{V}_{j,\cdot}^{T} \tilde{U}_{i, \cdot})) + \tr(t\Sigma^{*}(\tilde{V}_{j,\cdot}^{T} \tilde{U}_{i, \cdot}  - V^{*T}_{j,\cdot}U^{*}_{i,\cdot}))|\nonumber\\
    &\leq \norm{\tilde{\Sigma} - t\Sigma^{*}}_2 \norm{\tilde{V}_{j,\cdot}^{T} \tilde{U}_{i, \cdot}}_{*} + t\norm{\Sigma^{*}}_2 \norm{\tilde{V}_{j,\cdot}^{T} \tilde{U}_{i, \cdot}  - V^{*T}_{j,\cdot}U^{*}_{i,\cdot}}_{*} \label{eq:nuclear-norm}
\end{align}
where \cref{eq:nuclear-norm} is due to the triangle inequality and $|\tr(AB)| \leq \norm{A}_2 \norm{B}_{*}$ by the Von Neumann's trace inequality. We derive the bound on the term $\norm{\tilde{V}_{j,\cdot}^{T} \tilde{U}_{i, \cdot}  - V^{*T}_{j,\cdot}U^{*}_{i,\cdot}}_{*}$. Let $\hat{\gamma} = \kappa \sqrt{n p_{\O}} L / (\sigma_1^{*} t)$. Note that
\begin{align}
    \norm{\tilde{V}_{j,\cdot}^{T} \tilde{U}_{i, \cdot}  - V^{*T}_{j,\cdot}U^{*}_{i,\cdot}}_{*} &= \norm{(\tilde{V}_{j,\cdot}^{T} - V^{*T}_{j, \cdot})\tilde{U}_{i, \cdot} + V^{*T}_{j,\cdot} (\tilde{U}_{i, \cdot} - U^{*}_{i,\cdot})}_{*}\nonumber\\
    &\leq \norm{(\tilde{V}_{j,\cdot}^{T} - V^{*T}_{j, \cdot})\tilde{U}_{i, \cdot}}_{*} + \norm{V^{*T}_{j,\cdot} (\tilde{U}_{i, \cdot} - U^{*}_{i,\cdot})}_{*} \nonumber\\
    &\leq \norm{\tilde{V}_{j,\cdot}^{T} - V^{*T}_{j, \cdot}}_{2} \norm{\tilde{U}_{i, \cdot}}_{2} + \norm{V^{*T}_{j,\cdot}}_{2} \norm{\tilde{U}_{i, \cdot} - U^{*}_{i,\cdot}}_2 \label{eq:nuclear-two}\\
    &\lesssim \kappa^2\sqrt{\log(N)} \hat{\gamma} \sqrt{\mu r/N} \left(\norm{\tilde{U}_{i, \cdot}}_{2} + \norm{V^{*}_{j,\cdot}}_{2} \right) \label{eq:use-bound2}\\
    &\lesssim  \kappa^3\sqrt{\log(N)} \hat{\gamma} \mu r /N \label{eq:use-bound1}
\end{align}
where \cref{eq:nuclear-two} is due to $\norm{ab^{T}}_{*} = \norm{ab^{T}}_{2} \leq \norm{a}_2\norm{b}_2$ for any vector $a, b$, \cref{eq:use-bound2} is due to \cref{eq:bound2-in-proof}, and \cref{eq:use-bound1} is due to \cref{eq:bound1-in-proof}. We then bound $\norm{\tilde{V}_{j,\cdot}^{T} \tilde{U}_{i, \cdot}}_{*}$,
\begin{align}
    \norm{\tilde{V}_{j,\cdot}^{T} \tilde{U}_{i, \cdot}}_{*} &\leq \norm{V^{*T}_{j,\cdot}U^{*}_{i,\cdot}}_{*} + \norm{\tilde{V}_{j,\cdot}^{T} \tilde{U}_{i, \cdot}  - V^{*T}_{j,\cdot}U^{*}_{i,\cdot}}_{*}\nonumber \\
    &\lesssim \norm{V^{*T}_{j,\cdot}U^{*}_{i,\cdot}}_{*} +  \kappa^3\sqrt{\log(N)} \hat{\gamma} \frac{\mu r}{N}\label{eq:use-first-bound}\\
    &\lesssim \norm{V^{*}}_{2,\infty} \norm{U^{*}}_{2, \infty} + \kappa^3\sqrt{\log(N)} \hat{\gamma} \frac{\mu r}{N}\nonumber\\
    &\lesssim \frac{\mu r}{N} + \kappa^3\sqrt{\log(N)} \hat{\gamma} \frac{\mu r}{N}\nonumber\\
    &\lesssim \kappa^2\frac{\mu r}{N} \label{eq:final-VU-bound}
\end{align}
where \cref{eq:use-first-bound} is due to \cref{eq:use-bound1} and \cref{eq:final-VU-bound} is due to $\kappa \sqrt{\log(N)} \hat{\gamma} \lesssim 1$. Next we bound $\norm{\tilde{\Sigma}- t\Sigma^{*}}_{2}$. Note that
\begin{align}
    \norm{\tilde{\Sigma} - \Sigma}_2 &= \norm{\sgn(H)^{T}(\Sigma \sgn(H) - \sgn(H)\Sigma)}_2\nonumber\\
    &\leq \norm{\Sigma \sgn(H) - \sgn(H) \Sigma}_2 \nonumber\\
    &= \norm{(\Sigma H - H\Sigma) + \Sigma(\sgn(H)-H) + (H - \sgn(H))\Sigma}_2\nonumber\\
    &\leq \norm{\Sigma H - H \Sigma}_2 + 2\norm{\Sigma}_2\norm{\sgn(H) - H}_2.\label{eq:bound-for-tilde}
\end{align}

By Lemma 2 in \cite{abbe2017entrywise}, we have 
\begin{align}
    \norm{\sgn(H) - H}_2 \lesssim (\norm{\bar{X} - t\bar{M}^{*}}_2 / (t\sigma_r^{*}))^2 \lesssim \hat{\gamma}^2 \label{eq:sgnH-bound}\\
    \norm{\Sigma H - H \Sigma}_2 \leq 2\norm{\bar{X} - t\bar{M}^{*}}_2 \lesssim t\hat{\gamma} \sigma_r^{*} \label{eq:sigmaH-bound}
\end{align}
where $\norm{\bar{X} - t\bar{M}^{*}}_2 \leq \hat{\gamma}t\sigma_r^{*}$ by \cref{lem:A2}. By Weyl's inequality, we also have
$    \norm{\Sigma - t\Sigma^{*}}_2 \leq \norm{\bar{X} - t\bar{M}^{*}}_2 \lesssim t\hat{\gamma} \sigma_r^{*}.
$ Hence, 
\begin{align}
    \norm{\Sigma}_2 &\leq \norm{t\Sigma^{*}}_2 + \norm{\Sigma - t\Sigma^{*}}_2 \lesssim t\sigma_1^{*} + t\hat{\gamma} \sigma_r^{*} \lesssim t\sigma_1^{*}.\label{eq:two-norm-bound}
\end{align}
Plugging \cref{eq:sgnH-bound,eq:sigmaH-bound,eq:two-norm-bound} into \cref{eq:bound-for-tilde}, we have
$        \norm{\tilde{\Sigma} - \Sigma}_2 \lesssim t\hat{\gamma} \sigma_r^{*} + t\hat{\gamma}^2\sigma_1^{*} \lesssim t\hat{\gamma}\sigma_1^{*}.
$
Therefore,
\begin{align}
    \norm{\tilde{\Sigma}- t\Sigma^{*}}_{2} \leq \norm{\tilde{\Sigma} - \Sigma}_2 + \norm{\Sigma - t\Sigma^{*}}_2 \lesssim t\hat{\gamma} \sigma_1^{*}. \label{eq:tilde-star}
\end{align}

Plugging \cref{eq:use-bound1,eq:final-VU-bound,eq:tilde-star} into \cref{eq:nuclear-norm}, we arrive at 
\begin{align*}
    \norm{M^{'}-tM^{*}}_{\max}
    &\lesssim t\kappa^3\sqrt{\log(N)} \hat{\gamma} \frac{\mu r}{N} \sigma_1^{*}
    \lesssim t\kappa^3\sqrt{\log(N)} \frac{\mu r}{N} \sigma_1^{*} \frac{\kappa L\sqrt{N p_{\O}}}{t\sigma_1^{*}}
    \lesssim \kappa^4 \mu r L \sqrt{\frac{p_{\O}\log(N)}{N}}.
\end{align*}
\end{proof}

Next, we provide a lemma for the concentration bound of the sum over $\Omega$. 
\begin{lemma}\label{lem:concentration-bound}
Let $\Omega = \{(i, j) | O_{ij} = 1\} \subset [n]\times [m]$ where $O_{ij} \sim \Ber(p_{\O})$ are i.i.d random variables. Let $\{T_{ij} \in [0, 1] | (i,j) \in [n]\times[m]\}$ be independent random variables with $\E{T_{ij}} = p_{ij}$. Let $S = \sum_{(i, j) \in [n]\times [m]} p_{\O}p_{ij}$. Then, with probability $1-1/(nm)$,
\begin{align*}    
\left|\sum_{(i, j) \in \Omega} T_{ij} - S \right| \leq C\left(\sqrt{S \log (mn)} + \log(mn)\right)
\end{align*}
where $C$ is a constant. In particular, if $S \gtrsim \log(nm)$, then
\begin{align*}
\left|\sum_{(i, j) \in \Omega} T_{ij} - S \right| \leq C_1 S
\end{align*}
where $C_1$ is a constant.
\end{lemma}
\begin{proof}{Proof.}
Let $Z_{ij} = T_{ij}O_{ij}.$ Then $Z_{ij} \in [0, 1], \E{Z_{ij}} = p_{\O}p_{ij}.$ By the Bernstein's inequality \cite{bernstein1946theory}, we have
\begin{align*}
    \Prob{\left|\sum_{ij} Z_{ij} - \sum_{ij} p_{\O}p_{ij}\right| > t} \leq 2e^{-\frac{t^2/2}{\sum_{ij} \E{(Z_{ij}-p_{\O}p_{ij})^2} + \frac{t}{3}}}
    \leq 2e^{-\frac{t^2/2}{S + \frac{t}{3}}}
\end{align*}
due to 
\begin{align*}
\E{(Z_{ij}-p_{\O}p_{ij})^2} \leq \E{Z_{ij}^2} \leq  \E{Z_{ij}} = p_{\O} p_{ij}.
\end{align*} 
Take $t=C_2\left(\sqrt{S \log(nm)}+\log(nm)\right)$ where $C_2$ is a constant. Then we have 
\begin{align*}
\Prob{\left|\sum_{(i, j) \in \Omega} T_{ij} - \sum_{(i,j)\in[n]\times[m]} S \right|>t} \leq \frac{1}{nm}
\end{align*} 
for a proper $C_2$.
\end{proof}

\textbf{Proof of \cref{thm:entrywise-bound}:}

Next we proceed to the proof of  \cref{thm:entrywise-bound} based on \cref{prop:entry-wise-bound}. By the assumption in \cref{sec:model}, $\log^{1.5}(m) \mu r L \kappa^2 / (\norm{M^{*}}_{\max} \sqrt{m}) \lesssim \sqrt{p_{\O}}$ and $1-p_{\A}^{*} \gtrsim 1$. Note that $\norm{M^{*}}_{\max} \lesssim \sigma_1^{*}\mu r / m$, this implies that $\sqrt{\log(m)}\sqrt{m} L \kappa^2 \lesssim \sqrt{p_{\O}}\sigma_1^{*}$ and $\sqrt{p_{\O}} \gtrsim \frac{\log^{1.5}(m)}{\sqrt{m}}$, which is the condition required by \cref{prop:entry-wise-bound}. Also, by taking $T_{ij}=1$ in \cref{lem:concentration-bound} and noting that $p_{\O} \gtrsim \frac{\log^3(m)}{n}$, we have, with probability $1-O(\frac{1}{nm})$,  
\begin{align*}
\left| nmp_\O - |\Omega| \right| < C \sqrt{\log(nm)p_{\O}nm} 
\end{align*}
where $C$ is a constant. Then 
\begin{align*}
\left| \frac{nm}{|\Omega|} - \frac{1}{p_{\O}} \right| 
&= \frac{|nmp_\O - |\Omega||}{|\Omega| p_{\O}} \\
&\leq \frac{C\sqrt{\log(nm)p_{\O}nm}}{|\Omega|p_{\O}}\\
&\leq \frac{C'\sqrt{\log(nm)}}{\sqrt{p_{\O}nm}p_{\O}}
\end{align*}
where $C'$ is a constant. Finally, we can obtain
\begin{align}
\norm{M^{'} \frac{nm}{|\Omega|} - \frac{t}{p_{\O}} M^{*}}_{\max}
&= \norm{M^{'} \frac{nm}{|\Omega|} - M^{'}\frac{1}{p_{\O}} + M^{'}\frac{1}{p_{\O}} - \frac{t}{p_{\O}} M^{*}}_{\max} \nonumber\\
&\lesssim \norm{\frac{1}{p_{\O}} (M^{'} - tM^{*})}_{\max} + \norm{M^{'}}_{\max}\frac{\sqrt{\log(nm)/p_{\O}nm}}{p_{\O}}\nonumber\\
 &\lesssim \frac{\kappa^4 \mu r L}{p_{\O}} \sqrt{\frac{\log(m)p_{\O}}{m}} +  L\sqrt{\frac{\log(nm)}{p_{\O}nm}} \nonumber \\
 &\lesssim \kappa^4 \mu r L \sqrt{\frac{\log(m)}{p_{\O}m}}.
\end{align}
This completes the proof.

\vspace{1em}
\section{Analysis of $\pi^{\AD}$ and Proof of Theorem \ref{thm:main-theorem}}\label{sec:main-theorem-appendix}
To prove \cref{thm:main-theorem}, it is sufficient to prove \cref{lem:step2-F,lem:step2,lem:point-estimate}.
\vspace{1em}
\subsection{Moment Matching Estimator (Proof of Lemma \ref{lem:step2-F} and \ref{lem:step2})}\label{sec:moment-matching-appendix} 
\subsubsection{Proof of Lemma \ref{lem:step2-F}}
Recall that 
\begin{align*}
g_t(\theta, M) =\frac{1}{nm}\sum_{(i, j)\in[n]\times [m]} \left( p_\A\Probx{\Ano}{X_{ij}\leq t|\alpha, M_{ij}} + (1-p_\A)\Probx{\Pos}{X_{ij} \leq t|M_{ij}}\right).
\end{align*}
%

Let $\delta' = \kappa^4 \mu r L \sqrt{\frac{\log(m)}{p_{\O}m}}$ and 
\begin{align*}
h(\theta) = \sum_{t=0}^{T-1} \left(g_t(\theta, \hat{M}/e(\theta)) - \frac{|X_{ij}=t, (i,j)\in \Omega|}{|\Omega|}\right)^2.
\end{align*}
We have the following result.
\begin{lemma}\label{lem:smooth-on-M}
With probability $1-O(\frac{1}{nm})$, for any $\theta \in \Theta$ and $t = 0, 1, \dotsc, T$, 
\begin{align*} 
|g_t(\theta, M^{*}e(\theta^{*})/e(\theta)) - g_t(\theta, \hat{M}/e(\theta))| \lesssim (K+L)\delta'.
\end{align*} 
\end{lemma}
\begin{proof}{Proof.}
Note that $\Probx{\Ano}{X_{ij}=t|\alpha, M}$ is $K$-lipschitz on $M$. One also can verify that $\Probx{\Pos}{X_{ij}=t|M}$ is $L$-Lipschitz on $M$. Hence 
\begin{align*}
 \left( p_\A\Probx{\Ano}{X_{ij}\leq t|\alpha, M_{ij}} + (1-p_\A)\Probx{\Pos}{X_{ij} \leq t|M_{ij}}\right)
\end{align*} 
is $(K+L)$-Lipschitz on $M_{ij}$. Let $C_1, C_2$ be two constants. By \cref{thm:entrywise-bound}, with probability $1-O((nm)^{-1}),$ $|\hat{M}_{ij}/e(\theta^{*}) - M^{*}_{ij}| \leq C_1\delta'.$ This implies that
\begin{align*}
\left|\frac{M_{ij}^{*} e(\theta^{*})}{e(\theta)} - \frac{\hat{M}_{ij}}{e(\theta)}\right| \leq \frac{C_1\delta'}{e(\theta)} \leq C_2\delta'
\end{align*}
where we use that  $e(\theta) \geq (1-p_{A}) \geq c$ for some constant $c$. This implies that  
\begin{align*} 
\left|g_t\left(\theta, \frac{M^{*}e(\theta^{*})}{e(\theta)}\right) - g_t\left(\theta, \frac{\hat{M}}{e(\theta)}\right)\right| 
&\leq \frac{1}{nm} \sum_{ij} \left|\frac{M_{ij}^{*} e(\theta^{*})}{e(\theta)} - \frac{\hat{M}_{ij}}{e(\theta)}\right| (K+L)\\
&\lesssim (K+L)\delta'.
\end{align*}
\end{proof}

\begin{lemma}\label{lem:bound-theta-star}
With probability $1-O((nm)^{-1})$, $h(\theta^{*}) \lesssim (K+L)^2 (\delta')^2$.
\end{lemma}
\begin{proof}{Proof.}
Set $C_1, C_2, C_3, C_4, C_5$ be proper constants. 

Note that by \cref{lem:concentration-bound}, with probability $1-O((nm)^{-1})$, 
\begin{align*} 
\left||X_{ij}\leq t, (i,j)\in \Omega| - p_{\O} n m g_t(\theta^{*}, M^{*})\right|  \leq C_2 \sqrt{p_{\O} n m g_t(\theta^{*}, M^{*}) \log(nm)} + C_2\log(nm)
\end{align*} 
Also, we can similarly obtain $\left||\Omega| - p_{\O} nm\right| \leq C_3\sqrt{p_{\O} nm\log(nm)}$ by  \cref{lem:concentration-bound}. Then, one can verify that 
\begin{align*}
&\left|\frac{|X_{ij}=t, (i,j)\in \Omega|}{|\Omega|} - g_t(\theta^{*}, M^{*})\right| \\
&= \left|\frac{|X_{ij}=t, (i,j)\in \Omega|- |\Omega| g_t(\theta^{*}, M^{*})}{|\Omega|}\right| \\
&\leq \frac{1}{|\Omega|} \left( C_2 \sqrt{p_{\O} n m g_t(\theta^{*}, M^{*}) \log(nm)} + C_3\sqrt{p_{\O} nm\log(nm)} g_t(\theta^{*}, M^{*}) + C_2\log(nm)\right)\\
&\leq C_4 \frac{\sqrt{p_{\O}nm}\log(nm)}{nmp_\O}\\
&\leq C_4\frac{\log (nm)}{\sqrt{nmp_{\O}}}.
\end{align*}
Then, taking $\theta=\theta^{*}$ in \cref{lem:smooth-on-M}, we have 
\begin{align*}
    h(\theta^{*}) 
    &= \sum_{t=0}^{T} \left(g_t(\theta^{*}, \hat{M}/e(\theta^{*})) - |X_{ij}=t, (i,j)\in \Omega| /|\Omega|\right)^2\\
    &\leq \sum_{t=0}^{T} \left(|g_t(\theta^{*}, \hat{M}/e(\theta^{*})) - g_t(\theta^{*}, M^{*})| + C_5 \log(nm)/\sqrt{nmp_\O}\right)^2\\
    &\lesssim (K+L)^2 (\delta')^2 + \log(nm)\log(nm)/(nmp_{\O})\\
    &\lesssim (K+L)^2 (\delta')^2
\end{align*}
due to the fact that $\delta' \gtrsim \sqrt{\frac{\log(m)}{p_{\O}m}}$ and $p_{\O} \gtrsim \frac{\log^3(m)}{m}.$
\end{proof}

\textbf{Proof of \cref{lem:step2-F}:}

By \cref{lem:bound-theta-star}, with probability $1-O((nm)^{-1})$, $h(\hat{\theta}) \leq h(\theta^{*}) \lesssim (K+L)^2 (\delta')^2.$
This implies, for each $t<T$, $|g_t(\hat{\theta}, \hat{M}/e(\hat{\theta})) - \frac{|X_{ij}=t, (i,j)\in \Omega|}{|\Omega|}| \lesssim (K+L)\delta'.$ Combining with \cref{lem:bound-theta-star}, we have, for each $t<T$, 
\begin{align}\label{eq:gt-bound}
|g_t(\hat{\theta}, \hat{M}/e(\hat{\theta})) - g_t(\theta^{*}, \hat{M}/e(\theta^{*}))| \lesssim (K+L)\delta'. 
\end{align}
Note that 
\begin{align*}
    &|g_t(\theta^{*}, M^{*}) - g_t(\hat{\theta}, M^{*}e(\theta^{*})/e(\hat{\theta}))| \\
    &\leq |g_t(\theta^{*}, M^{*}) - g_t(\hat{\theta}, \hat{M}/e(\hat{\theta}))| + |g_t(\hat{\theta}, \hat{M}/e(\hat{\theta})) - g_t(\hat{\theta}, M^{*}e(\theta^{*})/e(\hat{\theta}))|. 
\end{align*}
By \cref{lem:smooth-on-M}, $ |g_t(\hat{\theta}, \hat{M}/e(\hat{\theta})) - g_t(\hat{\theta}, M^{*}e(\theta^{*})/e(\hat{\theta}))| \lesssim (K+L)\delta'.$ Also we have
\begin{align*}
 |g_t(\theta^{*}, M^{*}) - g_t(\hat{\theta}, \hat{M}/e(\hat{\theta}))|  \leq |g_t(\theta^{*}, M^{*}) - g_t(\theta^{*}, \hat{M}/e(\theta^{*}))| + |g_t(\hat{\theta}, \hat{M}/e(\hat{\theta})) - g_t(\theta^{*}, \hat{M}/e(\theta^{*}))|.
\end{align*}
By \cref{lem:smooth-on-M} agian, we have $|g_t(\theta^{*}, M^{*}) - g_t(\theta^{*}, \hat{M}/e(\theta^{*}))| \lesssim (K+L)\delta'.$ By \cref{eq:gt-bound}, we have $ |g_t(\hat{\theta}, \hat{M}/e(\hat{\theta})) - g_t(\theta^{*}, \hat{M}/e(\theta^{*}))| \lesssim (K+L)\delta'.$ In conclusion, 
\begin{align*}
|g_t(\theta^{*}, M^{*}) - g_t(\hat{\theta}, \hat{M}/e(\hat{\theta}))| \lesssim (K+L)\delta'.
\end{align*}

Therefore, $\norm{F(\hat{\theta}) - F(\theta^{*})} \lesssim (K+L)\delta'$ since $T$ is a constant.

\subsubsection{Proof of Lemma \ref{lem:step2}}
\begin{lemma}\label{lem:regularity-condtions}
Suppose $F$ satisfies the following condition:
\begin{itemize}
\item $F: \Theta \subset \R^{d_1} \rightarrow \R^{d_2}$ is  continuously differentiable and injective.
\item $B_{2C_2\delta}(\theta^{*}) \subset \Theta$ where $B_{r}(\theta^{*}) = \{\theta: \norm{\theta-\theta^{*}} \leq r\}.$
\item $\norm{J_{F}(\theta) - J_{F}(\theta^{*})}_{\max} \leq C_1 \norm{\theta-\theta^{*}}$ for $\theta \in B_{2C_2\delta}(\theta^{*}).$
\item $\norm{J_{F}(\theta^{*})^{-1}}_2 \leq C_2$.
\end{itemize} 
Suppose $2\sqrt{d_1d_2}C_1(C_2)^2 \delta < 1/2.$ For any $\theta \in \Theta$, 
\begin{align}
\norm{F(\theta)-F(\theta^{*})} \leq \delta \implies \norm{\theta-\theta^{*}} \leq 2C_2 \delta.
\end{align}
\end{lemma}
\begin{proof}{Proof.}
Suppose $\norm{F(\theta) - F(\theta^{*})} \leq \delta.$ We construct a sequence of $\theta_{i}$ such that $\lim_{i\rightarrow \infty}F(\theta_i) = F(\theta)$ while $\norm{\theta_i - \theta^{*}}$ is well bounded for every $i$. Let $\theta_1 - \theta^{*} = J_{F}^{-1}(\theta^{*})(F(\theta) - F(\theta^{*})).$ Note that 
\begin{align*}\norm{\theta_1 - \theta^{*}} \leq \norm{J_{F}^{-1}(\theta^{*})}_2 \norm{F(\theta)-F(\theta^{*})}
\leq C_2\delta.
\end{align*} Furthermore, by multivariate Taylor theorem, 
\begin{align*}
F(\theta_1) = F(\theta^{*}) + A (\theta_1-\theta^{*})^{T} 
\end{align*}
where the $i$-th row $A_i = (\nabla F_i (x_i))^{T}$ such that $x_i = \theta^{*} + c(\theta_1-\theta^{*})$ for some $c\in [0, 1].$ Hence, $F(\theta_1) = F(\theta^{*}) + J_F(\theta^{*}) (\theta_1-\theta^{*})^{T} + (A-J_{F}(\theta^{*}))(\theta_1-\theta^{*})^{T}.$ Note that $F(\theta^{*}) + J_F(\theta^{*}) (\theta_1-\theta^{*})^{T} = F(\theta)$ by the definition of $\theta_1$. Therefore,
\begin{align*}
&F(\theta_1) = F(\theta) + (A-J_F(\theta^{*}))(\theta_1-\theta^{*})^{T}\\
\implies& \norm{F(\theta_1) - F(\theta)} \leq  \norm{A-J_F(\theta^{*})}_{2} \norm{\theta_1-\theta^{*}}\\
\implies&\norm{F(\theta_1) - F(\theta)} \leq  \norm{A-J_F(\theta^{*})}_{\max} \sqrt{d_1d_2} \norm{\theta_1-\theta^{*}}\\
\implies&\norm{F(\theta_1) - F(\theta)} \leq  C_1\sqrt{d_1d_2} \norm{\theta_1-\theta^{*}}^2\\
\implies&\norm{F(\theta_1) - F(\theta)} \leq  C_1\sqrt{d_1d_2} C_2^2\delta^2.
\end{align*}
We can use the similar idea to the successive construction. In particular, let $t = 2\sqrt{d_1d_2}(C_1)(C_2)^2\delta < 1/2, a = 2C_1C_2\sqrt{d_1d_2}, \theta_0 = \theta^{*}.$ Suppose 
\begin{align*}
\norm{\theta_{k} - \theta_{k-1}} \leq \frac{1}{a} t^{k}, \norm{\theta_{k}-\theta^{*}} \leq \frac{1}{a}(2t - t^{k}), \norm{F(\theta_k)-F(\theta)} \leq \frac{1}{aC_2}t^{k+1}.
\end{align*} 
It is easy to verify that the above conditions are satisfied for $k=1.$ Then let $\theta_{k+1} - \theta_{k} = J_{F}^{-1}(\theta^{*})(F(\theta)-F(\theta_{k}))$ for $k>1.$

Then, we have $\norm{\theta_{k+1} - \theta_{k}} \leq C_2 \frac{t^{k+1}}{aC_2} \leq \frac{t^{k+1}}{a}.$ Also, $\norm{\theta_{k+1} - \theta^{*}} \leq \norm{\theta_{k+1}-\theta_{k}} + \norm{\theta_{k} - \theta^{*}} \leq \frac{2t - t^{k} + t^{k+1}}{a} \leq \frac{2t - t^{k+1}}{a}.$
Furthermore, 
\begin{align*}
&F(\theta_{k+1}) = F(\theta_{k}) + J_{F}(\theta^{*})(\theta_{k+1}-\theta_{k})^{T} + (A-J_F(\theta^{*}))(\theta_{k+1}-\theta_{k})^{T}\\
\implies& F(\theta_{k+1}) = F(\theta) + (A-J_F(\theta^{*}))(\theta_{k+1}-\theta_{k})^{T}\\
\implies& \norm{F(\theta_{k+1}) - F(\theta)} \leq  \norm{A-J_F(\theta^{*})}_{2} \norm{\theta_{k+1}-\theta_{k}}\\
\implies&\norm{F(\theta_{k+1}) - F(\theta)} \leq  \norm{A-J_F(\theta^{*})}_{\max} \sqrt{d_1d_2} \norm{\theta_{k+1}-\theta_{k}}\\
\implies&\norm{F(\theta_{k+1}) - F(\theta)} \leq  C_1\sqrt{d_1d_2}(\norm{\theta_{k}-\theta_{*}}+\norm{\theta_{k+1}-\theta_{k}})\norm{\theta_{k+1}-\theta_{k}}\\
\implies&\norm{F(\theta_{k+1}) - F(\theta)} \leq  C_1\sqrt{d_1d_2} \frac{2(t)}{a}\frac{t^{k+1}}{a}\\
\implies &\norm{F(\theta_{k+1}) - F(\theta)} \leq \frac{C_1C_2\sqrt{d_1d_2} 2}{a} \frac{t^{k+2}}{aC_2} \leq \frac{t^{k+2}}{aC_2}.
\end{align*}

Note that $\norm{\theta_{k} - \theta^{*}} \leq \frac{2t}{a}$. Therefore $\theta_{k} \in \Theta$ is well-defined. Furthermore, we can conclude for any $\epsilon>0$, there exists $N$, if $k_1,k_2>N$, $\norm{\theta_{k_1} - \theta_{k_2}} \leq \epsilon.$ Therefore, the sequence converges. Suppose $\lim_{k} \theta_{k} = \theta'.$  Note that $\norm{\theta'-\theta^{*}} \leq \frac{2t}{a}$ due to $\norm{\theta_{k} - \theta^{*}} \leq \frac{2t}{a}.$ Also note that $\norm{J_{F}(\theta)}$ is bounded for $\norm{\theta-\theta^{*}} \leq \frac{2t}{a}$. This implies that $\lim_{k} F(\theta_{k}) = F(\theta').$ On the other hand, due to the convergence of $F(\theta_{k})$, $\lim_{k} F(\theta_{k}) = F(\theta).$ By injectivity, $\theta = \theta'$ and $\norm{\theta-\theta^{*}} \leq 2t/a.$ This completes the proof.   
\end{proof}

\begin{proof}{\textbf{Proof of \cref{lem:step2}}.} \cref{lem:step2} is then a simple combination \cref{lem:step2-F} and \cref{lem:regularity-condtions}. In particular, \cref{lem:step2-F} implies $\norm{F(\theta)-F(\theta^{*})} \lesssim (K+L)\delta' $ and \cref{lem:regularity-condtions} implies $\norm{F(\theta)-F(\theta^{*})} \lesssim (K+L)\delta' \implies \norm{\theta-\theta^{*}} \lesssim (K+L)\delta'.$ This finishes the proof. 
\end{proof}

\vspace{1em}
\subsection{Plug-in estimator and Proof of Lemma \ref{lem:point-estimate}}
Let $\delta' = (K+L)\kappa^4 \mu r L \sqrt{\frac{\log(m)}{p_{\O}m}}.$ Let 
\begin{align*}
\hat{x}_{ij} &:= [\hat{p}_\A \mathbb{P}_{\Ano}(X_{ij}|\hat{\alpha}, \hat{M}_{ij}/e(\hat{\theta}))]\\ \hat{y}_{ij} &:= [(1-\hat{p}_\A)\mathbb{P}_{\Pos}(X_{ij}|\hat{M}_{ij}/e(\hat{\theta}))].
\end{align*} 
Let $x_{ij}=p_{\A}^{*}\Probx{\Ano}{X_{ij}|\alpha^{*},M^{*}_{ij}}, y_{ij}=(1-p_{\A}^{*})\Probx{\Pos}{X_{ij}|M_{ij}^{*}}$. We have the following result.
\begin{lemma}\label{lem:x-y-bound}
With probability $1-O(1/(nm)),$ $\max(|\hat{x}_{ij}-x_{ij}|, |\hat{y}_{ij}-y_{ij}|) \leq C(L+K)^2L\delta'$ for any $(i, j) \in \Omega.$
\end{lemma}
\begin{proof}{Proof.}

By \cref{lem:step2}, with probability $1-O(1/(nm)),$ we have $\norm{\hat{\theta}-\theta^{*}} \lesssim \delta'.$ 

Note that $g(\theta)$ is $K$-Lipschitz in $\theta$ and $e(\theta) = p_{\A}g(\theta) + (1-p_{\A}).$ Hence
\begin{align*}
|e(\hat{\theta}) - e(\theta^{*})| 
&\leq |\hat{p}_{\A}-p_{\A}^{*}|(1-g(\hat{\theta})) + p_{\A}^{*}|g(\hat{\theta}) - g(\theta^{*})|\\
&\lesssim (K+1)\delta'.
\end{align*} 
Furthermore
\begin{align*}
\left|\frac{\hat{M}_{ij}}{e(\hat{\theta})} - M^{*}\right| 
&= \frac{1}{e(\hat{\theta})} |\hat{M} - M^{*} e(\hat{\theta})| \\
&\leq \frac{1}{e(\hat{\theta})} \left( |\hat{M} - M^{*} e(\theta^{*})| + M^{*}|e(\theta^{*}) - e(\hat{\theta})| \right)\\
&\lesssim \frac{\delta'}{K+L} + L(K+1)\delta'\\
&\lesssim L(K+1)\delta'.
\end{align*}

Note that $\Probx{\Ano}{\alpha, M}$ is $K$-Lipschitz in $\alpha$ and $M$. The implies that
\begin{align*}
|\hat{x}_{ij} - x_{ij}| 
&\leq |\hat{p}_\A \Probx{\Ano}{X_{ij}|\hat{\alpha}, \hat{M}_{ij}} - p_{A}^{*}\Probx{\Ano}{X_{ij}|\alpha^{*},M^{*}_{ij}}|\\
&\leq |\hat{p}_{\A}-p_{\A}^{*}| \Probx{\Ano}{X_{ij}|\alpha^{*},M^{*}_{ij}} + |\Probx{\Ano}{X_{ij}|\alpha^{*},M^{*}_{ij}} - \Probx{\Ano}{X_{ij}|\hat{\alpha}, \hat{M}_{ij}}| \hat{p}_{\A}\\
&\lesssim \delta' + KL(K+1)\delta'\\
&\lesssim KL(K+1)\delta'.
\end{align*}

Similarly, one can obtain $|\hat{y}_{ij} - y_{ij}| \lesssim L^2(K+1) \delta'.$ In conclusion, 
\begin{align*}
\max(|\hat{x}_{ij} - x_{ij}|, |\hat{y}_{ij} - y_{ij}|) \lesssim (L+K)^2 L.
\end{align*}

\end{proof}

\begin{lemma}\label{lem:ratio-bound}
Suppose $|\hat{x}-x|\leq \delta, |\hat{y}-y| \leq \delta$ where $x,y, \hat{x}, \hat{y} \in [0, 1], x+y>0$. Let $\hat{s} = \frac{\hat{x}}{\hat{x}+\hat{y}}$ if $\hat{x}+\hat{y}>0$ otherwise $\hat{s} = 0$. Then,  
\begin{align}
    \left|\hat{s} - \frac{x}{x+y}\right| \leq \min\left(\frac{\delta}{x+y}, \frac{\delta}{\hat{x}+\hat{y}}, 1\right).
\end{align}
\end{lemma}
\begin{proof}{Proof.}
Note that $ \left|\hat{s} - \frac{x}{x+y}\right| \leq 1$ is trivial since $\hat{s} \in [0, 1]$ and $\frac{x}{x+y} \in [0 ,1]$. 

When $\hat{x}=\hat{y}=0$, $\frac{x}{x+y} \leq \frac{\delta}{x+y}$ due to $x \leq \delta.$

When $\hat{x}+\hat{y} > 0$, 
\begin{align*}
    \left|\hat{s} - \frac{x}{x+y}\right| 
    &= \left|\frac{\hat{x}(x+y)-x(\hat{x}+\hat{y})}{(\hat{x}+\hat{y})(x+y)}\right|\\
    &= \left|\frac{\hat{x}y-x\hat{y}}{(\hat{x}+\hat{y})(x+y)}\right|\\
    &= \left|\frac{\hat{x}(\hat{y}-(\hat{y}-y))-(\hat{x}-(\hat{x}-x))\hat{y}}{(\hat{x}+\hat{y})(x+y)}\right|\\
    &= \left|\frac{-\hat{x}(\hat{y}-y)+(\hat{x}-x)\hat{y}}{(\hat{x}+\hat{y})(x+y)}\right| \\
    &\leq \frac{\hat{x}}{\hat{x}+\hat{y}} \frac{|\hat{y}-y|}{x+y} + \frac{\hat{y}}{\hat{x}+\hat{y}} \frac{|\hat{x}-x|}{x+y}\\
    &\leq  \frac{\hat{x}}{\hat{x}+\hat{y}} \frac{\delta}{x+y} + 
    \frac{\hat{y}}{\hat{x}+\hat{y}} \frac{\delta}{x+y}\\
    &= \frac{\delta}{x+y}.
\end{align*}
By symmetry, $\left|\hat{s} - \frac{x}{x+y}\right|  \leq \frac{\delta}{\hat{x}+\hat{y}}$, which completes the proof.
\end{proof}
\begin{proof}{\textbf{Proof of \cref{lem:point-estimate}}}
The result of \cref{lem:point-estimate} is a simple corollary of \cref{lem:x-y-bound} and \cref{lem:ratio-bound}.
\end{proof}

\subsection{General cost model}\label{sec:general-cost}
In practice, one may also want to incorporate the rewards when $A_{ij}^{\pi}=1$ and $B_{ij}=1$ (the anomaly is detected correctly). In general, consider the costs (or rewards) associated with the following four scenarios:
\begin{itemize}
\item $c_{ij}^{(01)}:$ $A_{ij}^{\pi}=0$ and $B_{ij}=1.$
\item $c_{ij}^{(11)}:$ $A_{ij}^{\pi}=1$ and $B_{ij}=1.$
\item $c_{ij}^{(00)}:$ $A_{ij}^{\pi}=0$ and $B_{ij}=0.$
\item $c_{ij}^{(10)}:$ $A_{ij}^{\pi}=1$ and $B_{ij}=0.$
\end{itemize}
The costs of the algorithm in such a general model can be defined as
\begin{align*}
\mathrm{cost}^{\pi}(X_{\Omega}) := \frac{1}{|\Omega|}\E{\sum_{(i,j) \in \Omega} c_{ij} \Big| X_{\Omega}},
\end{align*}
where $c_{ij}$ is the cost incurred at entry $(i,j)$ given by
\begin{align*}
c_{ij} &:=   c_{ij}^{(10)} \1{A_{ij}^{\pi}=1, B_{ij} = 0}
 + c_{ij}^{(01)} \1{A_{ij}^{\pi}=0, B_{ij} = 1} \\
 &\quad + c_{ij}^{(00)} \1{A_{ij}^{\pi}=0, B_{ij} = 0}
 + c_{ij}^{(11)} \1{A_{ij}^{\pi}=1, B_{ij} = 1}.
\end{align*}

\noindent
\textbf{Reduction.} To minimize the cost in such a generalized model, one can simply let $c_{ij}^{(0)} := c_{ij}^{(10)} - c_{ij}^{(00)}$ and $c_{ij}^{(1)} := c_{ij}^{(01)} - c_{ij}^{(11)}$ and reduce it to the original model with only two types of costs. One can easily show that the solution provided by EW for such constructed $\{c_{ij}^{(0)}, c_{ij}^{(1)}\}$ will also achieve the (near) optimal regret $\tilde{O}(1/\sqrt{m})$ in the general model. The proof is simply by rewriting the cost formula and hence omitted for simplicity. 

\section{Proof of Proposition \ref{prop:lower-bound}}\label{sec:lower-bound-appendix}

We consider the following special model: let $p_{\O}=1$ and $p_{\A}^{*} = \frac{1}{2}$, and when $B_{ij}=1$, let $X_{ij}=0.$ We refer to this in notational form as $X\sim \mathrm{H}(M^{*})$.

We construct $\mathcal{M}_n =\{M^{b} \in \R^{n\times n}, b \in \{0,1\}^{n/2} \}$ as follows. Fix a constant $c^{*}\leq \frac{1}{2e}$. Consider $b \in \{0, 1\}^{n/2}$. For any $i \in [n/2], j \in [n]$,  if $b_{i} = 0$, $M^{b}_{2i,j} = 1$ and $M^{b}_{2i+1,j} = 1 - \frac{c^{*}}{\sqrt{n}};$ if $b_{i} = 1$, $M^{b}_{2i,j} = 1 - \frac{c^{*}}{\sqrt{n}}$ and $M^{b}_{2i+1,j} = 1.$ Let $M^{b}$ be drawn uniformly form $\mathcal{M}_{n}$ and $X \sim \mathrm{H}(M^{b}).$ For convenience, we let $M^{-} := 1-\frac{c^{*}}{\sqrt{n}}$ and $M^{+} := 1.$

We write $\sum_{(i,j)\in [n]\times [n]}$ as 
$\sum_{ij}$ if there is no ambiguity. Note that anomaly can only occur when $X_{ij} = 0.$ One can verify that
\begin{align*}
f_{ij}^{*} 
&= \frac{1/2 e^{-M_{ij}^{*}}}{1/2 + 1/2e^{-M_{ij}^{*}}} \1{X_{ij}=0} + \1{X_{ij} > 0}\\
&= \frac{e^{-M_{ij}^{*}}}{1+e^{-M_{ij}^{*}}} \1{X_{ij} = 0} + \1{X_{ij} > 0}\\
&= \frac{1}{e^{M_{ij}^{*}}+1} \1{X_{ij} = 0} + \1{X_{ij} > 0}
\end{align*} 

We set $c^{(1)}_{ij} = \frac{1}{2} \left(\frac{1}{e^{M^{+}}+1} + \frac{1}{e^{M^{-}}+1}\right), c^{(0)}_{ij} = 1 - c^{(1)}_{ij}.$ Then one can verify that the following policy $\pi^{*}$ with $\{t^{*}_{ij}\}$ is optimal: $t_{ij}^{*} = 1$ if $X_{ij} = 0$ and $M_{ij}^{*} = M^{+};$ otherwise $t_{ij}^{*} = 0.$ For any policy $\pi$, let $t_{ij}(X) := P(A_{ij}^{\pi}=1|X).$ 

Let $\delta = \frac{1}{2} (\frac{1}{e^{M^{-}}+1} - \frac{1}{e^{M^{+}}+1}).$ Then the regret of $\pi$ is 
\begin{align*}
n^2(\cost^{\pi}(X) - \cost^{\pi^{*}}(X)) 
&= 
\sum_{ij} (t_{ij}(X) - t_{ij}^{*}) (-c_{ij}^{(1)} + f_{ij}^{*}) \\
&\overset{(i)}{\geq} \sum_{ij} (t_{ij}(X) - t_{ij}^{*}) (-c_{ij}^{(1)} + f_{ij}^{*}) \1{X_{ij}=0}\\
&= \sum_{ij} ((t_{ij}(X) - 1) (-\delta) \1{M_{ij} = M^{+}}+ t_{ij}(X) \delta \1{M_{ij}=M^{-}}) \1{X_{ij}=0}.
\end{align*}
Here (i) is due to the non-negativity of $(t_{ij}(X) - t_{ij}^{*}) (-c_{ij}^{(1)} + f_{ij}^{*})$ (since $t_{ij}^{*}$ minimizes $t_{ij}^{*}(-c_{ij}^{(1)} + f_{ij}^{*}$)).

Next, let $p_{ij}(M) = E_{X|M} \left(t_{ij}(X)\1{X_{ij}=0}\right).$ Then
\begin{align*}
&E_{X|M}(n^2(\cost^{\pi}(X) - \cost^{\pi^{*}}(X))) \\
&\geq \delta \sum_{ij} \left\{\1{M_{ij} = M^{+}}P(X_{ij}=0) + p_{ij}(M)(\1{M_{ij}=M^{-}} - \1{M_{ij}=M^{+}}) \right\}\\
&= \delta \sum_{ij} \left\{\1{M_{ij} = M^{+}}e^{-M_{ij}} + p_{ij}(M)(\1{M_{ij}=M^{-}} - \1{M_{ij}=M^{+}}) \right\}
\end{align*}

This further implies
\begin{align}
&E_{M \sim \mathcal{M}_{n}}E_{X|M}(n^2(\cost^{\pi}(X) - \cost^{\pi^{*}}(X))) \nonumber \\
&\geq \frac{\delta}{2^{n/2}} \sum_{M \in \mathcal{M}_{n}} \sum_{ij} \left\{\1{M_{ij} = M^{+}}e^{-M_{ij}} + p_{ij}(M)(\1{M_{ij}=M^{-}} - \1{M_{ij}=M^{+}}) \right\} \nonumber \\
&= \frac{\delta}{2^{n/2}} \sum_{ij}\sum_{M \in \mathcal{M}_{n}}  \left\{\1{M_{ij} = M^{+}}e^{-M_{ij}} + p_{ij}(M)(\1{M_{ij}=M^{-}} - \1{M_{ij}=M^{+}}) \right\} \label{eq:E-M-bound}
\end{align}

Next, for any fixing $(i, j)$, let's consider two matrices $M^{a}, M^{b}$ where $a \in \{0, 1\}^{n/2}$ and $b \in \{0, 1\}^{n/2}$ are only different in the $\lfloor \frac{i}{2} \rfloor$-th bit ($a_{\lfloor \frac{i}{2} \rfloor} = 0, b_{\lfloor \frac{i}{2} \rfloor} = 1$). There are $2^{n/2-1}$ pair of such matrices in $\mathcal{M}_{n}.$ Then
\begin{align}
&\left\{\1{M_{ij}^a = M^{+}}e^{-M^a_{ij}} + p_{ij}(M^a)(\1{M^a_{ij}=M^{-}} - \1{M^a_{ij}=M^{+}}) \right\} \nonumber \\
&+ \left\{\1{M_{ij}^b = M^{+}}e^{-M^b_{ij}} + p_{ij}(M^b)(\1{M^b_{ij}=M^{-}} - \1{M^b_{ij}=M^{+}}) \right\} \nonumber\\
&\quad = e^{-M^{+}} + p_{ij}(M^{a}) - p_{ij}(M^{b}) \nonumber \\
&\quad \geq e^{-M^{+}} - |p_{ij}(M^{a}) - p_{ij}(M^{b})| \nonumber\\
&\quad \geq \frac{1}{2}  e^{-M^{+}} \label{eq:bound-Mij}
\end{align}
where is the last inequality is due to following lemma to bound $|p_{ij}(M^{a}) - p_{ij}(M^{b})|.$
\begin{lemma}
$$
 |p_{ij}(M^{a}) - p_{ij}(M^{b})| \leq \frac{1}{2} e^{-M^{+}}.
$$
\end{lemma}
\begin{proof}{Proof.}
Let $X(M)$ be $X\sim \mathrm{H}(M)$, $\delta(X || Y)$ be the total variation distance between $X$ and $Y$, $D_{\mathrm{KL}}(X || Y)$ be the KL-divergence between $X$ and $Y$. By definition of $p_{ij}(M)$, we have
\begin{align*}
 |p_{ij}(M^{a}) - p_{ij}(M^{b})| 
 &= \sum_{X} P(A_{ij}^{\pi}=1|X)\1{X_{ij}=0} \left(P(X|M^{a}) - P(X|M^{b})\right)\\
 &\leq \sum_{X} |P(X|M^{a}) - P(X|M^{b})| \\
 &= \delta(X(M^{a}) || X(M^{b})) & \text{total variation distance}\\
    &\leq \sqrt{\frac{1}{2} D_{KL}(X(M^{a}) || X(M^{b}))} &\text{Pinsker's inequality}\\
    &= \sqrt{\frac{1}{2} \sum_{ij} D_{KL}(X(M^{a})_{ij} || X(M^{b})_{ij})} &\text{$X_{ij}$ are independent}.
\end{align*}
Note that there are only two rows that are different between $M^{a}$ and $M^{b}$. 
Let $X^{+}$ be the observation of the entry with value $M^{+}$ and $X^{-}$ be the observation of the entry with the value $M^{-}.$ Then we have
\begin{align*}
\sum_{ij} D_{KL}(X(M^{a})_{ij} || X(M^{b})_{ij}) = n D_{KL}(X^{+} || X^{-}) + nD_{KL}(X^{-} || X^{+}).
\end{align*}
Note that $X^{+} = Y^{+}b, X^{-}=Y^{-}b$ where $Y^{+} = \Pos(M^{+}), Y^{-} = \Pos(M^{-})$, and $b$ indicates whether the anomaly occurs. Hence by the data processing inequality and formula of KL-divergence of Poisson random variables, 
\begin{align*}
    D_{KL}(X^{+}|| X^{-}) 
    &\leq D_{KL}(Y^{+} || Y^{-})\\
    &= (M^{+} \log (M^{+}/M^{-}) + M^{-} - M^{+}) \\
    &= - \log(1-\frac{c^{*}}{\sqrt{n}}) - \frac{c^{*}}{\sqrt{n}}\\
    &=   \frac{c^{*}}{\sqrt{n}} + \frac{(c^{*})^2}{2n} + \sum_{k=3}^{\infty}\frac{1}{k} (\frac{c^{*}}{\sqrt{n}})^{k} - \frac{c^{*}}{\sqrt{n}}\\
    &\leq  \frac{(c^{*})^2}{2n}  + \frac{1}{3} (\frac{c^{*}}{\sqrt{n}})^{3} \sum_{k=0}^{\infty}  (\frac{c^{*}}{\sqrt{n}})^{k} \\
    &\leq \frac{(c^{*})^2}{2n} + \frac{2c^{*}}{3} \frac{(c^{*})^2}{n^2} \leq \frac{(c^{*})^2}{n}. 
 \end{align*}
where $c^{*} < \frac{1}{2}$.
Similarly, 
\begin{align*}
D_{KL}(X^{-}|| X^{+}) 
&\leq D_{KL}(Y^{+} || Y^{-})\\
&= (M^{-} \log (M^{-}/M^{+}) + M^{+} - M^{-}) \\
&= (1-\frac{c^{*}}{\sqrt{n}}) \log (1-\frac{c^{*}}{\sqrt{n}}) + \frac{c^{*}}{\sqrt{n}}\\
&\leq  (1-\frac{c^{*}}{\sqrt{n}}) (-\frac{c^{*}}{\sqrt{n}}) + \frac{c^{*}}{\sqrt{n}}\\
&\leq \frac{(c^{*})^2}{n}.
\end{align*}

Hence, 
$$
 |p_{ij}(M^{a}) - p_{ij}(M^{b})|  \leq c^{*} = \frac{1}{2}e^{-M^{+}}.
$$
\end{proof}

Plug \cref{eq:bound-Mij} into \cref{eq:E-M-bound}, we then have
\begin{align*}
E_{M \sim \mathcal{M}_{n}}E_{X|M}(\cost^{\pi}(X) - \cost^{\pi^{*}}(X)) &\geq \frac{\delta}{2} \frac{1}{2}e^{-M^{+}} = \Omega\left(\frac{1}{\sqrt{n}}\right).
\end{align*}
This completes the proof.

\vspace{1em}

\vspace{1em}
\section{Additional Experimental Details}\label{sec:experiments-appendix}
In this section, we provide further implementation details of the experiments. 

\textbf{Computing Infrastructure.} all experiments are done in a personal laptop equipped with 2.6 GHz 6-Core Intel Core i7 and 16 GB 2667 MHz DDR4. The operating system is macOS Catalina. For each instance, the running time is within seconds for our algorithm. 

We present the implementation details of our algorithm and three state-of-the-arts. For practical consideration, we implemented a slight variant of the EW algorithm where (i) the matrix completion step used the typical soft impute algorithm~\citep{mazumder2010spectral}; (ii) the anomaly model estimation used MLE; and (iii) solving $\PAD$ by replacing $f_{ij}^{*}$ directly by $\frac{\hat{y}_{ij}}{\hat{x}_{ij}+\hat{y}_{ij}}$ when AUC curve is needed to generate. Given the observation $X_{\Omega}$, the soft impute algorithm solves the optimization problem $\min_{M} \norm{P_{\Omega}(X-M)}_{\F}^2 + \lambda \norm{M}_{*}$ where $\lambda$ is a hyper-parameter. To tune $\lambda$, we start with a small $\lambda$ and gradually increase it until the rank of the solution fits the true rank of $M^{*}$ (all other algorithms also use the knowledge of the true rank). In order to generate the AUC curve for each instance, we vary $\gamma$ in our algorithm. In the real data, the rank is identified through cross-validation. 

In the implementation of Stable-PCP, we solve the following optimization problem $(\hat{M}, \hat{A}) = \arg\min_{M, A} \norm{M}_{*} + \lambda \norm{A}_1 + \mu \norm{P_{\Omega}(M+A-X)}_{\F}^2$ by alternating optimization \citep{ma2018efficient}. The set of anomalies is identified from $\{(i,j)~|~\hat{A}_{ij} \neq 0\}.$ In order to choose suitable $(\lambda,\mu)$ and generate the AUC curve, note that when $\hat{M}$ fixed, the ratio of $\lambda/\mu$ decides the portion that will be classified as anomalies (i.e., different points on the AUC curve). Hence, we iterate the ratio $\lambda/\mu$ and then tune $\lambda$ (accordingly, $\mu$) such that the solution $\hat{M}$ fits the true rank of $M^{*}.$ This provides an AUC curve. 

In the DRMF algorithm, we implement the Algorithm 1 in \cite{xiong2011direct} to solve the following optimization problem $(\hat{M}, \hat{A}) = \arg\min \norm{P_{\Omega}(X-A-M)}_{\F}$ with the constraints $\text{rank}(\hat{M}) \leq r, \norm{A}_{0} \leq e$ (although \cite{xiong2011direct} does not consider the partial observation scenario, but the generalization to address missing entries is straightforward). The set of anomalies is identified from $\{(i,j)~|~\hat{A}_{ij} \neq 0\}.$ Here, we provide the true rank $r$ and vary $e$ for the DRMF algorithm to generate the AUC curve. 

For the RMC algorithm \cite{klopp2017robust}, the authors propose the following optimization problem $(\hat{M}, \hat{A}) = \arg\min_{M, A} \norm{M}_{*} + \lambda \norm{A}_1 + \mu \norm{P_{\Omega}(M+A-X)}_{\F}^2$ with constraints $\norm{M}_{\max} \leq a, \norm{A}_{\max} \leq a.$ This is effectively the Stable-PCP algorithm with the max norm constraints. We choose $a = k\norm{M^{*}}_{\max}$ for some constant scale $k>1.$ Then we implement RMC based on Stable-PCP and a projection of $(M, A)$ into the set with max norm constraints in every iteration during the alternating optimization. 

\section{Generalization to Detection Rate Optimization}\label{sec:cost-model}
Another interesting metric related to anomaly detection in inventory management, other than the average cost/benefits, is the rate of successfully detecting anomalies. In this section, we will generalize the results in the previous sections to the detection rate optimization problem and show that a variant of \cref{alg:EW} can achieve the optimal detection rate up to logarithmic factors. 

To make this precise, consider the goal of an algorithm $\pi$ is to correctly classify the entries into ``anomaly set'' and ``non-anomaly set''. This effectively amounts to a classification task, and as such we can measure performance via the standard true positive and false positive rates. Specifically, we will call an entry $(i, j)$ {\em positive} if $B_{ij} = 1$, and {\em negative} if $B_{ij} = 0.$ Furthermore, an entry $(i, j)$ is called ``true positive'' if $A_{ij}^{\pi} =  1$ and $B_{ij} = 1$; ``false positive'' if $A_{ij}^{\pi} = 1$ and $B_{ij} = 0.$ 

Consider the following definition for true positive and false positive rates. 
\begin{definition}
The true positive rate (TPR) of an algorithm $\pi$ given an observation $X_{\Omega}$ is denoted by
\begin{align}
    \TPR_{\pi}(X_{\Omega}) := \frac{\E{\sum_{(i, j)\in \Omega}\1{A_{ij}^{\pi}=1, B_{ij} = 1}\Big|X_{\Omega}}}{\E{\sum_{(i,j)\in \Omega} \1{B_{ij}=1}\big|X_{\Omega}}},\label{eq:TPR-definition}
\end{align}
i.e., the ratio between the expected number of true positive samples over the expected number of positive samples. Similarly, the false positive rate (FPR) of an algorithm $\pi$ given an observation $X_{\Omega}$ is denoted by
\begin{align}
    \FPR_{\pi}(X_{\Omega}) := \frac{\E{\sum_{(i,j)\in \Omega}\1{A_{ij}^{\pi}=1, B_{ij} = 0}\Big|X_{\Omega}}}{\E{\sum_{(i,j) \in \Omega} \1{B_{ij}=0}\big| X_{\Omega}}} \label{eq:FPR-definition}, 
\end{align}
i.e., the ratio between the expected number of false positive samples over the expected number of negative samples.\footnote{We suppose $\TPR_{\pi}(X_{\Omega}) = 0$ if $\E{\sum_{(i,j)\in \Omega} 1\{B_{ij}=1\}} = 0$; $\FPR_{\pi}(X_{\Omega}) = 0$ if $\E{\sum_{(i,j)\in \Omega} 1\{B_{ij}=0\}} = 0$.} 
\end{definition}
By the linearity of expectation, one can simplify \cref{eq:FPR-definition}:
\begin{align*}
 	\FPR_{\pi}(X_{\Omega}) = \frac{\sum_{(i,j)\in \Omega}\Prob{A_{ij}^{\pi}=1, B_{ij} = 0 ~\Big|~X_{\Omega}}}{\sum_{(i,j)\in \Omega}\Prob{B_{ij}=0~|~X_{\Omega}}}.
\end{align*}
Note that the output of $\pi$ solely depends on $X_{\Omega}$. Therefore, conditioned on $X_{\Omega}$, the variables $A_{ij}^{\pi}$ and $B_{ij}$ are independent. This implies that
 	\[\FPR_{\pi}(X_{\Omega}) = \frac{\sum_{(i,j)\in \Omega}\Prob{A_{ij}^{\pi}=1~|~X_{\Omega}} \Prob{B_{ij} = 0 ~|~X_{\Omega}}}{\sum_{(i,j)\in \Omega}\Prob{B_{ij}=0~|~X_{\Omega}}}.\]
 Let $f^{*}_{ij}:=\Prob{B_{ij} = 0~|~X_{\Omega}}$ be the probability that an entry $(i,j)$ does not have an anomaly given the observation $X_{\Omega}$. This provides a succinct characterization for FPR defined in \cref{eq:FPR-definition}:
\begin{align}
 	\FPR_{\pi}(X_{\Omega}) = \frac{\sum_{(i,j)\in \Omega}\Prob{A_{ij}^{\pi}=1~|~X_{\Omega}} f_{ij}^{*}}{\sum_{(i,j)\in \Omega}f_{ij}^{*}}.\label{eq:simplify-FPR}
\end{align}

Similarly, one can simplify TPR given in \cref{eq:TPR-definition} into:
\begin{align}
    \TPR_{\pi}(X_{\Omega}) 
    &= \frac{\sum_{(i,j)\in \Omega}\Prob{A_{ij}^{\pi}=1, B_{ij} = 1 ~\Big|~X_{\Omega}}}{\sum_{(i,j)\in \Omega}\Prob{B_{ij}=1~|~X_{\Omega}}} \nonumber\\
   &=\frac{\sum_{(i,j)\in \Omega}\Prob{A_{ij}^{\pi}=1~|~X_{\Omega}} \Prob{B_{ij} = 1 ~|~X_{\Omega}}}{\sum_{(i,j)\in \Omega}\Prob{B_{ij}=1~|~X_{\Omega}}} \nonumber\\
   &= \frac{\sum_{(i,j)\in \Omega}\Prob{A_{ij}^{\pi}=1~|~X_{\Omega}}(1-f_{ij}^{*})}{\sum_{(i,j)\in \Omega}(1-f_{ij}^{*})}.\label{eq:simplify-TPR}
\end{align}
Our goal will be to maximize $\TPR$ for some bound on $\FPR$. In establishing the quality of our algorithm we will compare, for a given constraint on $\FPR$, the $\TPR$ achieved under our algorithm to that achieved under the the (clairvoyant) optimal estimator that knows $M^{*}$, $p_{\A}^{*}$, and $\alpha^{*}$. We will show that in large matrices this gap grows negligibly small at a min-max optimal rate.

Besides the observed data $X_\Omega$, the only other input into the EW algorithm is a target FPR which we denote as $\gamma$. We propose \cref{alg:EW-rate} to maximize the TPR with a target FPR constraint. One can view \cref{alg:EW-rate} as a generalized version of \cref{alg:EW} with constraints in optimizing the decision rules. 

\begin{algorithm}
\caption{Entrywise Rate Optimization Algorithm $\pi^{\mathrm{EW}}(\gamma)$ } \label{alg:EW-rate}
{\bf Input:} $X_\Omega$, $\gamma \in (0,1]$
\begin{algorithmic}[1]
\State{Set \[\hat{M} = \frac{nm}{|\Omega|} \SVD(X_{\Omega})_{r}.\] Here, $ \SVD(X_{\Omega})_{r} := \arg\min_{\rank(M)\leq r} \norm{M-X'}_{\F}$, where $X'$ is obtained from $X_{\Omega}$ by setting unobserved entries to 0.}    
\State{Estimate $(\hat{p}_{\A}, \hat{\alpha})$ based on a moment matching estimator. } 
\State{Estimate a confidence interval $[f_{ij}^{\LL},f_{ij}^{\RR}]$ for $f_{ij}^{*}$ for $(i, j) \in \Omega$. }
\State{Let $\{t_{ij}^{\AD}\}$ be an optimal solution to the following optimization problem:
    \begin{align*}
        \PAD: \max_{\{0\leq t_{ij}\leq 1, (i, j)\in\Omega\}}& \sum_{(i,j) \in \Omega} t_{ij}\\
        \text{subject to}& \sum_{(i, j) \in \Omega} t_{ij} f_{ij}^{\RR} \leq \gamma \sum_{(i, j) \in \Omega} f_{ij}^{\LL}
    \end{align*} }
For every $(i, j) \in \Omega$, generate $A_{ij} \sim \mathrm{Ber}(t_{ij}^{\AD})$ independently. 
\end{algorithmic}
{\bf Output:} $A_\Omega$
\end{algorithm}

Step 1 and 2 in \cref{alg:EW-rate} are the same as \cref{alg:EW}. Step 3 uses plug-in confidence interval estimators instead point estimators (the full details will be specified in \cref{eq:conf_int}). Step 4 solves an constrained optimization problem $\PAD$ to determine the decision rules. 

The goal of the \cref{alg:EW-rate} is to maximize the TPR subject to a FPR below the input target value of $\gamma$. Our result is the following guarantee, which states that (a) the `hard' constraint on the FPR is satisfied with high probability, and (b) the TPR is within an additive {\em regret} of a certain unachievable policy we use as a proxy for the best achievable policy. Specifically, for any $\gamma \in (0,1]$, let $\pi^*(\gamma)$ denote the optimal policy when $M^*$, $p_{\A}^*$, and $\alpha^*$ are known (this policy is described later in this section). One can verify that, for any $\gamma$, $X_{\Omega}$ and policy $\pi$, $\TPR_{\pi^{*}(\gamma)}(X_{\Omega}) \geq \TPR_{\pi}(X_{\Omega})$ if $\FPR_{\pi}(X_{\Omega}) \leq \gamma.$

\begin{theorem}\label{thm:main-theorem-rate}
Assume that the regularity conditions $\mathrm{(RC)}$ hold. With probability $1-O(\frac{1}{nm})$, for any $0<\gamma \leq 1$, 
\begin{align*} 
\FPR_{\pi^{\AD}(\gamma)}(X_\Omega) &\leq \gamma, \\
\TPR_{\pi^{\AD}(\gamma)}(X_\Omega) &\geq \TPR_{\pi^{*}(\gamma)}(X_\Omega)  - C\frac{(K+L)^3L^3 \kappa^4 \mu r}{p_{\A}^{*}  \gamma}\frac{\log^{1.5}(m)}{\sqrt{p_\O m}}.
\end{align*}
\end{theorem}

To parse this result, consider that in a typical application, we can expect the problem parameters to fall in the following scaling regime: $K, L, \kappa, r, \mu = O(1)$, $p_\O,p_\A^{*},\gamma = \Omega(1)$, and $m/n = \Theta(1)$. For this regime, the regret is $O\left(n^{-1/2}\log^{1.5}{n}\right)$, which is in fact optimal up to logarithmic factors. To be precise,
we fix a particular value of $\gamma$ for which the following proposition states that, for any $n$, there exists a family of anomaly models $\mathcal{M}_{n}$ for which no algorithm can achieve a regret on $\TPR$  lower than $O(n^{-1/2})$ across all models within the family. To allow for direct comparison to \cref{thm:main-theorem}, let $\Pi_\gamma$ denote the set of all policies $\pi$ such that 
\begin{align*}
    \Probx{X_{\Omega}|M^*}{\FPR_{\pi}(X_{\Omega}) \leq \gamma} \geq 1-C/n^2 \;\;\; \text{ for all } M^* \in \mathcal{M}_{n}.
\end{align*}

\begin{proposition}\label{prop:lower-bound-rate}
For any algorithm $\pi \in \Pi_\gamma$, there exists $M^{*} \in \mathcal{M}_{n}$ such that
\begin{align*}
    \Ex{X_{\Omega}|M^*}{\TPR_{\pi^{*}(\gamma)}(X_{\Omega}) - \TPR_{\pi}(X_{\Omega})} \geq  C/\sqrt{n}.
\end{align*}
\end{proposition}
The proof of \cref{prop:lower-bound-rate} uses the same construction as the proof of \cref{prop:lower-bound}, which is omitted for simplicity. 

\vspace{1em}
\subsection{Steps 3--4: Confidence Intervals and the Optimization Problem $\PAD$}

Recall that the plug-in point estimator $\hat{f}_{ij}$ has been used for $f^{*}_{ij}$ in \cref{sec:plug-in-f} 
\begin{align*}
\hat{f}_{ij} = \frac{\hat{y}_{ij}}{\hat{x}_{ij} + \hat{y}_{ij}}.
\end{align*}
We construct the confidence interval by simply finding a small interval that centers at $\hat{f}_{ij}$, as shown below.
\begin{lemma}\label{lem:confidence-interval}
Let \[\delta =  (K+L)^3\kappa^4 \mu r L^2 \sqrt{\frac{\log m}{p_{\O}m}}.\] There exists a (known) constant $C_1$ such that, if 
\begin{equation} \label{eq:conf_int}
	f_{ij}^{\LL} := \left[\frac{\hat{y}_{ij}-C_1\delta}{\hat{x}_{ij}+\hat{y}_{ij}}\right]\;\;\; \text{ and } \;\;\; f_{ij}^{\RR} := \left[\frac{\hat{y}_{ij}+C_1\delta}{\hat{x}_{ij}+\hat{y}_{ij}}\right],
\end{equation}
then with probability $1-O(\frac{1}{nm})$, for every $(i, j) \in \Omega$, we have \[f_{ij}^{\LL} \leq f_{ij}^{*} \leq f_{ij}^{\LL}+\epsilon_{ij} \;\;\; \text{ and } \;\;\; f_{ij}^{\RR} - \epsilon_{ij} \leq f_{ij}^{*} \leq f_{ij}^{\RR},\] where $\epsilon_{ij} = \min(4C_1\delta/(x_{ij}^{*}+y_{ij}^{*}), 1)$.
\end{lemma}

The final step involves solving   $\PAD$. To motivate its particular form, consider the `ideal' anomaly detection algorithm if the $f_{ij}^{*}$'s were known. Intuitively, one should identify anomalies at entries with the smallest values of $f_{ij}^{*}$. This leads to the following idealized algorithm, which we will call $\pi^{*}(\gamma)$:
\begin{enumerate}
\item Let $\{t_{ij}^{*}\}$ be an optimal solution to the following optimization problem.
    \begin{align*}
     \Ps: \max_{\{0\leq t_{ij}\leq 1, (i, j)\in\Omega\}}& \sum_{(i,j) \in \Omega} t_{ij} \\
        \text{subject to}& \sum_{(i, j) \in \Omega} t_{ij} f_{ij}^{*} \leq \gamma\sum_{(i, j) \in \Omega} f_{ij}^{*}
    \end{align*}
\item For every $(i, j) \in \Omega$, generate $A_{ij} \sim \mathrm{Ber}(t_{ij}^*)$ independently.
\end{enumerate}
%
%
The following claim establishes the optimality of $\pi^{*}(\gamma).$
\begin{claim}
For any $\pi, \gamma$, and $X_{\Omega}$, if $~\FPR_{\pi}(X_\Omega) \leq \gamma$, then $\TPR_{\pi}(X_\Omega) \leq \TPR_{\pi^{*}(\gamma)}(X_\Omega).$
\end{claim}

%
%


Now notice that $\PAD$ is obtained from $\Ps$ by replacing $f_{ij}^{*}$ with the confidence interval estimators $f_{ij}^{\LL}$ and $f_{ij}^{\RR}$ defined in the previous step. Intuitively, we could expect that $\PAD \approx \Ps$, and therefore the algorithm $\pi^{\AD}$ should achieve the desired performance. In fact, $\FPR_{\pi^{\AD}(\gamma)}(X) \leq \gamma$ holds immediately because $f_{ij}^{\LL} \leq f_{ij}^{*} \leq f_{ij}^{\RR}$ and so $\{t_{ij}^{\AD}\}$ is a feasible solution of $\Ps.$  The guarantee for $\TPR_{\pi^{\AD}}(X)$ can be established based on a fine-tuned analysis of \cref{lem:confidence-interval}. See the Appendix for the formal proof.

\section{Proof of \cref{thm:main-theorem-rate}}
\subsection{Analysis of the optimization problem $\PAD$}
Note that $\PAD$ is obtained from $\Ps$ by replacing $f_{ij}^{*}$ with the confidence interval estimators $f_{ij}^{\LL}$ and $f_{ij}^{\RR}$. Intuitively, we could expect that $\PAD \approx \Ps$, and therefore the algorithm $\pi^{\AD}$ should achieve the desired performance. We first have the following lemma to show that $\FPR_{\pi^{\AD}(\gamma)}(X) \leq \gamma$ since $f_{ij}^{\LL} \leq f_{ij}^{*} \leq f_{ij}^{\RR}$ and so $\{t_{ij}^{\AD}\}$ is a feasible solution of $\Ps.$  
\begin{lemma}\label{claim:FRP-bound}
With probability $1-O(1/(nm))$, for any $0<\gamma \leq 1$
\begin{align*}
    \FPR_{\pi^{\AD}(\gamma)}(X_{\Omega}) \leq \gamma.
\end{align*}
\end{lemma}
\begin{proof}{Proof.}
This is because 
\begin{align*}
    \sum_{(i,j) \in \Omega} t_{ij}^{\AD} f_{ij}^{*} 
    \leq \sum_{(i,j) \in \Omega} t_{ij}^{\AD} f^{\RR}_{ij}
    \leq \gamma\sum_{(i,j) \in \Omega} f^{\LL}_{ij}
    \leq \gamma\sum_{(i, j) \in \Omega} f_{ij}^{*}.
\end{align*}
due to that $f_{ij}^{\LL} \leq f_{ij}^{*} \leq f_{ij}^{\RR}$ and the constraint of $t_{ij}^{\AD}$. 
\end{proof}

To show the desired performance guarantee for $\TPR_{\pi^{\AD}}(X)$, we provide the following Lemma that characterizes how $f_{ij}^{\LL}$ and $f_{ij}^{\RR}$ are close to $f_{ij}^{*}$ in an accumulated manner (the proof is shown momentarily):  
\begin{lemma}\label{lem:f-ij-bound}
Let $\delta =  (K+L)^3\kappa^4 \mu r L^2 \sqrt{\frac{\log(m)}{p_{\O}m}}.$  With probability $1-O(\frac{1}{nm})$, 
\begin{align*}
\sum_{(i,j)\in \Omega}\left(|f_{ij}^{\LL}-f_{ij}^{*}|+|f_{ij}^{\RR}-f_{ij}^{*}|\right) \leq CL\log(m)\delta p_\O nm.
\end{align*}
\end{lemma}

Next we proceed to the analysis of $\PAD$. For a fixed $\eta$, let $\{t_{ij}'\}$ be the optimal solution of $\pi^{*}(\gamma')$ for some $\gamma'$ such that $\frac{\sum_{(i,j)\in\Omega} t_{ij}'}{\sum_{(i,j)\in\Omega} t_{ij}^{*}} = \eta < 1.$ The key idea is to find some $\eta$ such that $\{t_{ij}'\}$ is a feasible solution of $\PAD$, while maintaining good $\TPR$ performance compared to $\pi^{*}(\gamma)$. Indeed, a sufficiently large $\eta$ can be achieved by \cref{lem:f-ij-bound}. In particular, we have (the proof is shown momentarily):
\begin{lemma}\label{lem:eta-bound}
Let $\delta =(K+L)^3\kappa^4 \mu r L^2 \sqrt{\frac{\log(m)}{p_{\O}m}}, \eta = 1-CL\delta \log(m)/\gamma$. Then $\{t_{ij}'\}$ is a feasible solution of $\PAD$. Furthermore, $\min\left(1, \frac{\sum_{(i,j)\in\Omega} t_{ij}^{*} - \sum_{(i,j)\in\Omega} t_{ij}'}{\sum_{(i,j)\in\Omega} (1-f_{ij}^{*})}\right) \leq C_1\frac{L\delta \log(m)}{\gamma p_{\A}^{*}}$
for a constant $C_1$.
\end{lemma}

\vspace{1em}
\subsubsection{Proof of Lemma \ref{lem:f-ij-bound}}
Next, we prove \cref{lem:f-ij-bound}, i.e., show that the accumulated error induced by the approximation of $f_{ij}^{*}$ by $f_{ij}^{\LL}$ and $f_{ij}^{\RR}$ has the desired bound. 

\begin{proof}{\textbf{Proof of \cref{lem:f-ij-bound}}}
Let $x_{ij} := p_{\A}^{*}\Probx{\Ano}{X_{ij}|\alpha^{*}, M_{ij}^{*}}, y_{ij} := (1-p_{\A}^{*})\Probx{\Pos}{X_{ij}|M_{ij}^{*}}.$ By \cref{lem:confidence-interval}, 
\begin{align*}
\max(|f_{ij}^{\LL} - f_{ij}^{*}|, |f_{ij}^{\RR} - f_{ij}^{*}|) \leq \epsilon_{ij}
\end{align*}
where $\epsilon_{ij} := \min\left( \frac{4C\delta}{x_{ij}+y_{ij}}, 1\right)$ for some constant $C$ and $\delta =   (K+L)^3\kappa^4 \mu r L^2 \sqrt{\frac{\log(m)}{p_{\O}m}}.$

Note that when $X_{ij} = t$,
\begin{align*}
x_{ij} + y_{ij} = \Prob{X_{ij} = t}.
\end{align*}

Note that $\norm{X_{ij}}_{\psi_1} \lesssim L$ is a sub-exponential random variable by \cref{lem:poisson-bound,lem:combination-bound}. Then, we have
\begin{align*}
&\Prob{X_{ij} > t} \leq \exp^{-t/C'L} \\ 
\implies &\Prob{X_{ij}' > C'L\log(1/\delta)} \leq \delta  
\end{align*}
where $C'$ is a proper constant. Let $z_{ij} = \min\left(\frac{\delta}{\Prob{X_{ij} = t}}, 1\right).$
Then,
\begin{align*}
\E{z_{ij}} &= \sum_{t=0}^{\infty} \min\left(1, \delta/\Prob{X_{ij} = t}\right) \Prob{X_{ij} = t}\\
&\leq \sum_{t=0}^{C'L\log(1/\delta)} \delta + \sum_{t=C'L\log(1/\delta)+1}^{\infty} \Prob{X_{ij}=t}\\
&\leq C'L\log(1/\delta)\delta + \delta.
\end{align*}

Note that $z_{ij} \in [0, 1]$ are independent random variables. Then, by \cref{lem:concentration-bound}, with probability $1-O(\frac{1}{nm})$, 
\begin{align*}
\sum_{(i,j)\in \Omega} z_{ij} 
&\lesssim L\log(1/\delta)\delta p_{\O} nm + \sqrt{p_\O nm \log (nm)}\\
&\lesssim L\log(m)\delta p_{\O} nm
\end{align*}
given that $\delta \gtrsim \sqrt{\frac{\log(m)}{p_\O m}}.$

Therefore,
\begin{align*}
\sum_{(i, j) \in \Omega} \max(|f_{ij}^{\LL} - f_{ij}^{*}|, |f_{ij}^{\RR} - f_{ij}^{*}|) \leq \sum_{(i, j) \in \Omega} \epsilon_{ij} \lesssim  \sum_{(i,j)\in \Omega} z_{ij}  \lesssim L\log(m)\delta p_{\O} nm.
\end{align*}
\end{proof}

\subsubsection{Proof of Lemma \ref{lem:eta-bound}}
Consider a concentration bound 
\begin{lemma}\label{lem:bound-N}
Let $C_1, C_2, C_3$ be constants. With probability $1-O(\frac{1}{nm}),$
\begin{align*}
 \sum_{(i,j)\in \Omega} f_{ij}^{*} &\geq C_1nmp_{\O}\\
  |\Omega| &\leq C_2 nmp_{\O}.
\end{align*}
Furthermore, if $p_{\A}^{*}p_{\O}nm \gtrsim \log(nm),$
\begin{align*}
\sum_{(i,j)\in \Omega} 1-f_{ij}^{*} \geq C_3 p_{\A}^{*}p_{\O} nm.
\end{align*}

\end{lemma}
\begin{proof}{Proof.}
Let $Z_{ij} = \Prob{B_{ij}=1|X_{ij}}$. Then $\sum_{(i,j)\in \Omega} 1-f_{ij}^{*} = \sum_{(i,j)\in \Omega} Z_{ij}$. Note that $\E{Z_{ij}} = p_{\A}^{*}$ and $Z_{ij} \in [0, 1]$ are independent. Hence, by \cref{lem:concentration-bound}, with probability $1-O(\frac{1}{nm})$, $\sum_{(i,j)\in \Omega} 1-f_{ij}^{*} \geq C p_{\A}^{*}p_{\O} nm$ where $C$ is a constant given that $p_{\A}^{*}p_{\O} nm \gtrsim \log(nm)$ Similar results for $ \sum_{(i,j)\in \Omega} f_{ij}^{*}$ (with $1-p_{\A}^{*} \geq c$ for some constant $c$) and $|\Omega|$ can also be obtained. 
\end{proof}
\begin{proof}{\textbf{Proof of \cref{lem:eta-bound}}}
Let $\{t_{ij}', (i, j) \in \Omega\}$ be the optimal solution of the algorithm $\pi^{*}(\gamma').$ Let $\{t_{ij}^{*}, (i, j) \in \Omega\}$ be the optimal solution of $\pi^{*}(\gamma)$. Suppose 
\begin{align*}
\frac{\sum_{(i, j) \in \Omega} t_{ij}'}{\sum_{(i, j) \in \Omega} t_{ij}^{*}} = \eta < 1.
\end{align*}

Order $f_{ij}^{*}$ by $f_{a_1b_1}^{*} \leq f_{a_2 b_2}^{*} \leq \dotsc \leq f_{a_{|\Omega|}b_{|\Omega|}}^{*}.$ One can easily verify that $t_{a_1b_1}' \leq t_{a_1b_1}^{*}, t_{a_2b_2}' \leq t_{a_2b_2}^{*}, \dotsc, t_{a_{|\Omega|}b_{|\Omega|}}' \leq t_{a_{|\Omega|}b_{|\Omega|}}^{*}$. Furthermore, for any $k$ and $l$ such that $t_{a_{k}b_{k}}^{'} > 0$ and $t^{*}_{a_{l}b_{l}} - t_{a_{l}b_{l}}' > 0$, we have $f_{a_{k}b_{k}}^{*} \leq f_{a_{l}b_{l}}^{*}.$ Let $A=\sum_{ij} t'_{ij}, B = \sum_{ij} t^{*}_{ij} - t'_{ij}, C = \sum_{ij} t'_{ij}f_{ij}^{*}, D = \sum_{ij}(t^{*}_{ij}-t'_{ij})f_{ij}^{*}.$ Then the following weighted average inequality holds: $\frac{C}{A} \leq \frac{D}{B}.$ This implies that $\frac{C}{A} \leq \frac{C+D}{A+B}$, i.e., 
\begin{align}
\frac{1}{\sum_{(i, j) \in \Omega} t_{ij}'}\sum_{(i,j) \in \Omega} t_{ij}' f_{ij}^{*} \leq \frac{1}{\sum_{(i, j) \in \Omega} t_{ij}^{*}}\sum_{(i,j) \in \Omega} t_{ij}^{*} f_{ij}^{*}. \label{eq:ideal-bound}
\end{align}
This implies that $\sum_{(i,j) \in \Omega} t_{ij}' f_{ij}^{*} \leq \eta \sum_{(i,j) \in \Omega} t_{ij}^{*} f_{ij}^{*}.$ Then, we have, 
\begin{align*}
    \sum_{(i,j)\in \Omega} t_{ij}' f_{ij}^{\RR} 
    &\leq  \sum_{(i,j) \in \Omega} t_{ij}'(f^{*}_{ij} + |f_{ij}^{\RR}-f_{ij}^{*}|) \\
    &\leq  \left(\eta\sum_{(i, j) \in \Omega}t_{ij}^{*} f^{*}_{ij}\right) + \sum_{(i,j)\in \Omega} |f_{ij}^{\RR}-f_{ij}^{*}|& \text{by \cref{eq:ideal-bound} and $0\leq t_{ij}'\leq 1$}\\
    &\leq \left( \gamma \eta\sum_{(i,j)\in \Omega} f_{ij}^{*}\right) + \sum_{(i,j)\in \Omega} |f_{ij}^{\RR}-f_{ij}^{*}|  & \sum_{(i,j)\in \Omega} t_{ij}^{*} f_{ij}^{*} \leq \gamma \sum_{(i,j)\in \Omega} f_{ij}^{*}\\
    &\leq \gamma \sum_{(i,j) \in \Omega} f_{ij}^{*} + \sum_{(i,j)\in \Omega}|f_{ij}^{\RR}-f_{ij}^{*}|  -\gamma(1-\eta) \sum_{(i,j)\in \Omega} f_{ij}^{*}.
\end{align*}
Note that
\begin{align*}
\gamma \sum_{(i, j) \in \Omega} f_{ij}^{*} \leq \gamma \sum_{(i, j) \in \Omega} f_{ij}^{\LL} + \sum_{(i, j) \in \Omega} |f_{ij}^{*} - f_{ij}^{\LL}|.
\end{align*}
Therefore, we have
\begin{align*}
\sum_{(i,j)\in \Omega} t_{ij}' f_{ij}^{\RR}  \leq \gamma \sum_{(i, j) \in \Omega} f_{ij}^{\LL} + \left(\sum_{(i, j) \in \Omega} \left(|f_{ij}^{*} - f_{ij}^{\LL}| + |f_{ij}^{*}-f_{ij}^{\RR}|\right)\right) -\gamma(1-\eta) \sum_{(i,j)\in \Omega} f_{ij}^{*}.
\end{align*}

By \cref{lem:f-ij-bound}, we have $\left(\sum_{(i, j) \in \Omega} \left(|f_{ij}^{*} - f_{ij}^{\LL}| + |f_{ij}^{*}-f_{ij}^{\RR}|\right)\right) \leq C_1 L \log(m)\delta p_\O nm. $ By \cref{lem:bound-N}, we have $\gamma(1-\eta) \sum_{(i,j)\in \Omega} f_{ij}^{*} \geq C_2 \gamma (1-\eta) p_{\O}nm.$ Take $\eta = 1-\frac{C_1}{C_2\gamma } L \log(m) \delta.$ We then have $\{t_{ij}'\}$ is a feasible solution of $\PAD$:
\begin{align*}
\sum_{(i,j) \in \Omega} t_{ij}' f_{ij}^{\RR} \leq   \gamma \sum_{(i, j) \in \Omega} f_{ij}^{\LL}.
\end{align*}

Furthermore, for any $0<\gamma \leq 1$, we can get 
\begin{align*}
      \frac{\sum_{(i,j)\in \Omega} (t_{ij}^{*}-t_{ij}')}{\sum_{(i, j) \in \Omega} (1-f_{ij}^{*})}  = \frac{(1-\eta) \sum_{(i, j) \in \Omega} t_{ij}^{*}}{\sum_{(i, j) \in \Omega} (1-f_{ij}^{*})}.\end{align*}
By \cref{lem:bound-N}, $\sum_{(i,j)\in\Omega} t_{ij}^{*} \leq |\Omega| \lesssim nmp_\O.$ Suppose $p_{\A}^{*}p_{\O} nm \gtrsim \log(nm)$, then by \cref{lem:bound-N}, $\sum_{(i, j) \in \Omega} (1-f_{ij}^{*}) \gtrsim nmp_\O p_{\A}^{*}.$ This leads to
\begin{align}
\frac{\sum_{(i,j)\in \Omega} (t_{ij}^{*}-t_{ij}')}{\sum_{(i, j) \in \Omega} (1-f_{ij}^{*})}  \lesssim \frac{(1-\eta)p_{\O}nm}{p_{\O}p_{\A}^{*}nm} \lesssim \frac{L\log(m) \delta}{\gamma p_{\A}^{*}}.
\label{eq:TPR-bound-here}
\end{align}
Note that $\delta \gtrsim \frac{1}{\sqrt{p_{\O}m}}$. Suppose $p_{\A}^{*}p_{\O}nm \lesssim \log(nm)$, then
\begin{align*}
\frac{L\log(n) \delta}{\gamma p_{\A}^{*}} \gtrsim \frac{1}{p_{\A}^{*}\sqrt{p_{\O} m}} \gtrsim \frac{nm\sqrt{p_\O}}{\log(nm)\sqrt{m}} \gtrsim 1. 
\end{align*}
This completes the proof. 
\end{proof}

\vspace{1em}
\subsection{Proof of Theorem \ref{thm:main-theorem-rate}}
\begin{proof}{\textbf{Proof of \cref{thm:main-theorem-rate}}}
Finally, we proceed the proof of  \cref{thm:main-theorem-rate}. Note that
\begin{align*}
  &\TPR_{\pi^{*}(\gamma)}(X_{\Omega}) - \TPR_{\pi^{\AD}(\gamma)}(X_{\Omega})\\
    &= \frac{\sum_{(i,j)\in \Omega}t_{ij}^{*}(1-f^{*}_{ij}) - \sum_{(i,j)\in \Omega} t_{ij}^{\AD}(1-f^{*}_{ij})}{\sum_{(i, j) \in \Omega} (1-f_{ij}^{*})} \\
    &\leq \frac{\sum_{(i,j)\in \Omega} (t_{ij}^{*} - t_{ij}^{\AD}) + (\sum_{(i, j) \in \Omega} t^{\AD}_{ij} f_{ij}^{*} - \sum_{(i, j) \in \Omega} t^{*}_{ij} f_{ij}^{*})}{\sum_{(i, j) \in \Omega} (1-f_{ij}^{*})}.
\end{align*}
Note that $\sum_{(i, j)\in \Omega}t_{ij}^{*} f_{ij}^{*} = \gamma \sum_{(i, j) \in \Omega} f_{ij}^{*}$ and $\sum_{(i, j)\in \Omega}t_{ij}^{\AD} f_{ij}^{*} \leq \gamma \sum_{(i, j) \in \Omega} f_{ij}^{*}$ by \cref{claim:FRP-bound}. Furthermore, $\sum_{(i, j) \in \Omega} t_{ij}^{\AD} \geq \sum_{(i, j) \in \Omega} t_{ij}'$ since $\{t_{ij}'\}$ is a feasible solution of $\PAD$ and the objective function of $\PAD$ maximizes $\sum_{(i, j) \in \Omega} t_{ij}^{\AD}$ given the constraint. Hence,  

\begin{align*}
    \TPR_{\pi^{*}(\gamma)}(X_{\Omega}) - \TPR_{\pi^{\AD}(\gamma)}(X_{\Omega}) \leq \frac{\sum_{(i,j)\in \Omega} (t_{ij}^{*}-t_{ij}^{\AD})}{\sum_{(i, j) \in \Omega} (1-f_{ij}^{*})} \leq \frac{\sum_{(i,j)\in \Omega} (t_{ij}^{*}-t_{ij}^{'})}{\sum_{(i, j) \in \Omega} (1-f_{ij}^{*})}.
    \end{align*}
Also, note that $\TPR_{\pi^{*}(\gamma)}(X_{\Omega}) - \TPR_{\pi^{\AD}(\gamma)}(X_{\Omega}) \leq 1$ since $\TPR \leq 1$ by definition. By \cref{lem:eta-bound}, 
\begin{align*}
 \TPR_{\pi^{*}(\gamma)}(X_{\Omega}) - \TPR_{\pi^{\AD}(\gamma)}(X_{\Omega})  \lesssim \frac{L\log(m) \delta}{\gamma p_{\A}^{*}},
\end{align*}
which completes the proof.
\end{proof}

\vspace{1em}

\vspace{1em}
\end{appendices}

\end{document}